\documentclass[smallcondensed]{svjour3}     
\smartqed  
\usepackage{times}
\usepackage{graphicx}
\usepackage{amsmath}
\usepackage{amssymb}
\usepackage{array}
\usepackage{multirow}
\usepackage{tabu}
\usepackage{subcaption}
\usepackage{color}
\usepackage{fancyhdr}
\usepackage{geometry}
\begin{document}
\date{Received: date / Accepted: date}

\title{Local Directional Relation Pattern for Unconstrained and Robust Face Retrieval}

\author{Shiv Ram Dubey}
\institute{S.R. Dubey \at Computer Vision Group \\ Indian Institute of Information Technology, Sri City, Chittoor, Andhra Pradesh - 517646, India \\
Tel.(Office): +91-7337324915 \\
\email{srdubey@iiits.in}
}

\maketitle
\thispagestyle{fancy}
\fancyhf{}
\lfoot{This paper is published in Multimedia Tools and Applications, Springer. Cite this article as:
S.R. Dubey, ``Local Directional Relation Pattern for Unconstrained and Robust Face Retrieval", \textit{Multimedia Tools and Applications}, 2019. https://doi.org/10.1007/s11042-019-07908-3}

\begin{abstract}
Face recognition is still a very demanding area of research. This problem becomes more challenging in unconstrained environment and in the presence of several variations like pose, illumination, expression, etc. Local descriptors are widely used for this task. The most of the existing local descriptors consider only few immediate local neighbors and not able to utilize the wider local information to make the descriptor more discriminative. The wider local information based descriptors mainly suffer due to the increased dimensionality. In this paper, this problem is solved by encoding the relationship among directional neighbors in an efficient manner. The relationship between the center pixel and the encoded directional neighbors is utilized further to form the proposed local directional relation pattern (LDRP). The descriptor is inherently uniform illumination invariant. The multi-scale mechanism is also adapted to further boost the discriminative ability of the descriptor. The proposed descriptor is evaluated under the image retrieval framework over face databases. Very challenging databases like PaSC, LFW, PubFig, ESSEX, FERET, AT\&T, and FaceScrub are used to test the discriminative ability and robustness of LDRP descriptor. Results are also compared with the recent state-of-the-art face descriptors such as LBP, LTP, LDP, LDN, LVP, DCP, LDGP and LGHP. Very promising performance is observed using the proposed descriptor over very appealing face databases as compared to the existing face descriptors. The proposed LDRP descriptor also outperforms the pre-trained ImageNet CNN models over large-scale FaceScrub face dataset. Moreover, it also outperforms the deep learning based DLib face descriptor in many scenarios.

\keywords{Local Descriptor \and Face \and Unconstrained \and Robust \and Retrieval \and Directional Relation}

\end{abstract}

\section{Introduction}
\subsection{Motivation}
Unconstrained and robust face recognition is the current demand for the betterment of the quality life. Most of the early days research has been conducted in a very controlled environment, where users have given their facial images in frontal pose, under consistent lighting, without glasses or occlusion, etc. Some researchers also tried to develop the face recognition approaches robust for specific geometric and photometric changes such as pose, illumination, motion blur, etc. \cite{wright2009}, \cite{ding2015multi}, \cite{kan2016multi}, \cite{ding2012}, \cite{abhi2015}. The face recognition approaches are surveyed time to time by many researchers \cite{zhao2003face}, \cite{zhang2009face}, \cite{ding2016comprehensive}. 

The face recognition approaches are categorized into three major areas, namely deep learning based face recognition \cite{facenet}, \cite{deepface}, traditional learning based face recognition \cite{cao2010}, \cite{wolf2011}, \cite{lei2014}, \cite{lu2015learning}, \cite{lu2015simultaneous}, and hand-crafted feature based face recognition \cite{ldp}, \cite{lvp}, \cite{lghp}. The deep learning based approaches are being popular due to high performance, but at the cost of increased complexity in terms of the time, computing power and data size. The deep learning based approaches are also biased towards the training data. 
The main drawback of the traditional learning based descriptors are the dependency over the training database and vocabulary size. The hand-designed local descriptors are very simple from design aspect. This class of descriptors have shown very promising performance in most of the computer vision problems \cite{lbpbook}. Some typical applications are local image matching \cite{iold}, image retrieval \cite{ltrp}, \cite{mdlbp}, texture classification \cite{lbptexture}, \cite{dlbp}, \cite{brint}, \cite{priclbp}, medical image retrieval \cite{ldep}, \cite{lbdp}, \cite{lwp}, 3D face recognition \cite{tang20133d}, \cite{elaiwat2014}, palmprint recognition \cite{hong2015novel}, activity recognition \cite{liu2015action2activity}, \cite{liu2016action}, \cite{liu2016recognizing}, etc. The main advantages of the handcrafted local descriptors are as follows: a) it is not dependent upon the database, b) it does not require very complex computing facility, and c) lower dimensional descriptors can boost the time efficiency significantly. 

\subsection{Related Works}
Several face descriptors have been investigated in the last decade. Ahonen et al. applied the local binary pattern (LBP) for the face recognition \cite{lbp}. Inspired from the simplicity and success of LBP, several variants are proposed for the face recognition \cite{huang2011local}, \cite{yang2013comparative}. Some researchers tried to use the transformations like Gabor, Wavelet and Weber in the framework of LBP \cite{lgbphs}, \cite{hgpp}, \cite{ahonen2008}, \cite{wld}, \cite{xie2010fusing}. The drawback of such transformation based descriptors is that the time and space complexity of the transformations are too high for practical face recognition.

Utilizing the gradient information is very common in LBP like descriptors. The binary concept of LBP is extended in ternary by local ternary pattern (LTP) \cite{ltp}. The local derivative pattern (LDP) computes the LBP over derivative images in four directions \cite{ldp}. The average values of block subregions are used to compute the multi-scale block local binary pattern \cite{mblbp}. Gradient edge map features are computed using a cascade of processing steps for illumination invariant frontal face representation \cite{gemf}. The relationships between gradient orientations and magnitudes like patterns of orientation difference and patterns of oriented edge magnitudes are exploited by Vu for face recognition \cite{vu2013}. Recently, Local intensity orders over gradient images are used to develop the local gradient order pattern for face recognition \cite{lgop}. Lumini et al. combined the multiple descriptors for face recognition \cite{lumini2017}. These descriptors encode the gradient information in a local neighborhood in circular fashion, but miss to utilize the directional information across different radius of the local neighborhood.

Some of the LBP variants tried to utilize the directions in order to improve it. Local directional number (LDN) pattern uses the masks to represent the image into different directions and encodes the directional numbers and sign \cite{ldn}. Local vector pattern (LVP) uses the pairwise direction of the vector with diverse distances for each pixel to represent the face image \cite{lvp}. Jabid et al. utilized the relative magnitude's strength of edge responses in eight directions to compute the local directional pattern \cite{jabid2010facial}. A single eight bit code for a block is used to reduce the dimensionality of the local directional pattern \cite{drldp}. Local directional gradient pattern (LDGP) uses the four directions to capture the local information for recognition \cite{ldgp}. The major drawback of these approaches is the computation of directional information as a separate stage similar to pre-processing. These approaches also do not consider the directional relationship at different radius. 

Some descriptors tried to consider the wider neighborhood to increase the discriminative ability of the descriptor. The LBP \cite{lbp} considers the neighbors at radius one, i.e., immediate neighbors. Basically, wider meighborhood refers to the local region with larger radius. Local quantized pattern (LQP) quantizes the binary code generated from the large neighborhood to reduce the dimensionality \cite{lqp}. The LQP is computed over regional features of the image and combined to form the multiscale LQP \cite{chan2013}. Dual cross pattern (DCP) considers the local neighbors at two radius with first derivative of Gaussian operator to encode the directional information \cite{dcp}. In order to reduce the dimension, DCP divides the local neighborhood into two groups: 1) horizontal and vertical neighbors, and 2) diagonal neighbors. The local directional ternary pattern (LDTP) converts the image into eight directional images using Robinson compass masks and then finds the primary and secondary directions to generate the feature vector \cite{ldtp}. Recently, the local gradient hexa pattern (LGHP) has been proposed for the face recognition and retrieval \cite{lghp}. LGHP basically works by encoding the relationship of center pixel with its neighboring pixels at different distances across different derivative directions. The major problem associated with these existing descriptors is the increased dimension while considering more local neighbors. The relationship between different neighbors in a particular direction is also not utilized by these descriptors. 

\subsection{Major Contribution}
It is pointed out from the related works that most of the descriptors use only immediate local neighbors which decreases the discriminative ability. Some descriptors tried to utilize the wider local neighborhood at the cost of increased dimension. The directional information is crucial to increase the discriminative power. The existing descriptors use filters to create the directional gradient image which increases the complexity of the descriptor. In order to overcome the above issues, this paper proposes a local directional relation pattern (LDRP). The LDRP first encodes the relationship among directional neighbors and then utilizes the encoded values with center pixel value to generate the final pattern. The major contributions are as follows:
\begin{itemize}
\item The proposed descriptor utilizes the wider local neighborhood without increasing the dimension.
\item In contrast to the existing descriptors which use derivatives to represent the directions, the LDRP descriptor uses the direction inherently. Basically, it encodes the relationship among the directional neighbors at multiple radius to transform it into a single value.
\item The proposed descriptor enriches the pattern with the relationship among local directional neighbors at multiple radius as well as the relationship between the center and transformed directional local neighboring values.
\item The relation among directional neighbors at multiple radius is computed by considering the binary relation between each pair in that direction.
\item The binary relation provides the robustness against uniform illumination while the wider local neighborhood increases the discriminative ability.
\end{itemize}

The rest of the paper is structured as follows: Section II describes the proposed descriptor; Section III illustrates the experimental setup; Section IV reports the experimental results and comparison; Section V presents the performance analysis; And finally Section VI sets the concluding remarks.

\section{Proposed Descriptor}
In this section, the construction process of proposed local directional relation pattern is described in detail. The whole process is divided into several steps such as local neighborhood extraction, local directional information coding, local directional relation pattern generation, feature vector computation and multiscale adaptation.

\subsection{Local Neighborhood Extraction}
\begin{figure}[!t]
    \centering
    \includegraphics[width=.7\linewidth]{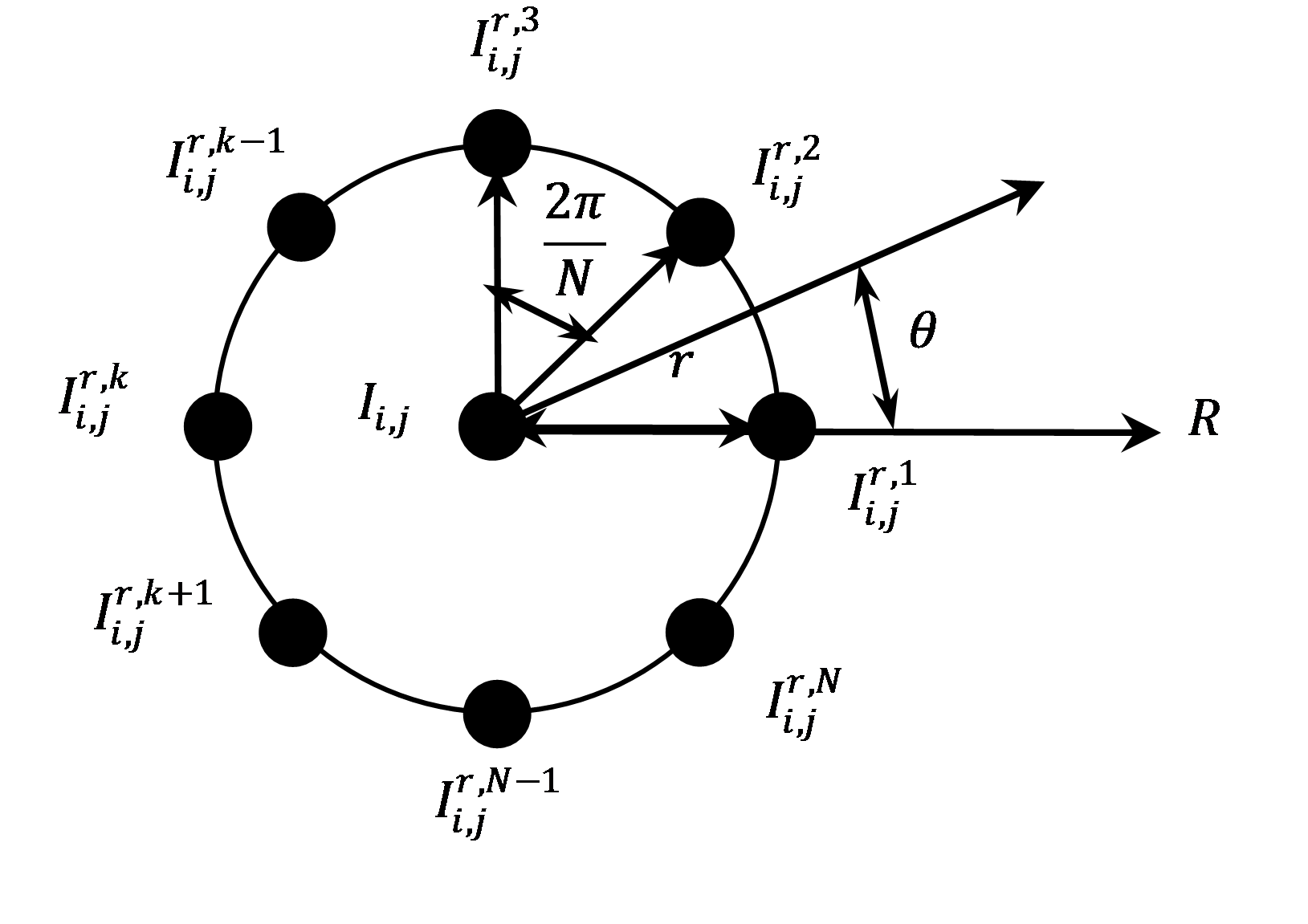}
    \caption{The $N$ neighbors in the local neighborhood of pixel $(i,j)$ at radius $r$. Note that the first neighbor is considered in the right side and rest neighbors are considered w.r.t. first neighbor in counter-clockwise direction.}
    \label{fig:neighbors}
\end{figure}

Let, $I$ is an image with dimension $x \times y$ and $I_{i,j}$ represents the intensity value for the pixel in $i^{th}$ row and $j^{th}$ column with $i \in [1,x]$ and $j \in [1,y]$. The coordinates of top and left corner is $(0,0)$ with positive x-axis downside across the rows and positive y-axis right side across the columns. The $N$ local neighbors of $I_{i,j}$ at a radius $r$ are represented by $I_{i,j}^{r}$, where $I_{i,j}^{r,k}$ is the $k^{th}$ neighbor with $k \in [1,N]$ as shown in Fig. \ref{fig:neighbors}. The coordinates of $k^{th}$ neighbor of pixel $(i,j)$ at a radius $r$ is given by $(i_k,j_k)$, and defined as follows,
\begin{equation}
i_k = i + r\cos\theta_k
\end{equation}
\begin{equation}
j_k = j - r\sin\theta_k
\end{equation}
where $\theta_k$ is the angular displacement of $k^{th}$ neighbor w.r.t. first neighbor and given as follows,
\begin{equation}
\theta_k = (k-1) \times \frac{360}{N} 
\end{equation}
So, $I_{i,j}^{r,k}$ can be written as follows,
\begin{equation}
I_{i,j}^{r,k} = I_{i_k, j_k}
\end{equation}
The first neighbor is considered in the right side of the center pixel and rest of the neighbors are computed w.r.t. first neighbor in the counter-clockwise direction (see Fig. \ref{fig:neighbors}). For example, if the number of local neighbors ($N$) is $8$, then the values of $\theta_k$ are $0$, $45$, $90$, $135$, $180$, $225$, $270$, $315$, and $360$ for $1^{st}$ (i.e., $k=1$), $2^{nd}$ (i.e., $k=2$), $3^{rd}$ (i.e., $k=3$), $4^{th}$ (i.e., $k=4$), $5^{th}$ (i.e., $k=5$), $6^{th}$ (i.e., $k=6$), $7^{th}$ (i.e., $k=7$), and $8^{th}$ (i.e., $k=8$) neighbors, respectively.

\subsection{Local Directional Information Coding}
In order to increase the discriminative ability of the proposed descriptor, the wider neighborhood is used in this work. The relation among local neighbors at multiple radius are utilized to encode the directional information. The $\theta_k$ represents the $k^{th}$ direction with $k\in[1,N]$. Considering the $M$ directional neighbors in $k^{th}$ direction, the binary codes are computed between each pair. The number of pairs out of $M$ neighbors are $\binom{M}{2}$. Let, the $t^{th}$ directional neighboring pair in $k^{th}$ direction is represented by ($I_{i,j}^{r_1,k}, I_{i,j}^{r_2,k}$). The index for $t^{th}$ pair can be computed from $r_1$ and $r_2$ as follows,
\begin{equation}
t = 
\begin{cases}
r_2-r_1, &\text{if $r_1=1$;}\\
r_2-r_1+\sum_{\eta=1}^{r_1-1}(m-\eta), &\text{otherwise.}
\end{cases}
\end{equation}
where $t\in[1,\binom{M}{2}]$, $r_1\in[1,M-1]$ and $r_2\in[r_1+1,M]$.

Let, $\beta_{i,j}^{k}$ denotes the local directional binary pattern for center pixel $(i,j)$ in $k^{th}$ direction. The binary code between $t^{th}$ directional neighboring pair (or between the two neighbors at radius $r_1$ and $r_2$) in $k^{th}$ direction for center pixel $(i,j)$ is generated as follows,
\begin{equation}
\beta_{i,j}^{k}(t) = 
\begin{cases}
1, &\text{if $I_{i,j}^{r_1,k} \leq I_{i,j}^{r_2,k}$;}\\
0, &\text{otherwise.}
\end{cases}
\end{equation}

\begin{figure*}[!t]
    \centering
    \includegraphics[width=.99\linewidth]{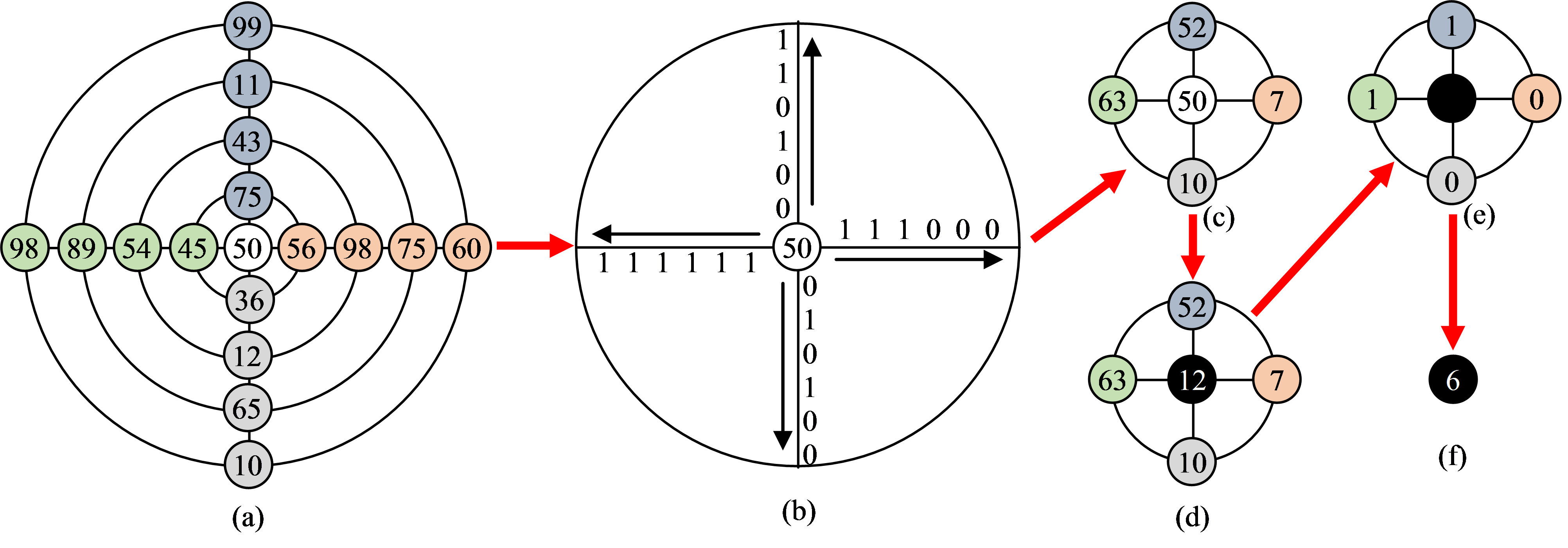}
    \caption{An illustration of the local directional relation pattern computation. (a) A local neighborhood with $N=4$ directions and $M=4$ neighbors in each direction. (b) Local direction binary bits are generated in each direction. For $M=4$, the number of binary values is $\mu=6$. (c) Local directional relation code, $\Gamma_{i,j}^{k}$, is computed in each direction for $k \in [1, 4]$ by converting $\mu=6$ binary bits into equivalent decimal. (d) The transformed center value ($\tau_{i,j}=12$) is computed from the original center pixel value ($I_{i,j}=50$) in order to match its range with local directional codes. (e) The local directional relation binary values (i.e., $\rho_{i,j}^{N,M}(k)$ for $N=4$, $M=4$ and $k \in [1,N]$) are computed for each direction. (f) Finally, the local directional relation pattern for a center pixel ($i,j$), $LDRP_{i,j}^{4,4}$, is generated from $\rho_{i,j}^{4,4}$ by converting the binary values into equivalent decimal.}
    \label{fig:example}
\end{figure*}

An example of local neighborhood with $N=4$ number of circular neighbors and $M=4$ number of directional neighbors is considered in Fig. \ref{fig:example}(a). The intensity value of the center pixel is 50 in this example. Since 4 directional neighbors at different radius is considered, $\mu=\binom{4}{2}=6$ number of binary values are generated in each direction as depicted in Fig. \ref{fig:example}(b). The local directional binary bits in $1^{st}$ direction (i.e. $0^o$) are $\beta_{i,j}^{1} = 1,1,1,0,0,0$ for the pairs (56,98), (56,75), (56,60), (98,75), (98,60), (75,60), respectively. Similarly, the local directional binary bits in $2^{nd}$ direction (i.e. $90^o$) is $\beta_{i,j}^{2} = 0,0,1,0,1,1$, in $3^{rd}$ direction (i.e. $180^o$) is $\beta_{i,j}^{3} = 1,1,1,1,1,1$, and in $4^{th}$ direction (i.e. $270^o$) is $\beta_{i,j}^{4} = 0,1,0,1,0,0$. 

For $M$ neighbors in a direction, $\mu=\binom{M}{2}$ number of binary values are generated. In order to reduce the dimension of the descriptor, it is required to code these binary values into a single value. A local directional information code ($\Gamma$) is generated in each direction from the binary values in that direction. The local directional information code, $\Gamma$, in $k^{th}$ direction for pixel $(i,j)$ is computed by the following equation, 
\begin{equation}
\Gamma_{i,j}^{k} = \sum_{\eta=1}^{\mu}(\beta_{i,j}^{k}(\eta) \times \xi(\eta))
\end{equation}
where, $\xi$ is a weight function to convert the directional binary string into the directional information code (i.e. a decimal value). The purpose of this weight, $\xi$ is to find the unique decimal values for each possible binary string in a particular direction. The weight value ($\xi(\eta)$) for $\eta^{th}$ bit of the directional binary string is defined as follows,
\begin{equation}
\xi(\eta)=2^{\eta-1}.
\label{weight}
\end{equation}
 
The local directional relation codes in the example of Fig. \ref{fig:example}(a) are computed in Fig. \ref{fig:example}(c). The weights for $\mu=6$ directional neighbors are $\xi=1,2,4,8,16,32$. The directional relation code in $1^{st}$ direction is $\Gamma_{i,j}^{1}=1\times1+1\times2+1\times4+0\times8+0\times16+0\times32=1+2+4+0+0+0=7$. Similarly, the directional relation codes in $2^{nd}$, $3^{rd}$, and $4^{th}$ directions are $\Gamma_{i,j}^{2}=52$, $\Gamma_{i,j}^{3}=63$, and $\Gamma_{i,j}^{4}=10$ as shown in Fig. \ref{fig:example}(c).

\subsection{Local Directional Relation Pattern}
The local directional relation code is computed in the previous sub-section for a direction by encoding the relationship among the neighbors at different radius in that direction. Now, the next step is to find out the relation between center pixel and local directional relation codes. The minimum and maximum values of local directional code are dependent upon the number of directional neighbors considered (i.e. $M$). The code is generated from the $\mu = \binom{M}{2}$ number of binary values. So, the different number of decimal values that can be generated from $\mu$ binary bits is $2^{\mu}$ with a minimum value as $0$ and maximum value as $2^{\mu}-1$. Whereas, the minimum and maximum values of center pixel are $0$ and $2^{B}-1$ respectively, where $B$ is the bit-depth of the image. Note that, the bit-depth ($B$) of the images is 8 in the databases used in this paper. A clear mismatch can be observed between the range of center pixel and local directional relation code. Thus, a transformation is required over either the center pixel or the local directional relation codes to match both the ranges. Due to efficiency reason, the center pixel is transformed into the range of local directional relation codes as follows,
\begin{equation}
\tau_{i,j} = \Upsilon (I_{i,j} \times \frac{2^{\mu}-1}{2^{B}-1}).
\end{equation}
where, $\tau_{i,j}$ is the transformed version of $I_{i,j}$, $\Upsilon(\eta)=\lfloor \eta \rfloor$ is a function to compute the floor of $\eta$ to the closest integer value equal to or less than $\eta$.
The transformed value of center pixel in Fig. \ref{fig:example}(d) for $\mu=6$ and $B=8$ is computed as $\tau_{i,j}=\Upsilon(50 \times \frac{2^6-1}{2^8-1})=\lfloor 12.35 \rfloor = 12$.

Let, $\rho$ is a binary pattern representing the relationship between center and directional relation code having $N$ values corresponding to each direction. The $\rho_{i,j}^{N,M}(k)$ for center pixel $(i,j)$ in $k^{th}$ direction is given as follows,
\begin{equation}
\rho_{i,j}^{N,M}(k)=
\begin{cases}
1,	&\text{if $\Delta_{i,j}^{k} \geq 0$;}\\
0,	&\text{otherwise.}
\end{cases}
\end{equation}
where, $\Delta_{i,j}^{k}$ is the difference between local directional relation code in $k^{th}$ direction and transformed value of the center pixel, i.e.,
\begin{equation}
\Delta_{i,j}^{k} = \Gamma_{i,j}^{k} - \tau_{i,j} 
\end{equation}

The local directional relation pattern $(LDRP)$ for pixel $(i, j)$ by considering the local neighbors in $N$ directions with $M$ neighbors in each direction is computed as follows,
\begin{equation}
LDRP_{i,j}^{N,M} = \sum_{k=1}^{N}(\rho_{i,j}^{N,M}(k)) \times \xi(k)),
\end{equation}
where, $\xi$ is a weight function defined in (\ref{weight}).

\subsection{LDRP Feature Vector}
The LDRP feature vector ($H$) is generated by finding the number of occurrences of LDRP values over the whole image. Note that, the minimum and maximum values of LDRP are $0$ and $2^N-1$ respectively. Thus, the length of feature vector is $2^N$. The LDRP feature vector for Image $I$ with local neighborhood from $N$ directions having $M$ neighbors in each direction is defined as follows,
\begin{equation}
H^{N,M}(\eta) = \sum_{i=M+1}^{x-M}\sum_{j=M+1}^{y-M} \zeta(LDRP_{i,j}^{N,M}, \eta)
\end{equation}
where, $\zeta$ is calculated by following rule,
\begin{equation}
\zeta(\alpha_1, \alpha_2) = 
\begin{cases}
1,	&\text{if $\alpha_1 = \alpha_2$;}\\
0,	&\text{otherwise.}
\end{cases}
\end{equation}
Note that, in this paper the number of directions (i.e., $N$) is considered as $8$, so the dimension of the LDRP feature vector is $2^8=256$ at radius $M$ (i.e. for $M$ directional neighbors).

\subsection{Multi-scale LDRP}

In order to make the LDRP descriptor more discriminative, the multi-scale directional neighborhood characteristics are utilized in this work. The LDRP feature descriptors are computed by varying the number of directional neighbor (i.e. $M$) in each direction. The values of $M$ are considered from $M_1$ to $M_2$ with $M_2 \geq M_1$. Finally, the LDRP feature vectors ($H^{N,M}$ for $M \in [M_1, M_2]$) are concatenated into a single feature vector. Mathematically, the final LDRP feature vector ($H^{N,M_1,M_2}$) can be written as follows,
\begin{equation}
\begin{split}
H^{N,M_1,M_2} & = H^{N,M_1}  ||  H^{N,M_1+1}  ||  ...  ||  H^{N,M_2} \\
& = {||}_{\eta=M_1}^{M_2} H^{N,\eta}
\end{split}
\end{equation}
The dimension of the multi-scale LDRP feature vector only depends upon the number of directions ($N$) and number of scales ($M_2 - M_1 + 1$) and given as follows,
\begin{equation}
d^{N,M_1,M_2}=(M_2 - M_1 + 1) \times 2^N. 
\end{equation}
In order to make the final feature vector invariant to the image resolution, $H^{N,M_1,M_2}$ is normalized as follows,
\begin{equation}
H_{Normalized}^{N,M_1,M_2}(\delta) = \frac{H^{N,M_1,M_2}(\delta)}{\sum_{\lambda = 1}^{d^{N,M_1,M_2}}H^{N,M_1,M_2}(\lambda)}
\end{equation}
for $\forall \delta \in [1, d^{N,M_1,M_2}]$. In the experiments, the normalized version of feature vector is considered for all descriptors.
In case of $M_1=M_2$, the multiscale LDRP feature vector is equivalent to the single scale LDRP feature vector (i.e., $H^{N,M}=H^{N,M_1}=H^{N,M_2}$).

\begin{figure}[!t]
    \centering
    \includegraphics[width=.9\linewidth]{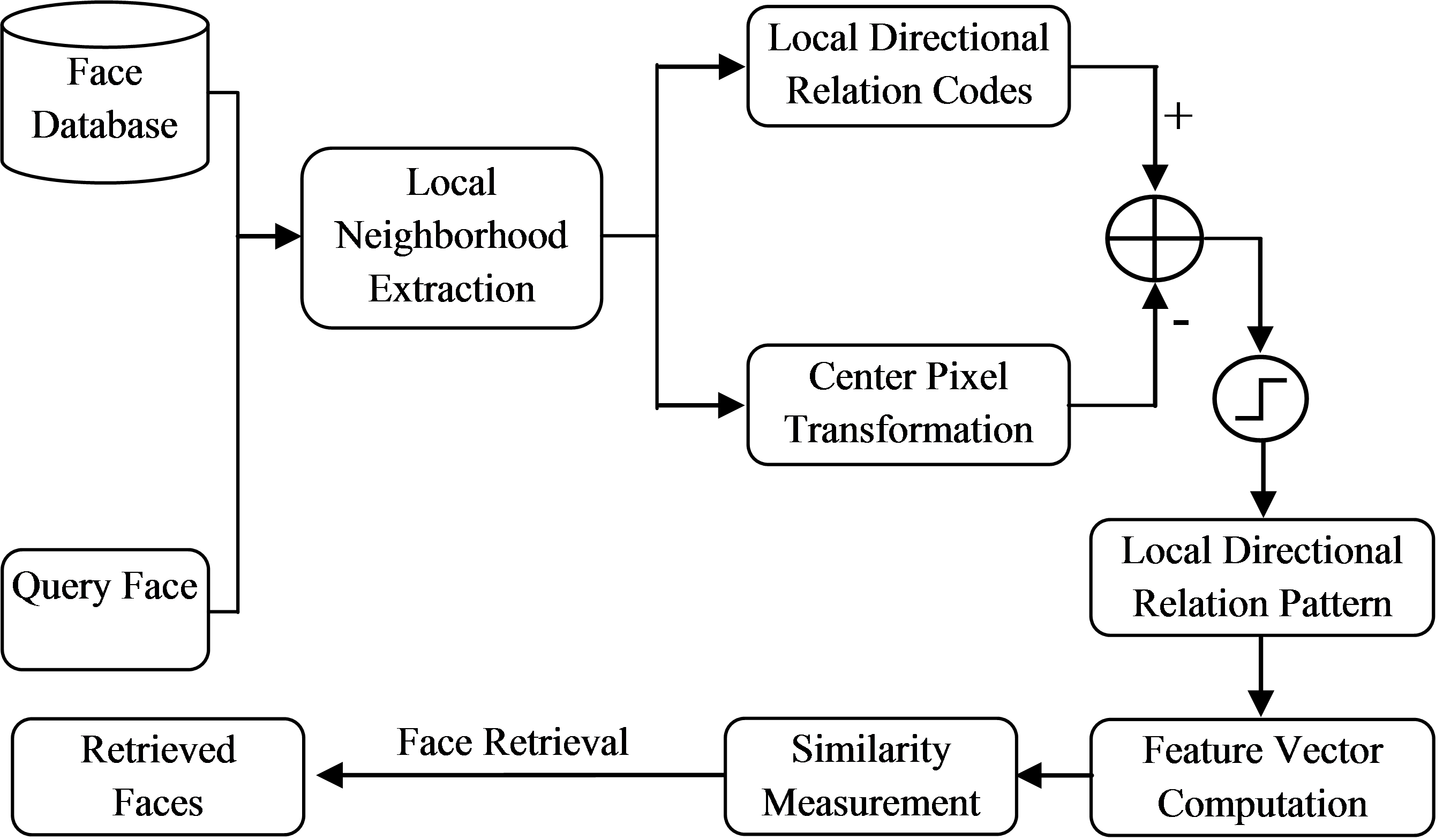}
    \caption{The face retrieval framework using proposed LDRP descriptor.}
    \label{fig:setup}
\end{figure}

\begin{figure*}[!t]
  \begin{subfigure}{.5\textwidth}
    \centering
    \includegraphics[clip=true, trim = 0 0 0.8cm 16cm, width=0.98\columnwidth]{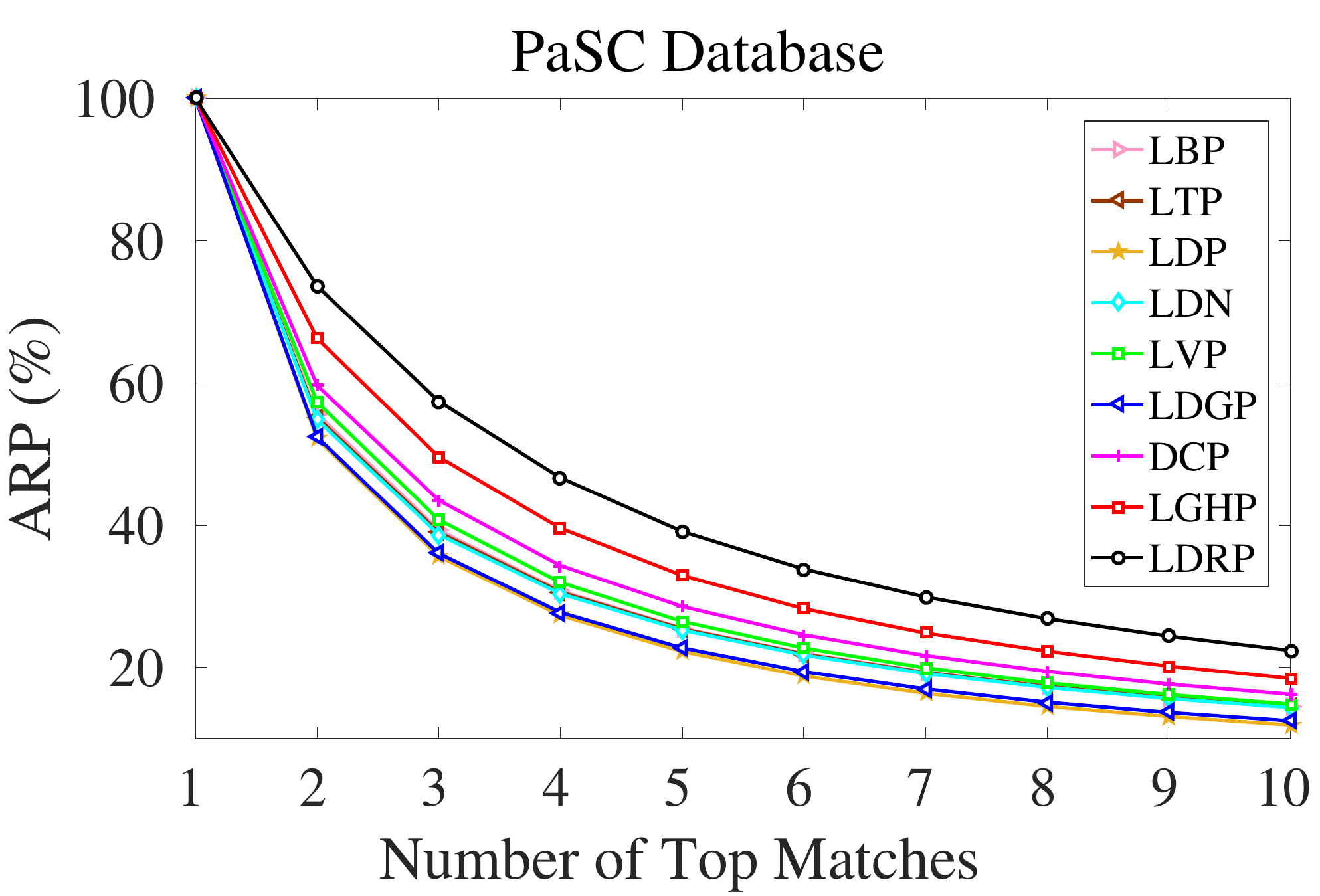}
    \caption{ARP}
    \label{fig:pasc-arp}
  \end{subfigure}%
    \begin{subfigure}{.5\textwidth}
    \centering
    \includegraphics[clip=true, trim = 0 0 0.8cm 16cm, width=0.98\columnwidth]{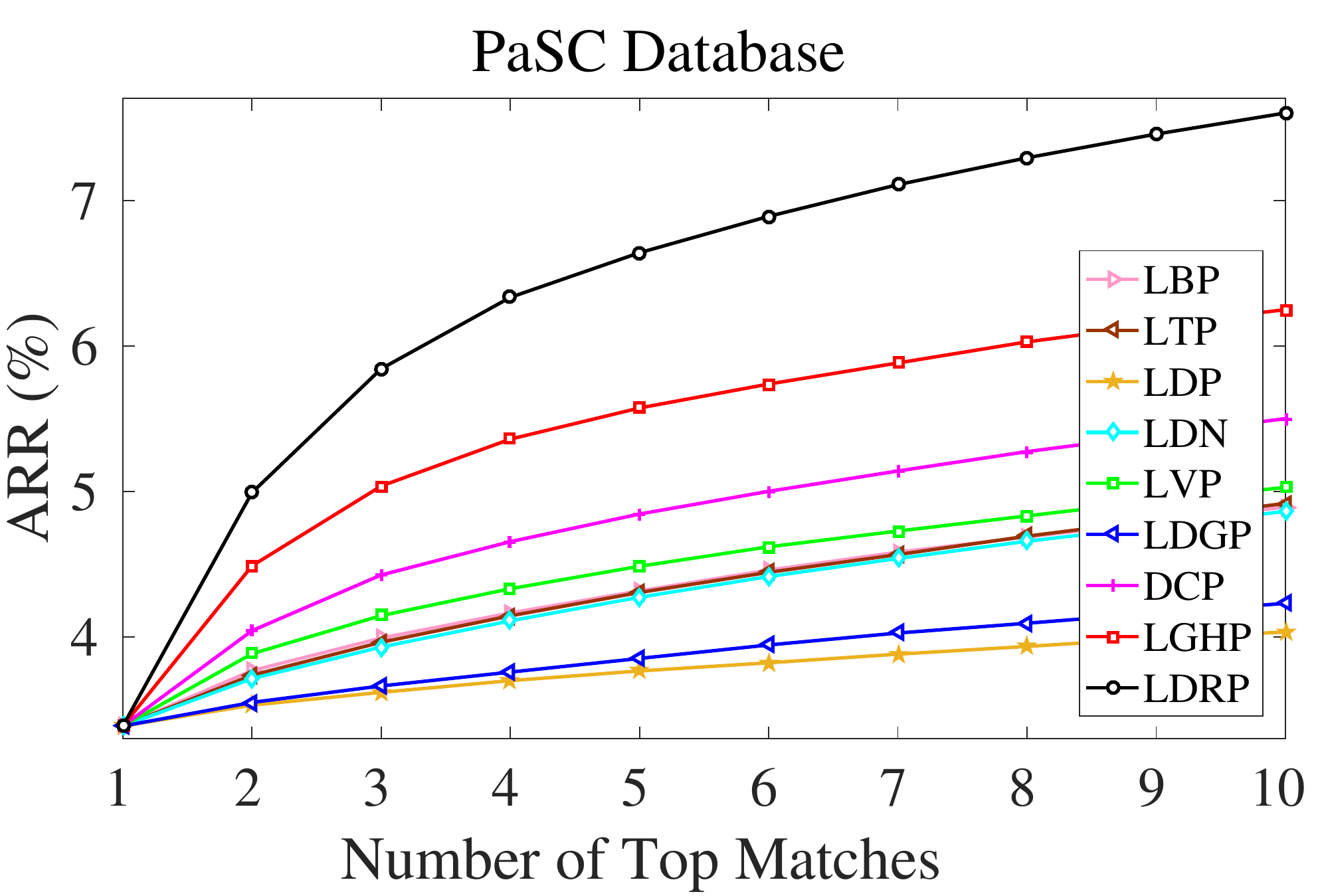}
    \caption{ARR}
    \label{fig:pasc-arr}
  \end{subfigure}
    \begin{subfigure}{.5\textwidth}
    \centering
    \includegraphics[clip=true, trim = 0 0 0.8cm 16cm, width=0.98\columnwidth]{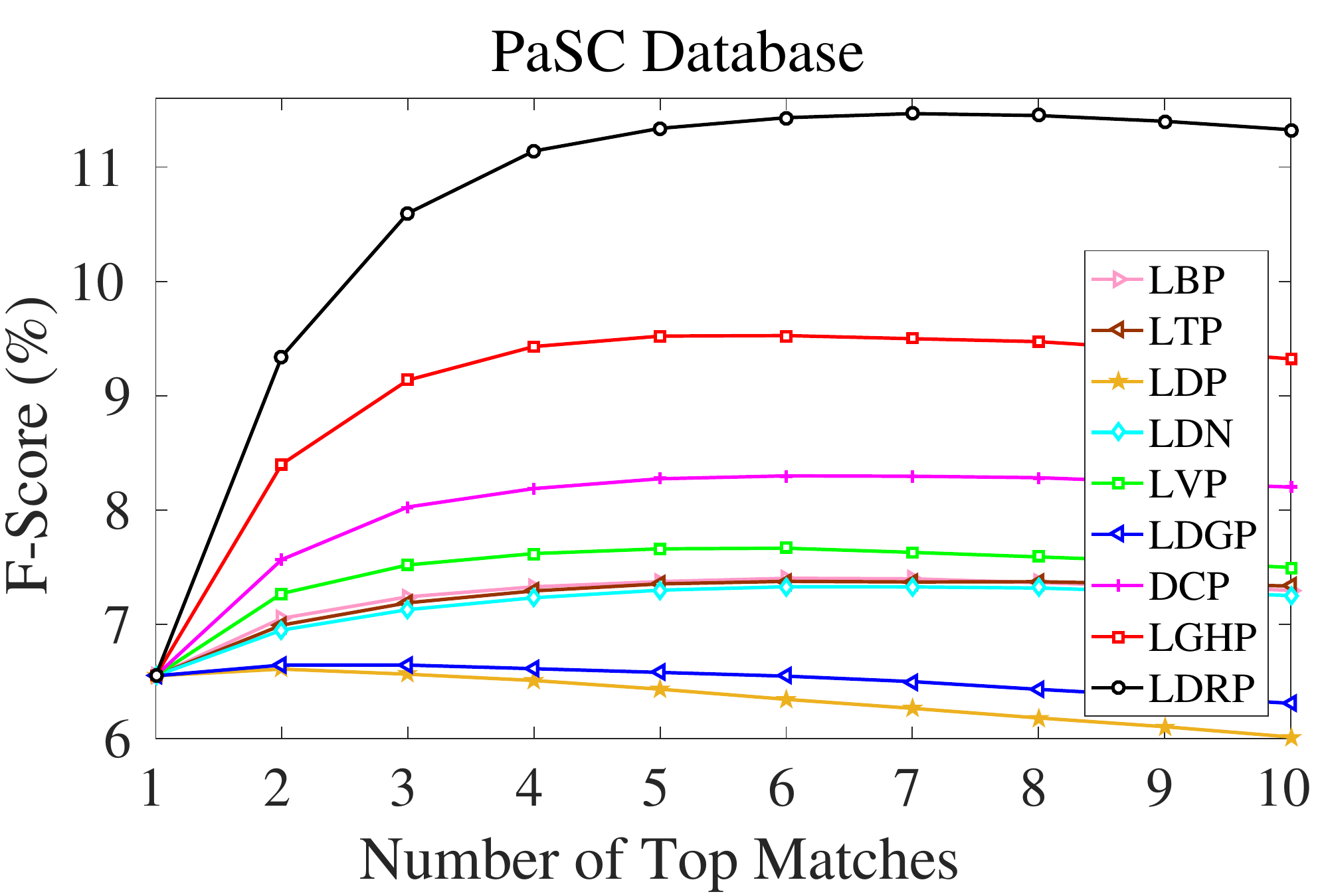}
    \caption{F-Score}
    \label{fig:pasc-f}
  \end{subfigure}%
    \begin{subfigure}{.5\textwidth}
    \centering
    \includegraphics[clip=true, trim = 0 0 0.8cm 16cm, width=.98\linewidth]{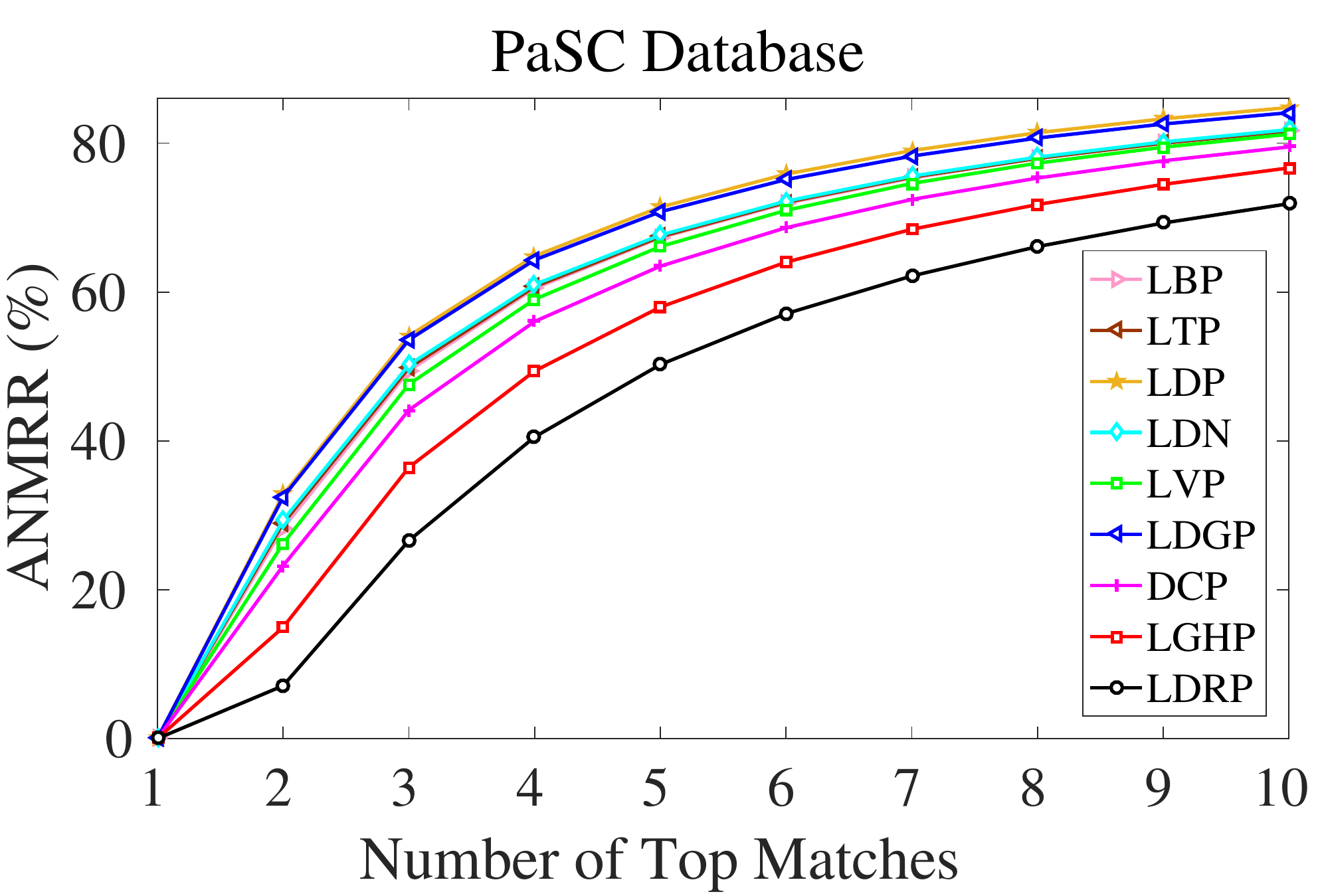}
    \caption{ANMRR}
    \label{fig:pasc-anmrr}
  \end{subfigure}
  \caption{The results over PaSC database in terms of the ARP, ARR, F-Score, and ANMRR vs number of retrieved images.}
  \label{fig:results_pasc}
\end{figure*}

\begin{figure*}[!t]
  \begin{subfigure}{.5\textwidth}
    \centering
    \includegraphics[clip=true, trim = 0 0 0.8cm 16cm, width=.98\linewidth]{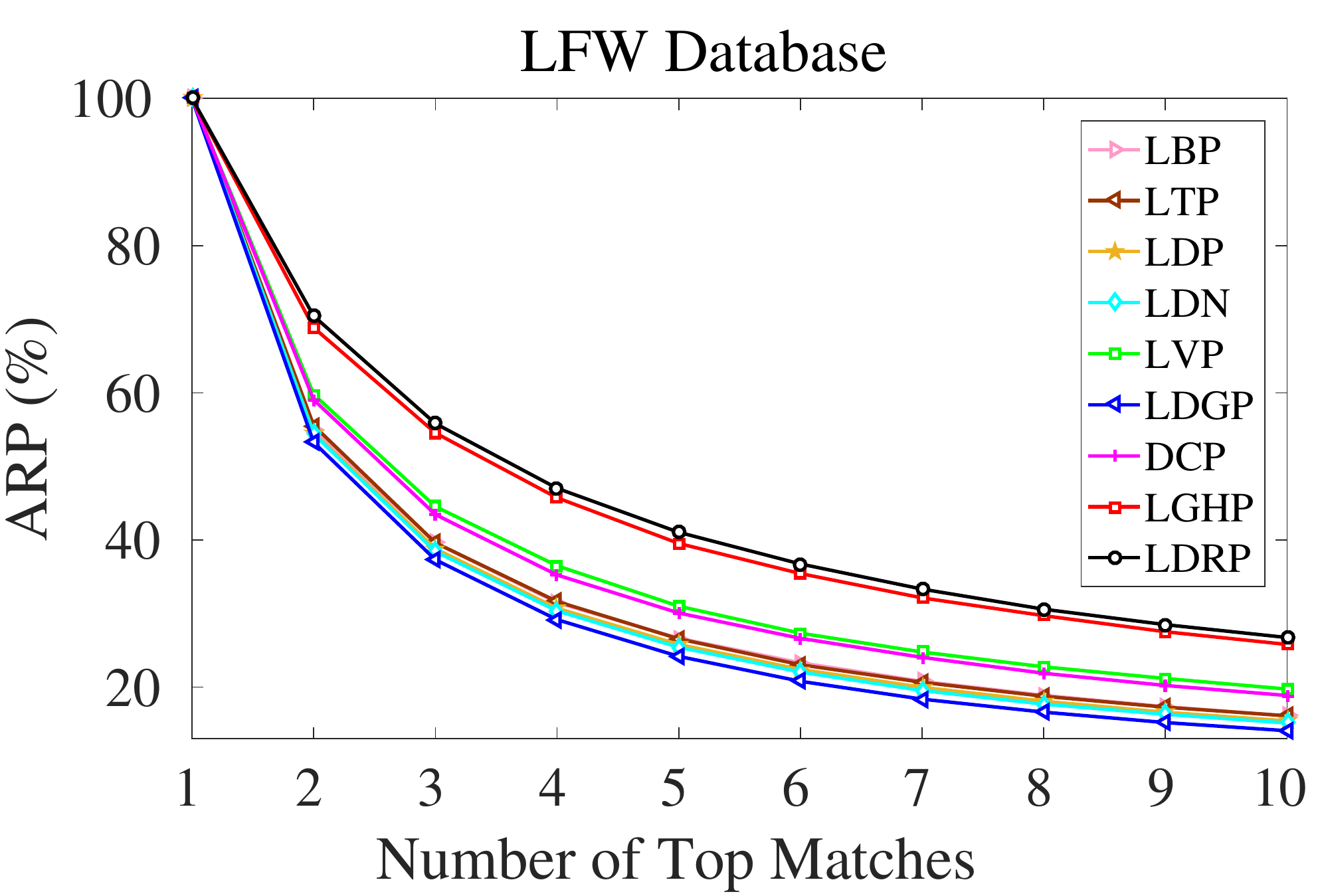}
    \caption{ARP}
    \label{fig:lfw-arp}
  \end{subfigure}%
    \begin{subfigure}{.5\textwidth}
    \centering
    \includegraphics[clip=true, trim = 0 0 0.8cm 16cm, width=.98\linewidth]{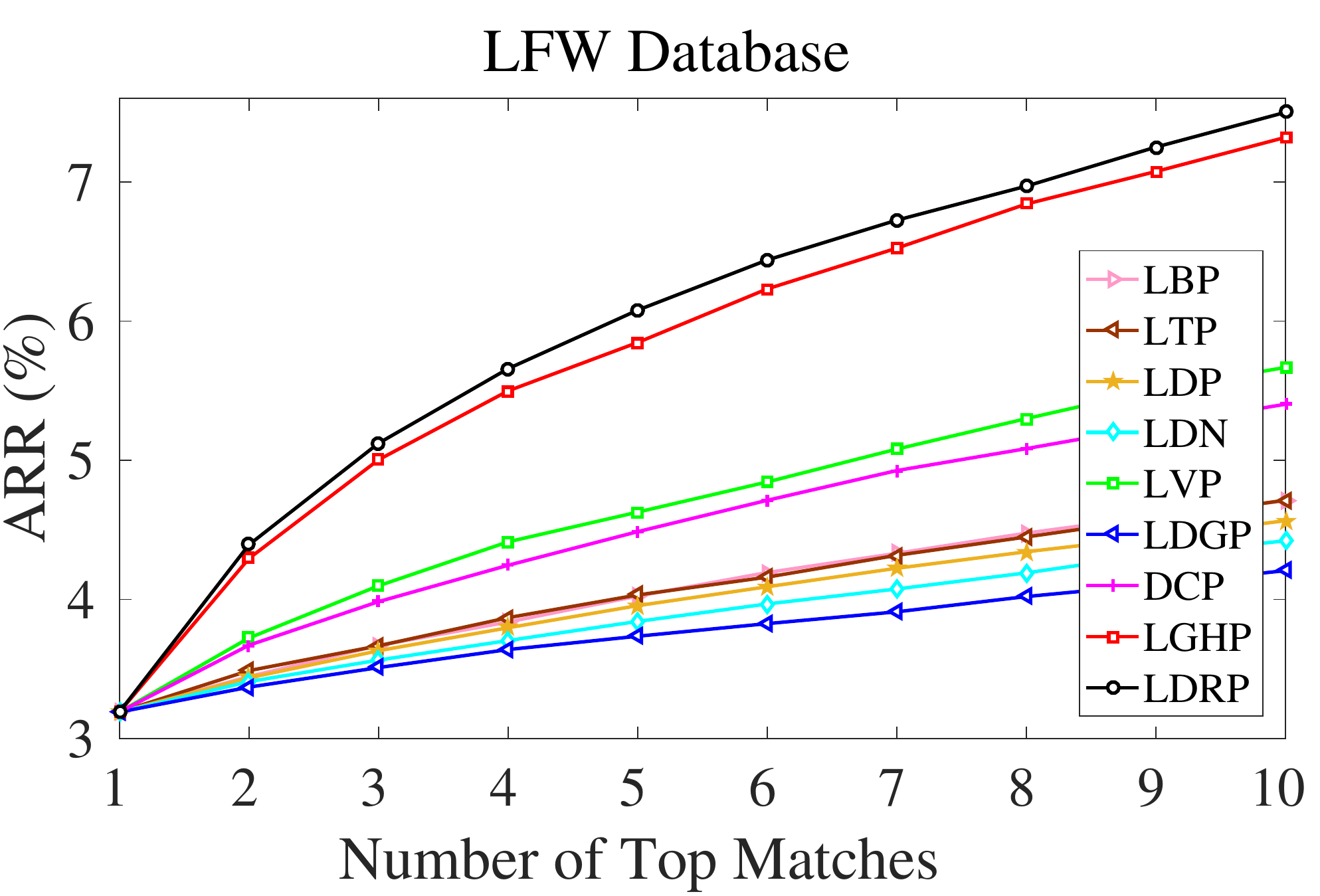}
    \caption{ARR}
    \label{fig:lfw-arr}
  \end{subfigure}
    \begin{subfigure}{.5\textwidth}
    \centering
    \includegraphics[clip=true, trim = 0 0 0.8cm 16cm, width=.98\linewidth]{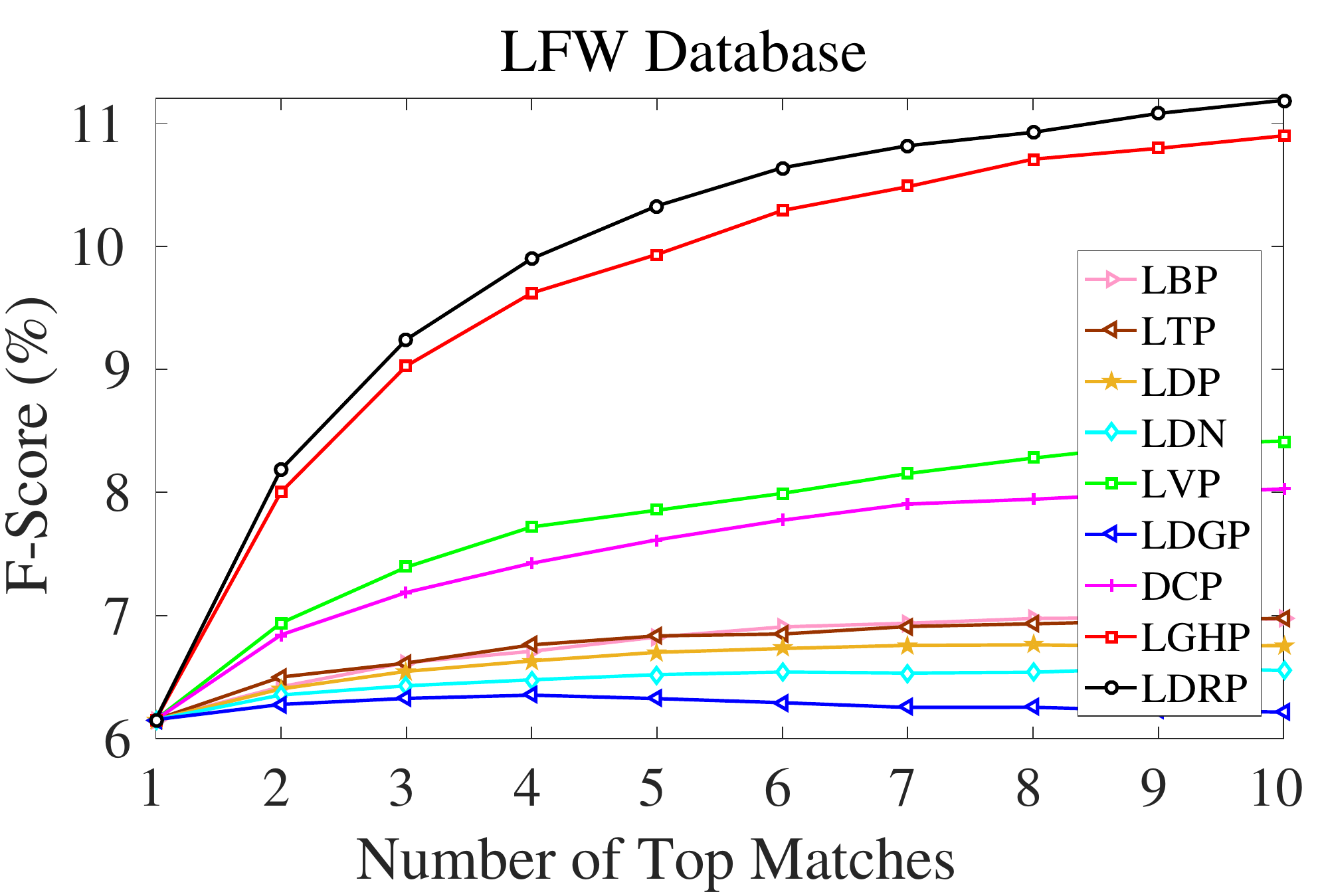}
    \caption{F-Score}
    \label{fig:lfw-f}
  \end{subfigure}%
    \begin{subfigure}{.5\textwidth}
    \centering
    \includegraphics[clip=true, trim = 0 0 0.8cm 16cm, width=.98\linewidth]{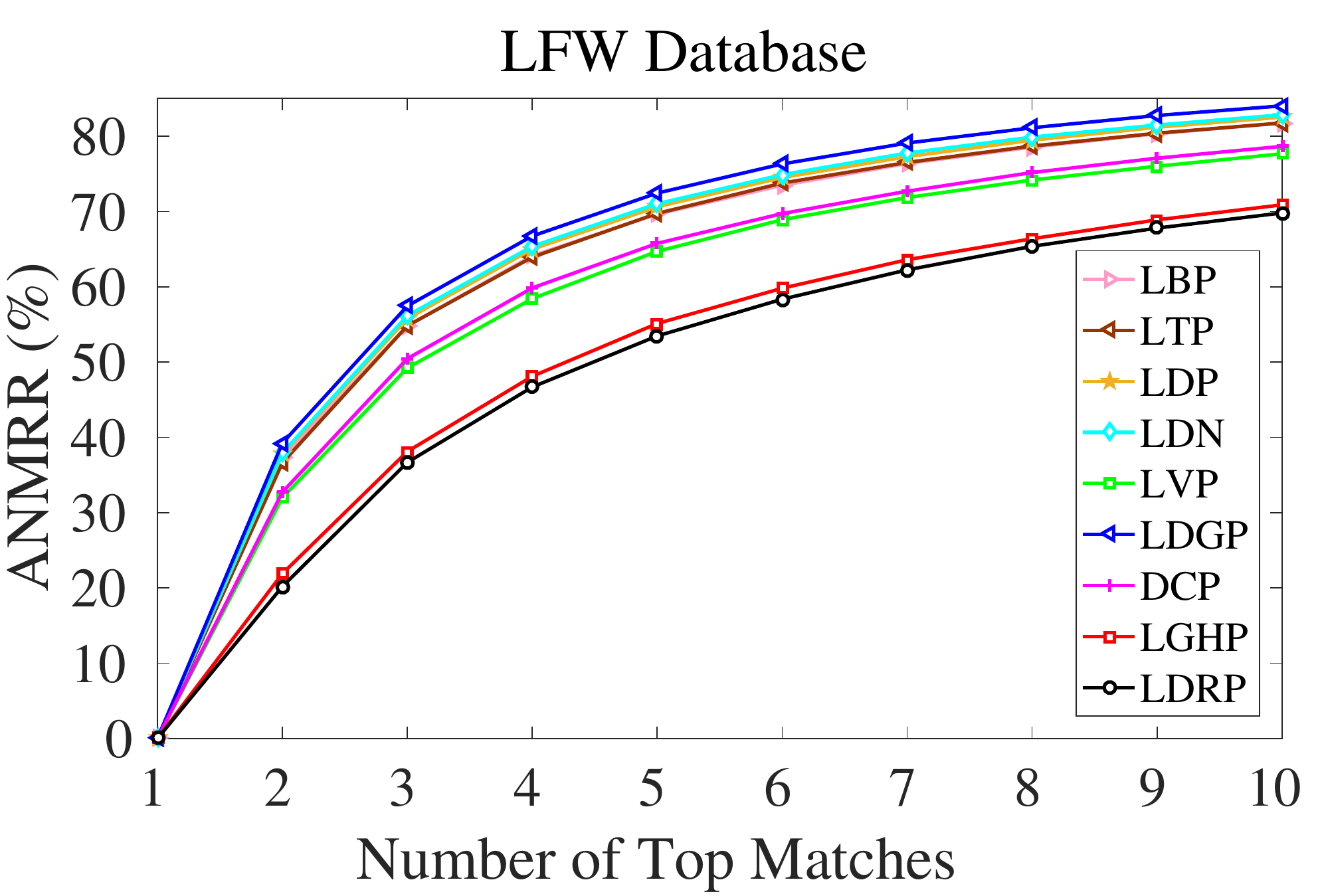}
    \caption{ANMRR}
    \label{fig:lfw-anmrr}
  \end{subfigure}
  \caption{The results over LFW database in terms of the ARP, ARR, F-Score, and ANMRR vs number of retrieved images.}
  \label{fig:results_lfw}
\end{figure*}

\begin{figure*}[!t]
  \begin{subfigure}{.5\textwidth}
    \centering
    \includegraphics[clip=true, trim = 0 0 0.8cm 16cm, width=.98\linewidth]{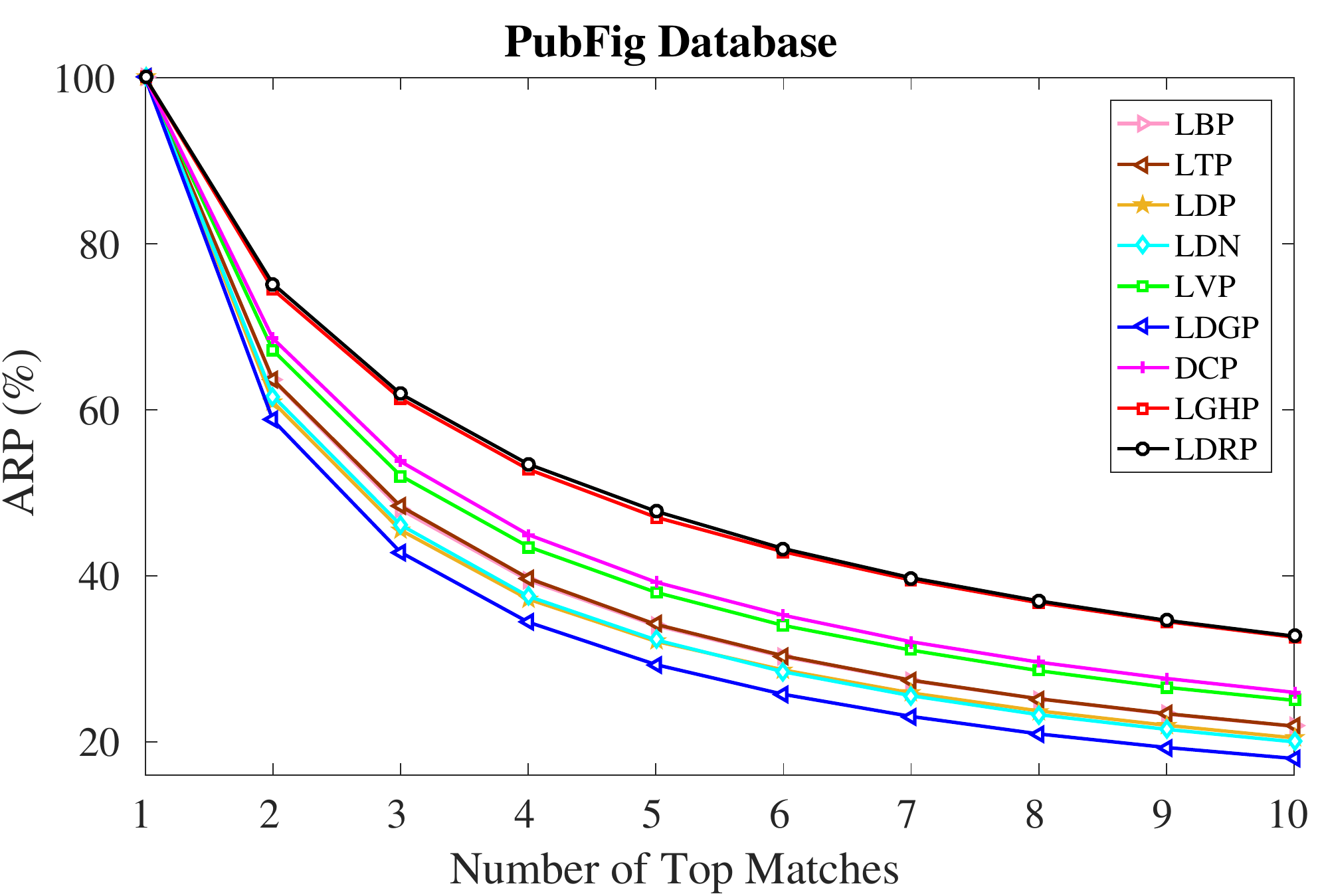}
    \caption{ARP}
    \label{fig:pubfig-arp}
  \end{subfigure}%
    \begin{subfigure}{.5\textwidth}
    \centering
    \includegraphics[clip=true, trim = 0 0 0.8cm 16cm, width=.98\linewidth]{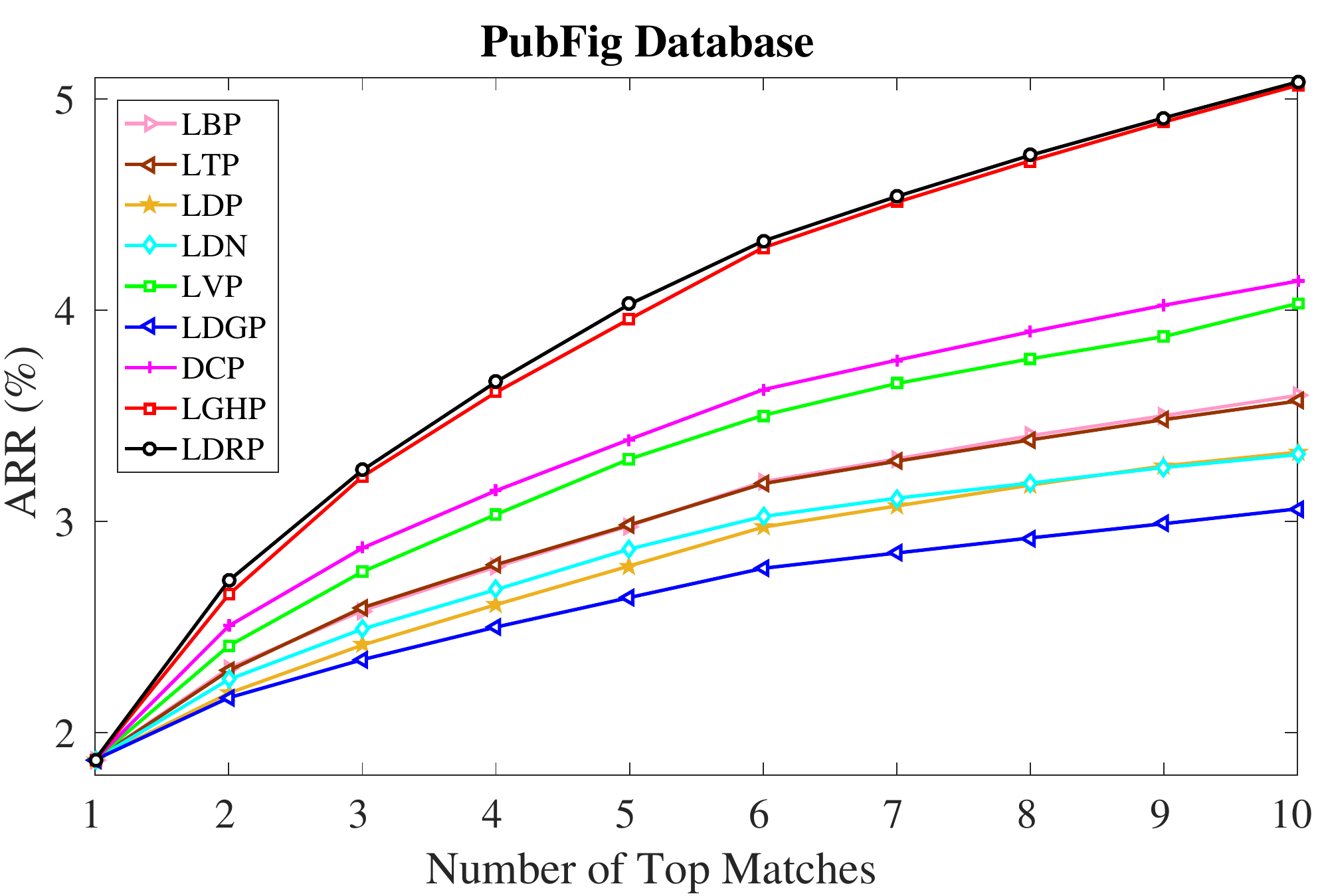}
    \caption{ARR}
    \label{fig:pubfig-arr}
  \end{subfigure}
    \begin{subfigure}{.5\textwidth}
    \centering
    \includegraphics[clip=true, trim = 0 0 0.8cm 16cm, width=.98\linewidth]{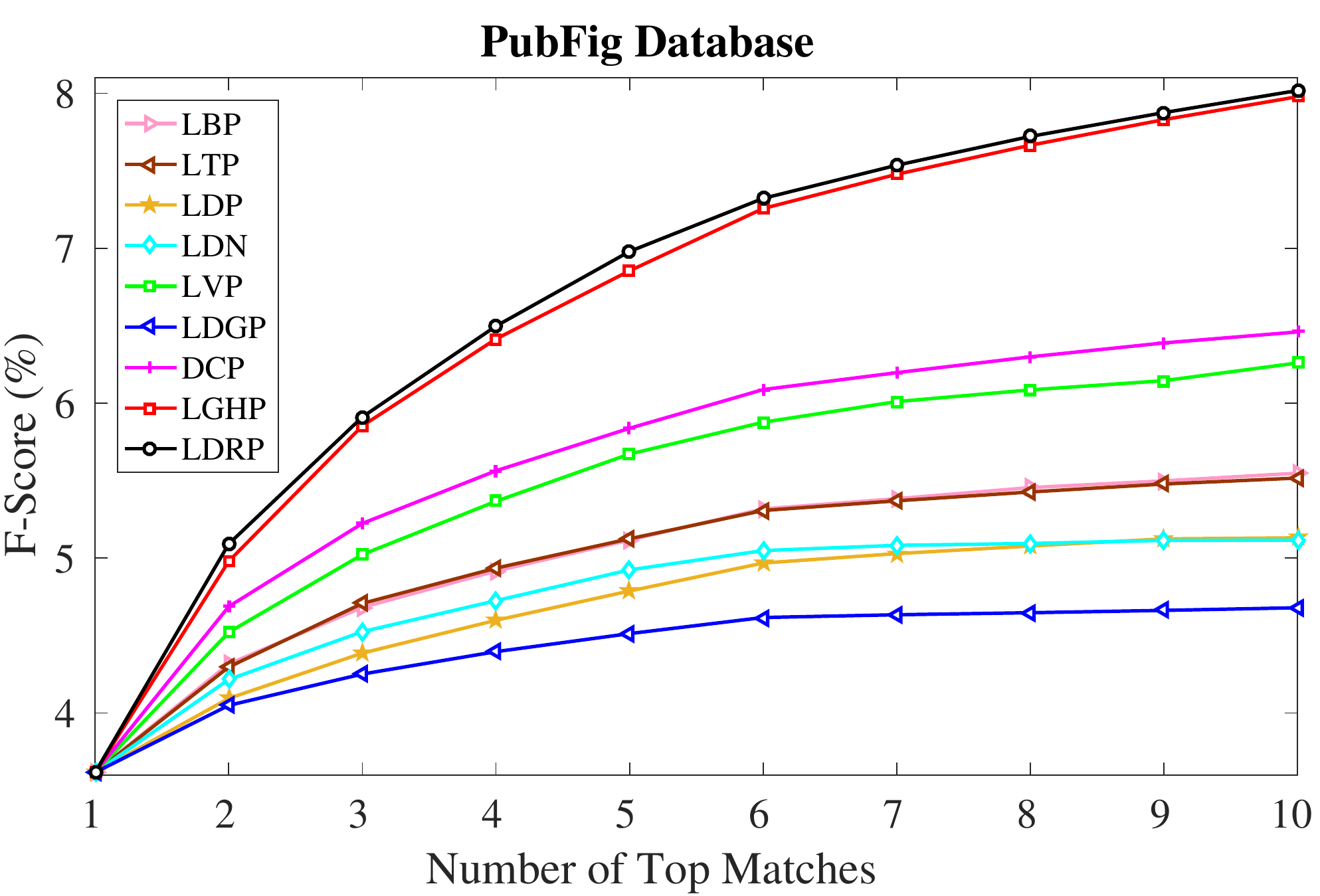}
    \caption{F-Score}
    \label{fig:pubfig-f}
  \end{subfigure}%
    \begin{subfigure}{.5\textwidth}
    \centering
    \includegraphics[clip=true, trim = 0 0 0.8cm 16cm, width=.98\linewidth]{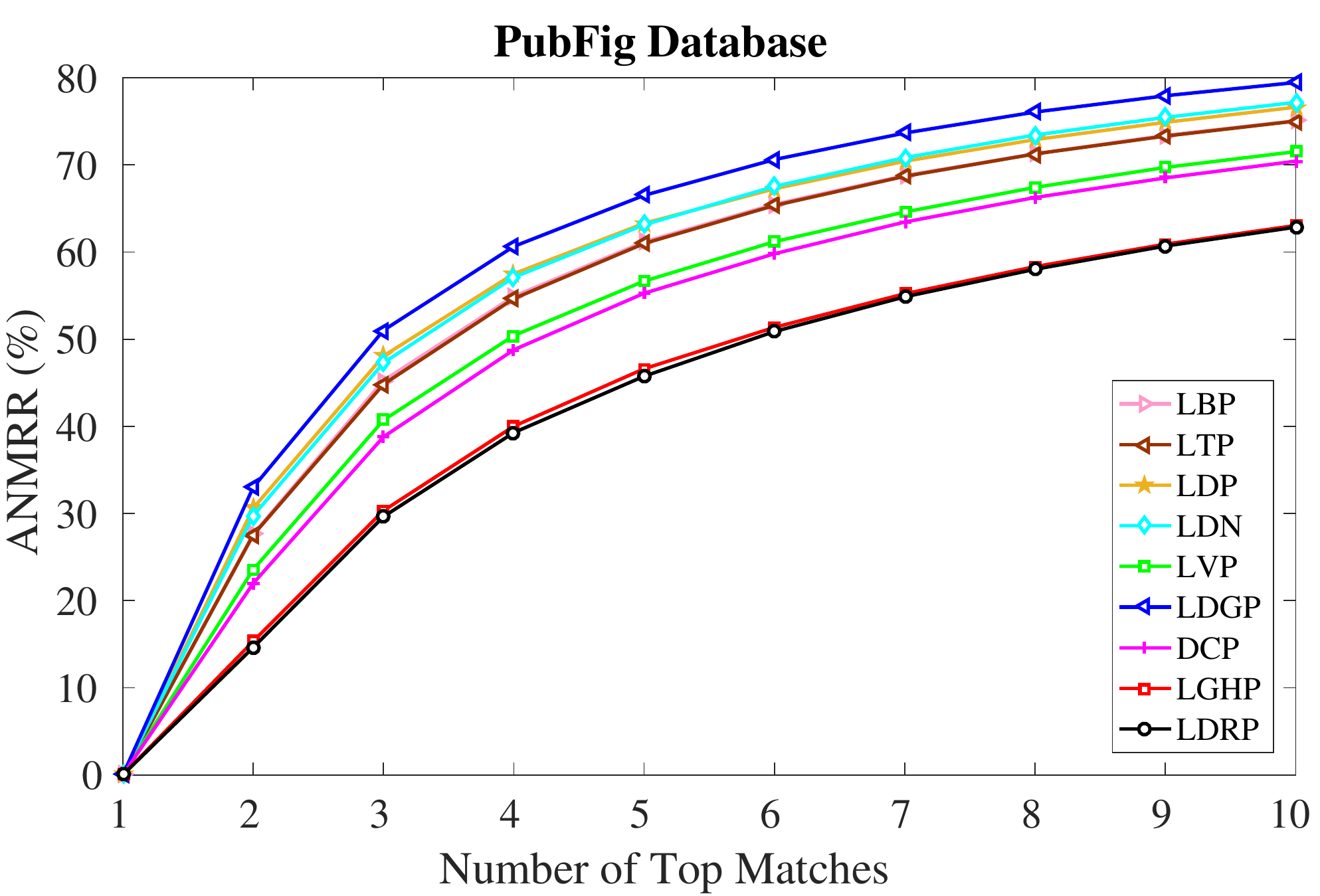}
    \caption{ANMRR}
    \label{fig:pubfig-anmrr}
  \end{subfigure}
  \caption{The results over PubFig database in terms of the ARP, ARR, F-Score, and ANMRR vs number of retrieved images.}
  \label{fig:results_pubfig}
\end{figure*}

\begin{figure*}[!t]
  \begin{subfigure}{.5\textwidth}
    \centering
    \includegraphics[clip=true, trim = 0 0 0.8cm 16cm, width=.98\linewidth]{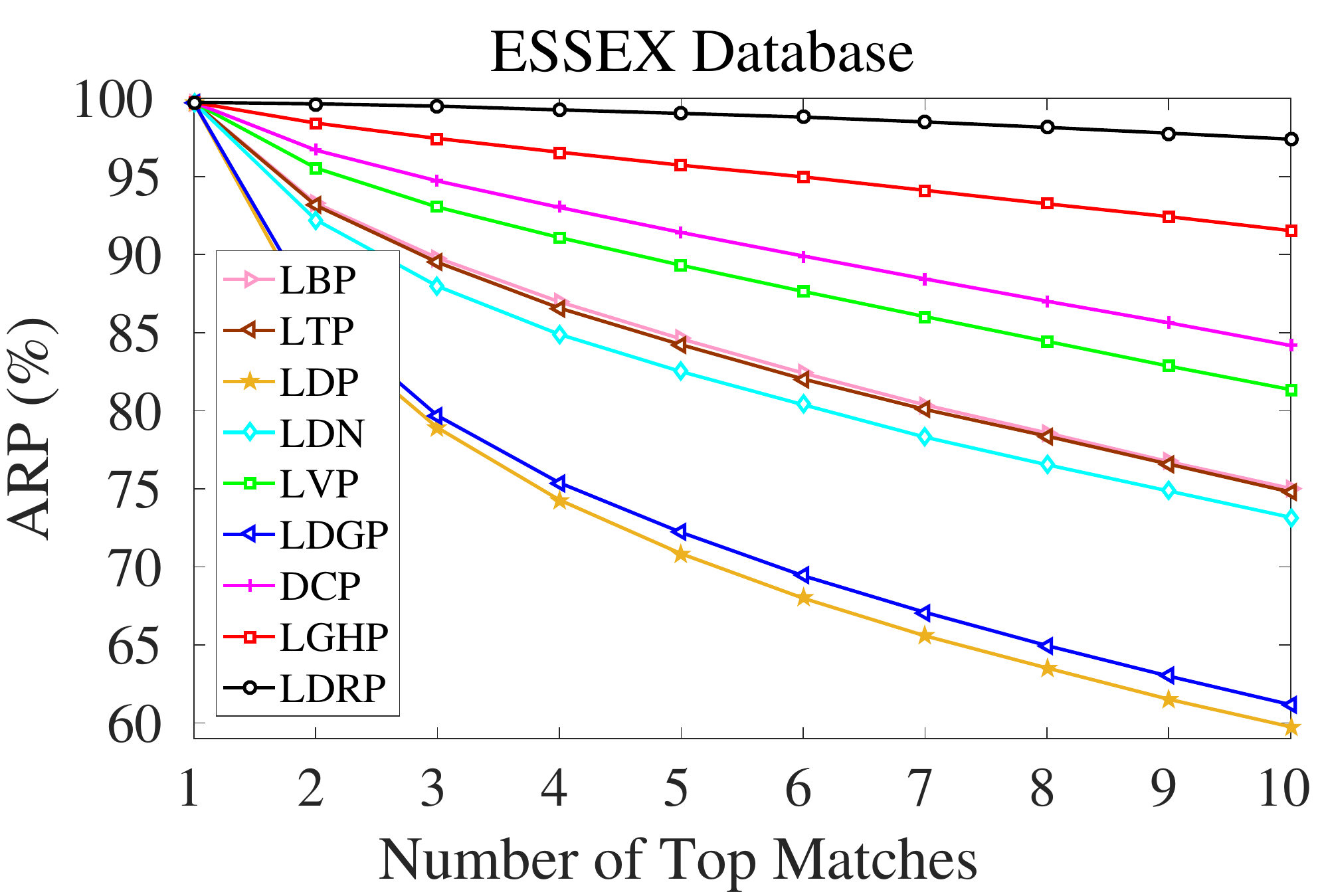}
    \caption{ARP}
    \label{fig:essex-arp}
  \end{subfigure}%
    \begin{subfigure}{.5\textwidth}
    \centering
    \includegraphics[clip=true, trim = 0 0 0.8cm 16cm, width=.98\linewidth]{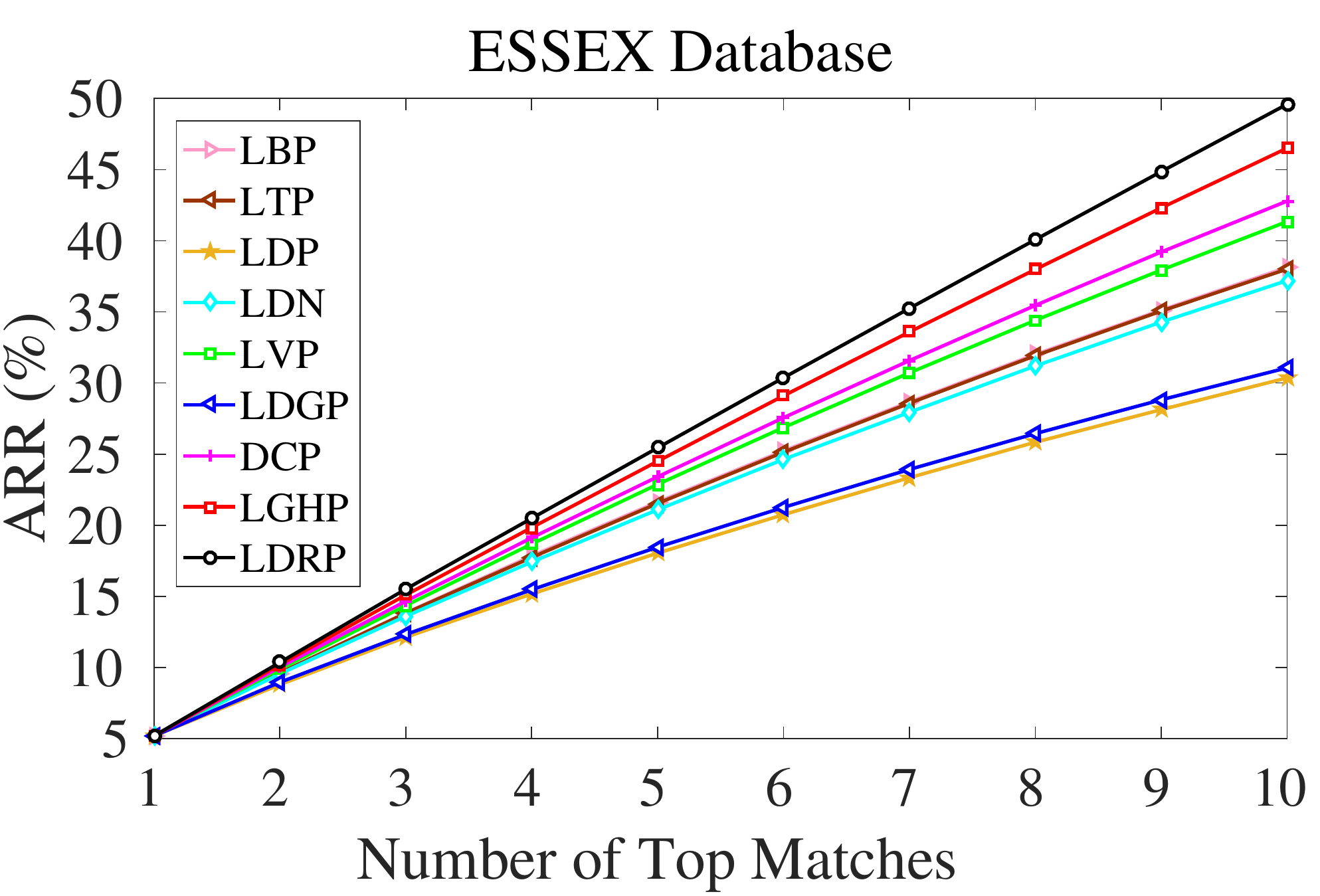}
    \caption{ARR}
    \label{fig:essex-arr}
  \end{subfigure}
    \begin{subfigure}{.5\textwidth}
    \centering
    \includegraphics[clip=true, trim = 0 0 0.8cm 16cm, width=.98\linewidth]{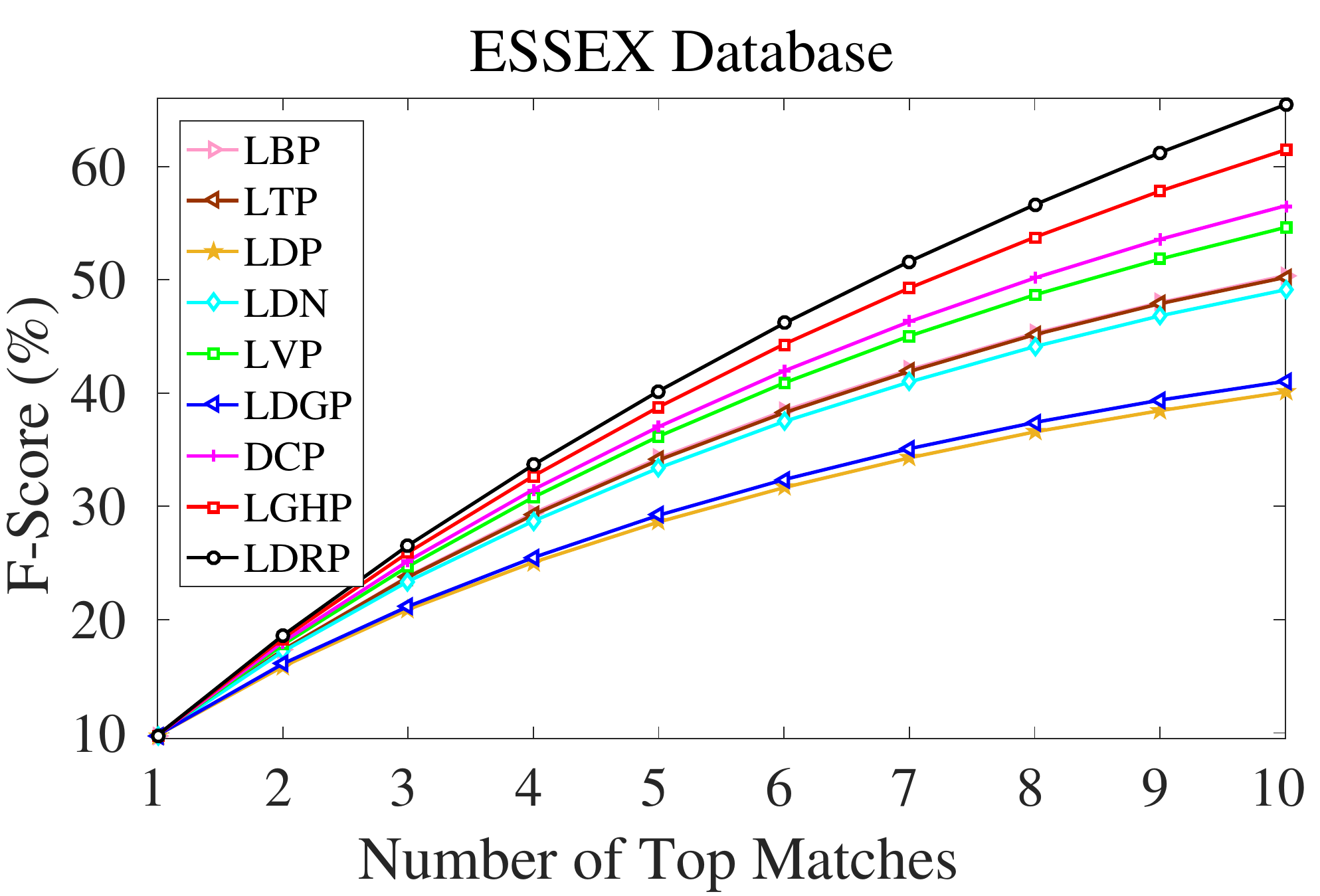}
    \caption{F-Score}
    \label{fig:essex-f}
  \end{subfigure}%
    \begin{subfigure}{.5\textwidth}
    \centering
    \includegraphics[clip=true, trim = 0 0 0.8cm 16cm, width=.98\linewidth]{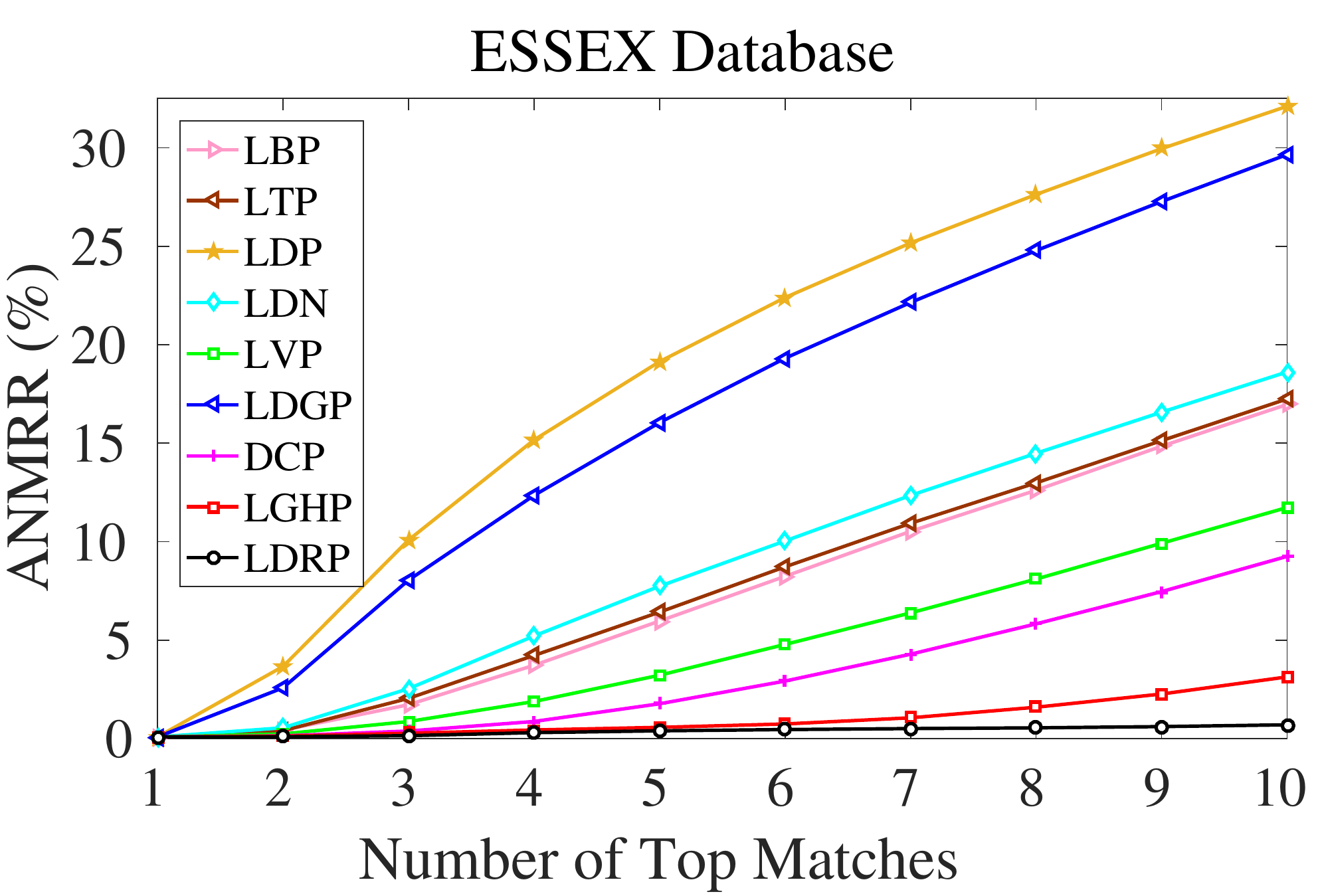}
    \caption{ANMRR}
    \label{fig:essex-anmrr}
  \end{subfigure}
  \caption{The results over ESSEX database in terms of the ARP, ARR, F-Score, and ANMRR vs number of retrieved images.}
  \label{fig:results_essex}
\end{figure*}

\begin{figure*}[!t]
  \begin{subfigure}{.5\textwidth}
    \centering
    \includegraphics[clip=true, trim = 0 0 0.8cm 16cm, width=.98\linewidth]{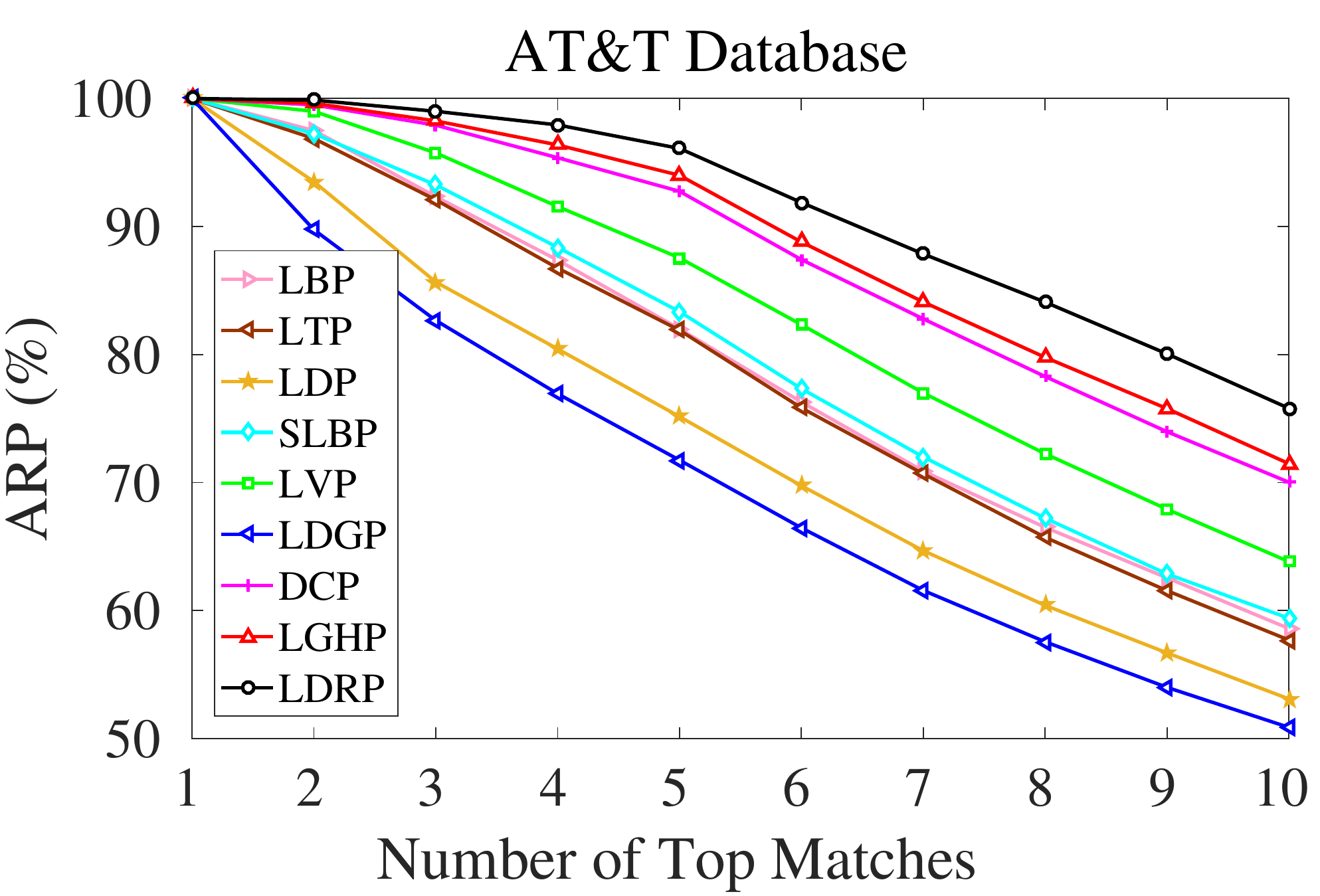}
    \caption{ARP}
    \label{fig:att-arp}
  \end{subfigure}%
    \begin{subfigure}{.5\textwidth}
    \centering
    \includegraphics[clip=true, trim = 0 0 0.8cm 16cm, width=.98\linewidth]{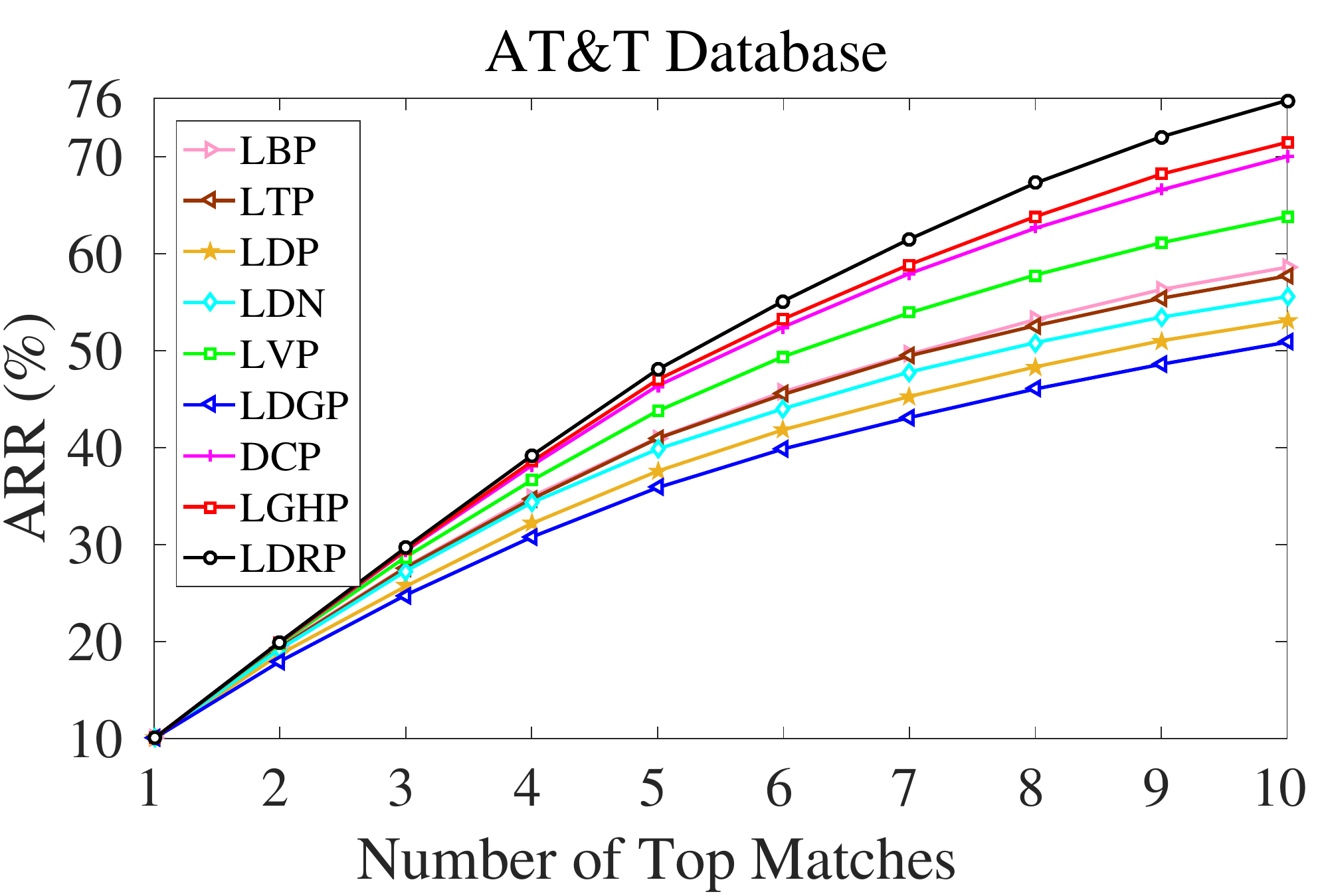}
    \caption{ARR}
    \label{fig:att-arr}
  \end{subfigure}
    \begin{subfigure}{.5\textwidth}
    \centering
    \includegraphics[clip=true, trim = 0 0 0.8cm 16cm, width=.98\linewidth]{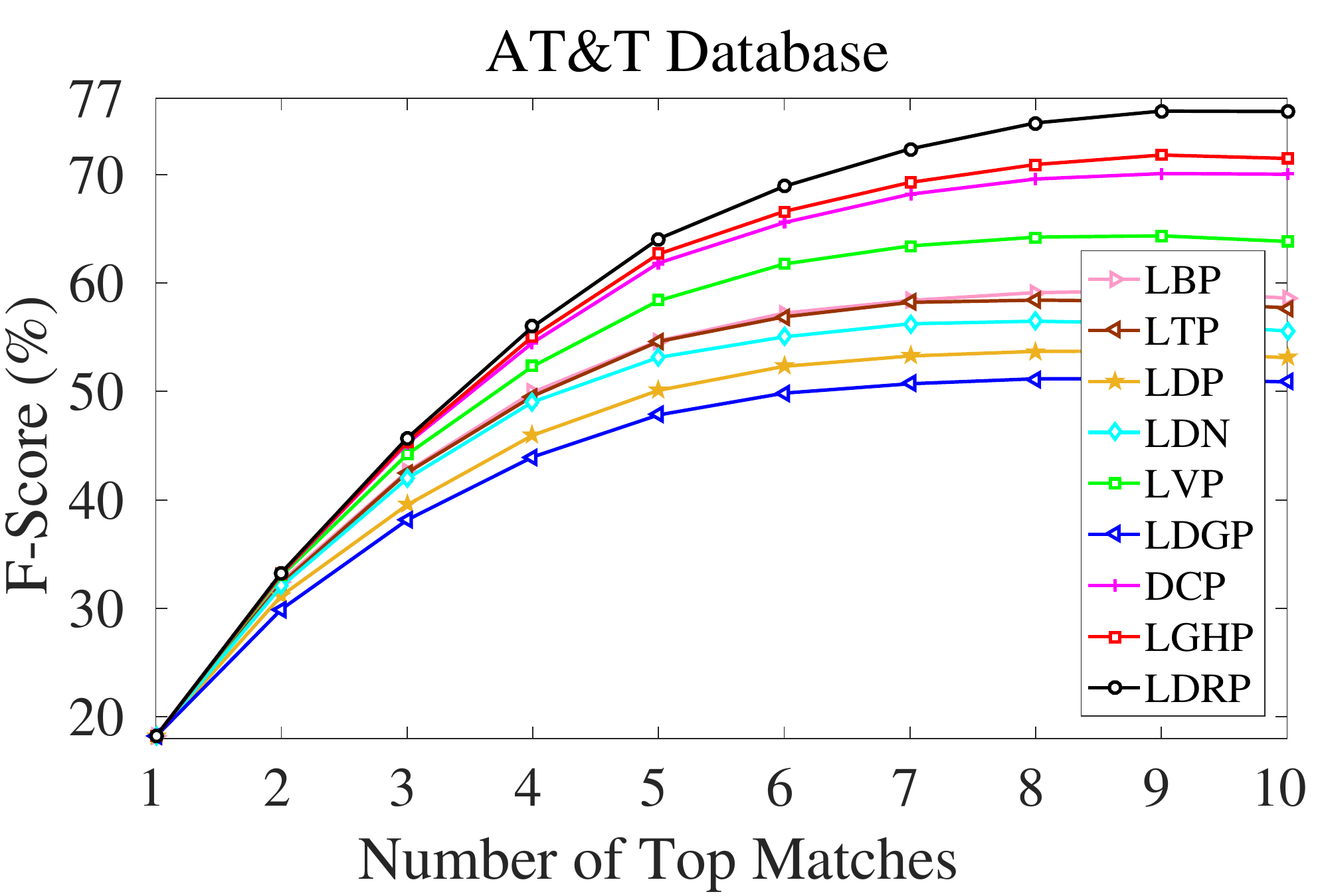}
    \caption{F-Score}
    \label{fig:att-f}
  \end{subfigure}%
    \begin{subfigure}{.5\textwidth}
    \centering
    \includegraphics[clip=true, trim = 0 0 0.8cm 16cm, width=.98\linewidth]{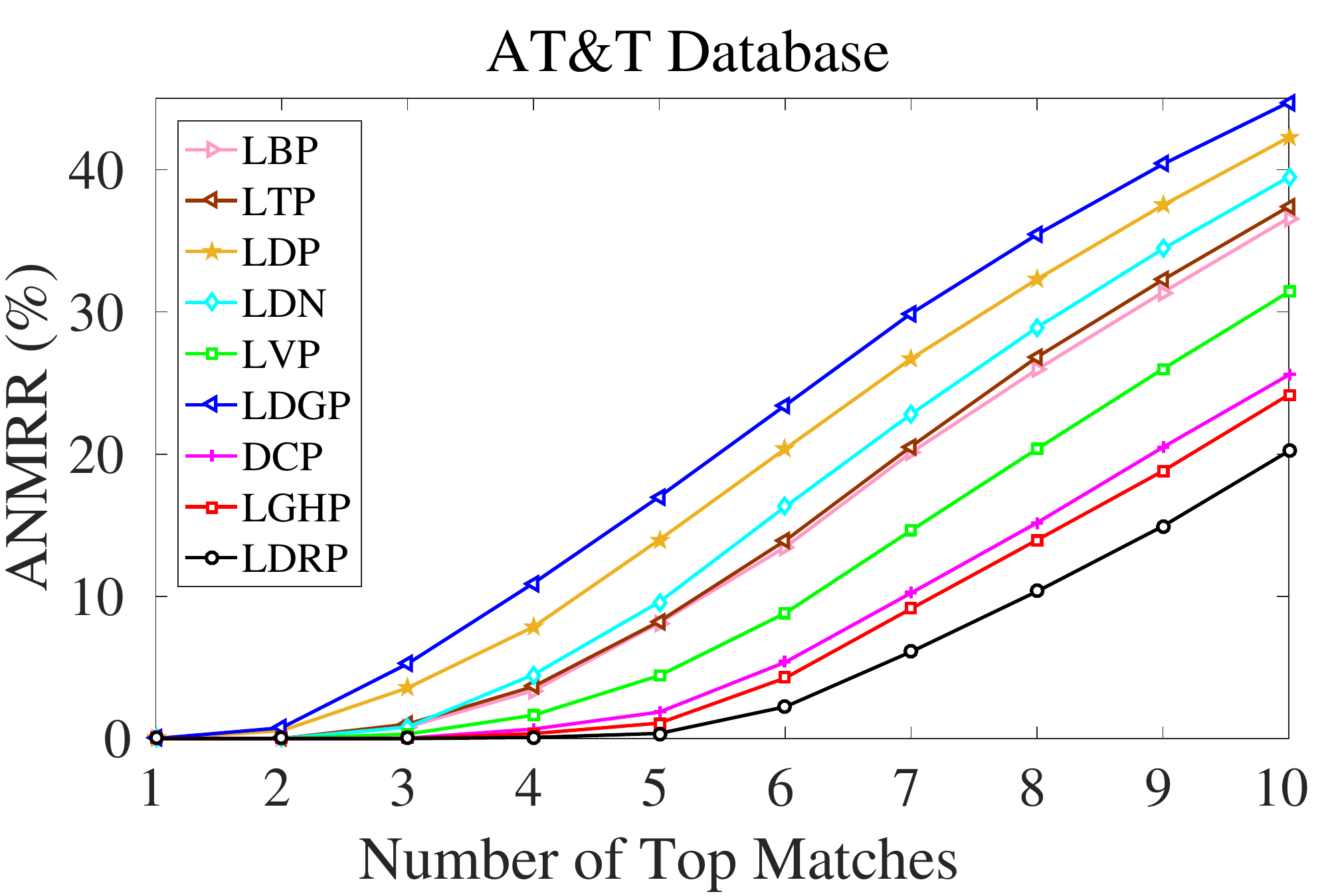}
    \caption{ANMRR}
    \label{fig:att-anmrr}
  \end{subfigure}
  \caption{The results over AT\&T database in terms of the ARP, ARR, F-Score, and ANMRR vs number of retrieved images.}
  \label{fig:results_att}
\end{figure*}

\begin{figure*}[!t]
  \begin{subfigure}{.5\textwidth}
    \centering
    \includegraphics[clip=true, trim = 0 0 0.8cm 16cm, width=.98\linewidth]{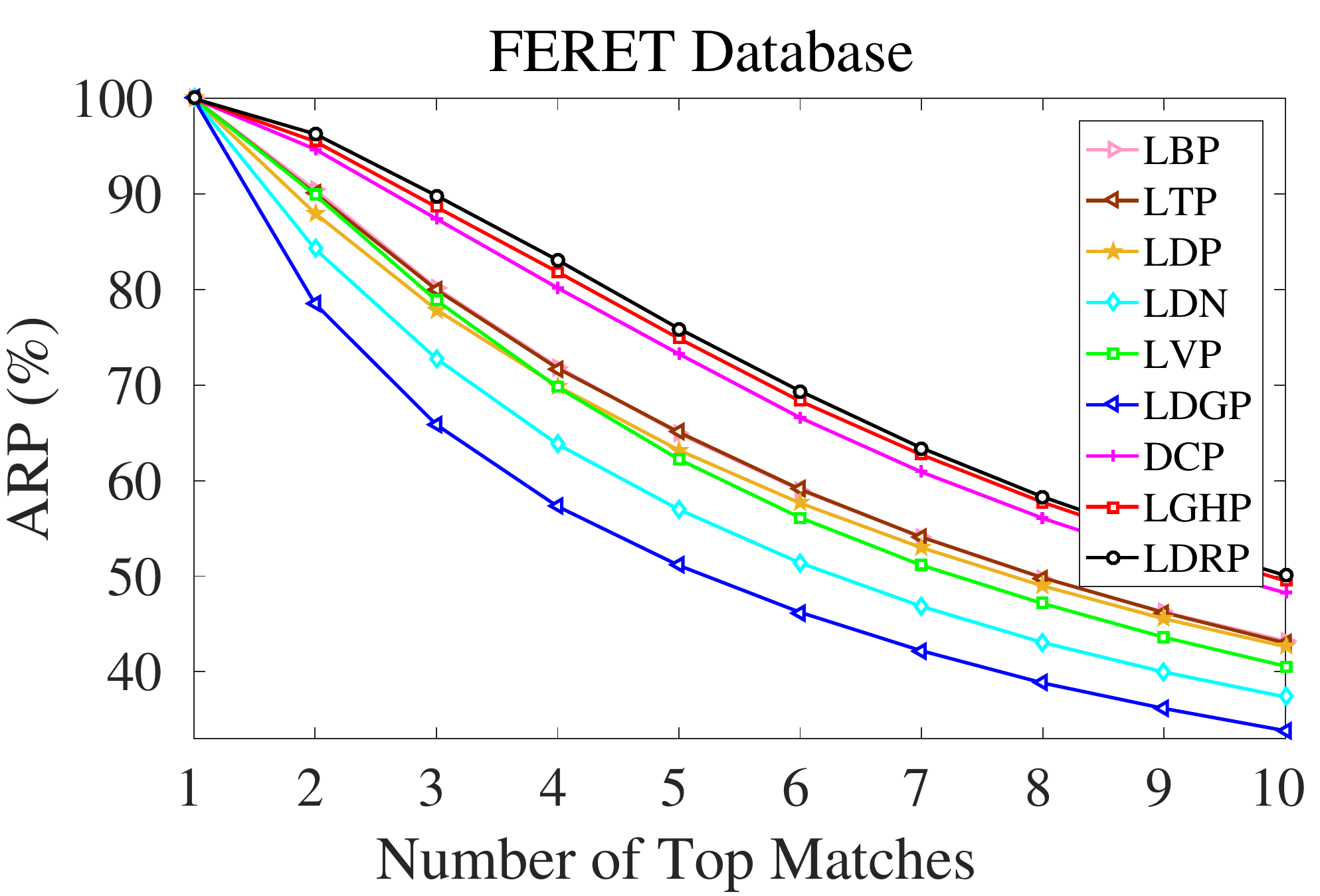}
    \caption{ARP}
    \label{fig:feret-arp}
  \end{subfigure}%
    \begin{subfigure}{.5\textwidth}
    \centering
    \includegraphics[clip=true, trim = 0 0 0.8cm 16cm, width=.98\linewidth]{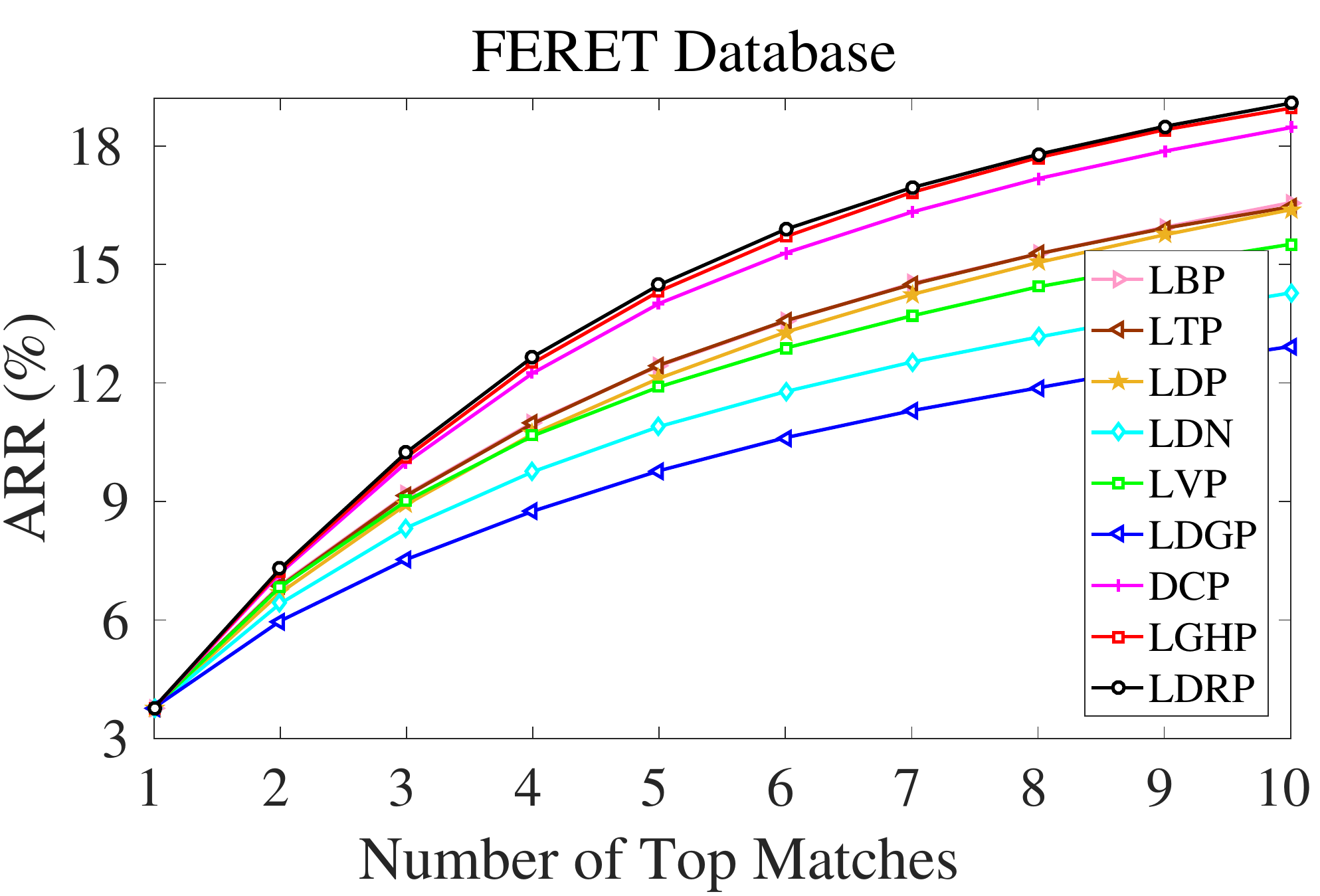}
    \caption{ARR}
    \label{fig:feret-arr}
  \end{subfigure}
    \begin{subfigure}{.5\textwidth}
    \centering
    \includegraphics[clip=true, trim = 0 0 0.8cm 16cm, width=.98\linewidth]{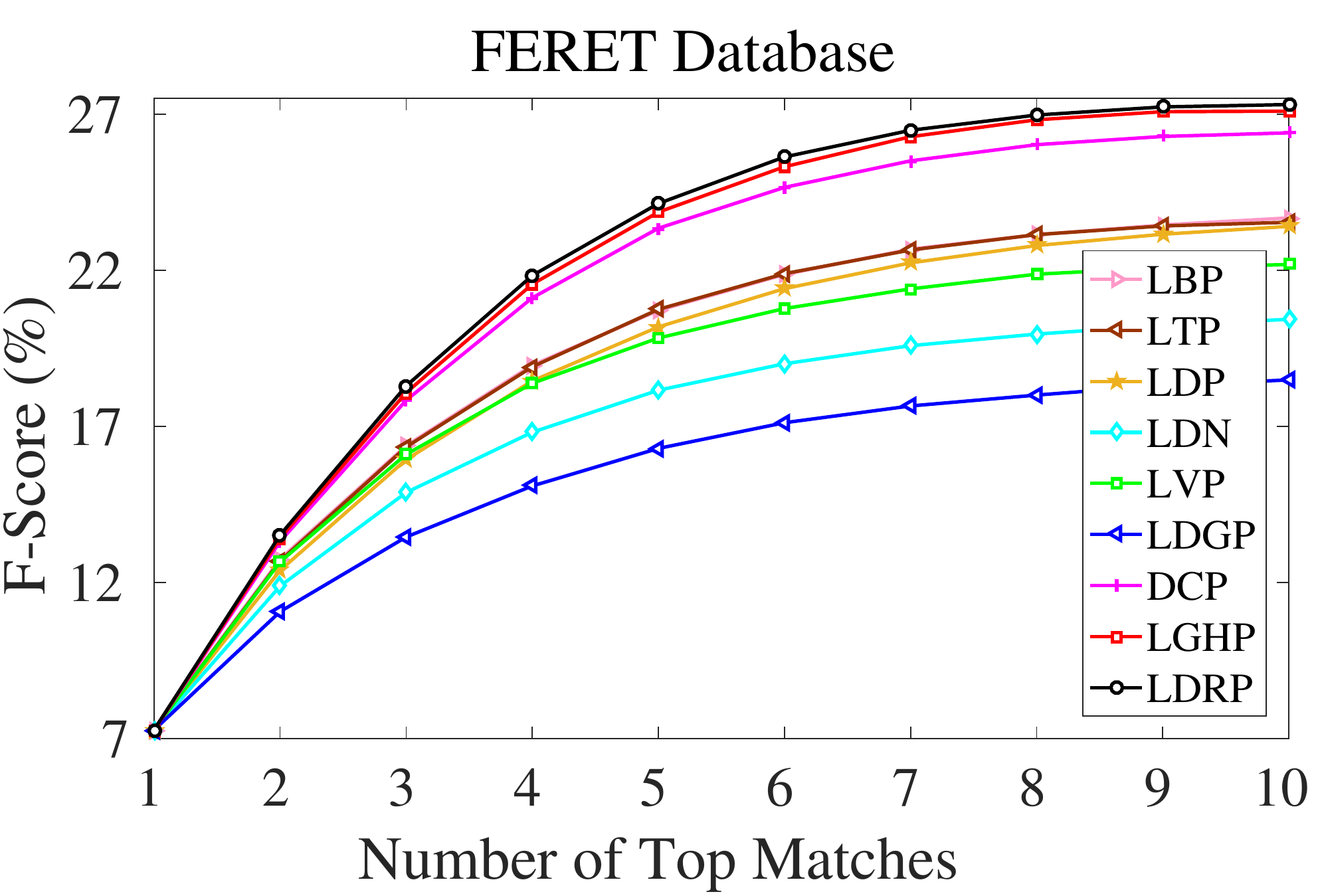}
    \caption{F-Score}
    \label{fig:feret-f}
  \end{subfigure}%
    \begin{subfigure}{.5\textwidth}
    \centering
    \includegraphics[clip=true, trim = 0 0 0.8cm 16cm, width=.98\linewidth]{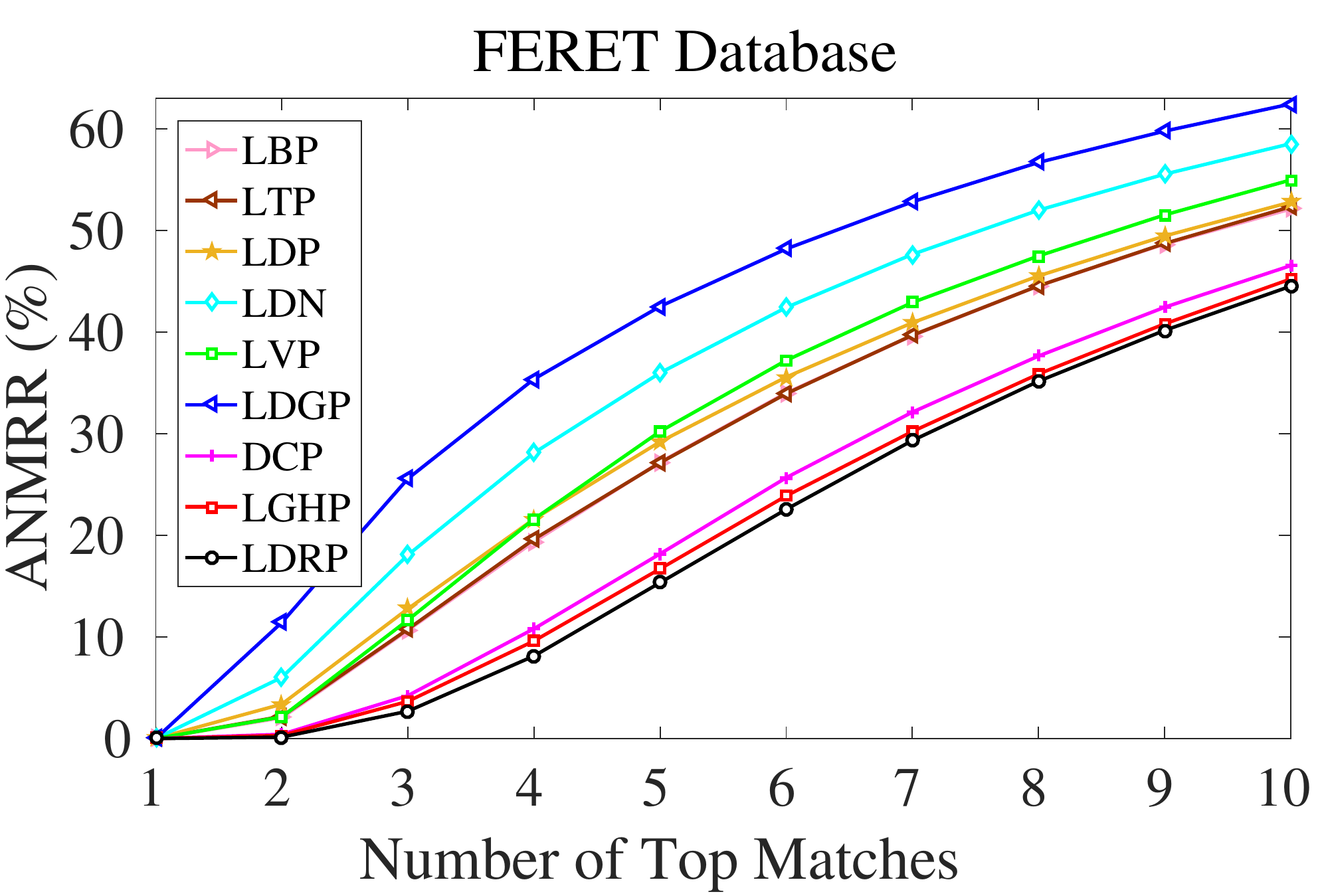}
    \caption{ANMRR}
    \label{fig:feret-anmrr}
  \end{subfigure}

  \caption{The results over FERET database in terms of the ARP, ARR, F-Score, and ANMRR vs number of retrieved images.}
  \label{fig:results_feret}
\end{figure*}

\section{Experimental Setup}
The image retrieval framework is used in this work for the experiments. The face retrieval using proposed local directional relation pattern (LDRP) descriptor is shown in Fig. \ref{fig:setup}. The best matching faces from a database are extracted against a query face using LDRP descriptor. The LDRP descriptor is generated for all the images of the database as well as for the query image. The similarity scores are computed between the LDRP descriptors of query image and database images using a distance measure technique. Note that high similarity score (or low distance) between two descriptors signifies that the corresponding images are more similar and vice-versa.

\subsection{Parameter Settings}
In this paper, the multi-scale LDRP is used for the comparison purpose with following by-default values of the parameters: $B=8$, $N=8$, $M_1=3$, and $M_2=6$. Four scales ($M=3,4,5,6$) are considered as $M\in[M_1,M_2]$ and LDRP descriptors are concatenated to form the final LDRP descriptor of dimension $4 \times 256 = 1024$. The other values of parameters $M_1$ and $M_2$ are also tested with the LDRP descriptor.

\subsection{Distances Measures}
The distance measures play an important role in image matching. The top $n$ number of faces is retrieved based on the lower distances, computed using the distance measures. The Chi-square distance measure is generally used in the experiments in this paper, whereas other distances like Euclidean, Cosine, L1, and D1 are also tested with the proposed descriptor to find its suitability \cite{ltrp}, \cite{mdlbp}.

\subsection{Evaluation Criteria}
The image retrieval algorithms are generally evaluated using precision, recall, f-score, and retrieval rank metrics. We have also used these metrics to judge the performance of the proposed method. In order to find the performance over a database, all the images of that database are converted as the query image one by one and metrics are computed. The average retrieval precision (ARP) and average retrieval rate (ARR) over whole database are calculated by taking the average of mean precision (MP) and mean recall (MR) of each category of that database respectively. The MP and MR for a category are computed by taking the mean of precision and recall by turning all of the images in that category as the query image one by one respectively. The F-score is computed from the ARP and ARR values as follows,
$$F-score=2 \times \frac{ARP\times ARR}{ARP+ARR}.$$
The average normalized modified retrieval rank (ANMRR) metric is also calculated to judge the rank of correctly retrieved faces \cite{anmrr}. The higher value of ARP, ARR and F-Score indicates the better retrieval performance and vice-versa, whereas the lower value of ANMRR indicates the better retrieval performance and vice-versa.

\subsection{Databases Used}
In the experiments, six challenging face databases are used to demonstrate the performance of the proposed LDRP descriptor. The used six face databases are PaSC \cite{pasc}, LFW \cite{lfw}, \cite{lfw1}, PubFig \cite{pubfig}, FERET \cite{feret}, \cite{feret1}, ESSEX \cite{essex}, and AT\&T (or ORL face database) \cite{att}. All the face images are down-sampled in $64\times64$ dimension. The PaSC still images face database is one of the challenging database having the variations like pose, illumination and blur \cite{pasc}. This database is having 293 subjects with total 9376 images (i.e., 32 images per subject). Viola Jones object detection method \cite{viola} is used for the facial part extraction over PaSC images. Finally, 8718 faces are successfully detected using Viola Jones detector in PaSC database. The unconstrained face retrieval is very challenging and close to real scenarios. The LFW and PubFig databases are having the images from the Internet. These images are taken in totally unconstrained scenarios without subjects cooperations. The variations like pose, lighting, expression, scene, camera, etc. are part of these databases. The gray-scaled version of LFW cropped database \cite{lfw1} is used in this work for the experiments. In the image retrieval framework, it is required to retrieve more than one (typically 5, 10, etc.) best matching images. So, it is required that the sufficient number of images should be present in each category of the database. Thus, the subjects having at least 20 images are considered. Total 3023 face images from 62 individuals are present in the LFW database used. The Public Figure database (i.e., PubFig) consists the images from 60 individuals and have 6472 number of total images \cite{pubfig}. The images are downloaded from the Internet directly following the urls given in this database (dead urls are removed). 

The ESSEX face database is very appealing database with a variety of background, scale, illumination, blur, and extreme variation of expressions \cite{essex}. The viola jones algorithm is used to extract the faces from the image \cite{viola}. Total 7740 faces are present from 392 subjects with nearly 20 images per subject.
The AT\&T face database (formerly known as the ORL Database of Faces) consisting of 10 images per subject from 40 different individuals \cite{att}. Different lighting conditions and facial expressions are present in the images of some subjects. The images in AT\&T database are captured in an upright and frontal position with a dark homogeneous background.
"Portions of the research in this paper use the FERET database of facial images collected under the FERET program, sponsored by the DOD Counterdrug Technology Development Program Office" \cite{feret}, \cite{feret1}. The Color-FERET database is considered due to the severe variations in the expression and pose (13 different poses). The subjects having at least 20 images are considered and all the color images are converted into the grayscale images. In this work, 4053 images from 141 subjects are present in FERET database.

\section{Face Retrieval Experimental Results}

In order to demonstrate the superior performance of proposed local directional relation pattern (LDRP) descriptor for face retrieval, the state-of-the-art face descriptors like LBP \cite{lbp}, LTP
\cite{ltp}, LDP \cite{ldp}, LDN \cite{ldn}, LVP \cite{lvp}, LDGP \cite{ldgp}, DCP \cite{dcp}, and LGHP \cite{lghp} are used for the comparison over all six databases. The dimensions of LBP, LTP, LDP, LDN, LVP, LDGP, DCP, LGHP, and LDRP descriptors are 256, 512, 1024, 64, 1024, 65, 512, 9216, and 1024 respectively. Note that, all these descriptors have shown very promising results for facial analysis under varying conditions such as rotation, scale, background, blur, illumination, pose, masking, etc. The parameters for all the compared descriptors are  used as per their source papers.
The results comparison using different descriptors in terms of the ARP (\%), ARR (\%), F-score (\%) and ANMRR (\%) vs number of retrieved images ($n$) over PaSC database in Fig. \ref{fig:results_pasc}, LFW database in Fig. \ref{fig:results_lfw}, PubFig database in Fig. \ref{fig:results_pubfig}, ESSEX database in Fig. \ref{fig:results_essex}, AT\&T database in Fig. \ref{fig:results_att} and FERET database in Fig. \ref{fig:results_feret}, respectively. 
The LDRP descriptor outperforms the existing face descriptors over PaSC database as it has the highest values for ARP, ARR, and F-Score and lowest values for ANMRR (see Fig. \ref{fig:pasc-arp}-\ref{fig:pasc-anmrr}). PaSC database is having variations like scale, blur, pose and illumination. The LFW and PubFig databases are fully unconstrained database. It is observed that the performance of LDRP descriptor is comparable with the LGHP descriptor over both LFW and PubFig databases as depicted in Fig. \ref{fig:lfw-arp}-\ref{fig:lfw-anmrr} and Fig. \ref{fig:pubfig-arp}-\ref{fig:pubfig-anmrr} respectively, whereas the dimension of LDRP(dim: 1024) is much lower than the dimension of LGHP(dim: 9216). Thus, the time efficiency of LDRP is far better than LGHP while maintaining similar performance over the unconstrained databases. 

The performance of LDRP is improved significantly over the frontal face databases, but with other variations like scale, background, illumination, blur, expressions, etc. as shown in Fig. \ref{fig:essex-arp}-\ref{fig:essex-anmrr} over ESSEX database and Fig. \ref{fig:att-arp}-\ref{fig:att-anmrr} over AT\&T database. The proposed LDRP descriptor is outstanding over both ESSEX and AT\&T databases as compared to the state-of-the-art face descriptors. In order to test the suitability of LDRP in pose and scale variations, FERET database is considered because it has the faces with huge pose and scale variations. The results over FERET database is summarized in Fig. \ref{fig:feret-arp}-\ref{fig:feret-anmrr}. It is found that the LDRP is equivalent to the other top performing descriptors such as LGHP over FERET database, whereas its dimension is much lower than LGHP. 

From the experimental results of Fig. \ref{fig:results_pasc}-\ref{fig:results_feret}, it is observed that LDRP and LGHP descriptors outperforms the other descriptors over the PaSC, LFW, PubFig, ESSEX, AT\&T, and FERET face databases. It is also noticed that the LGHP descriptor is mostly the second best performing method and having very good discriminative features, but redundant and at the cost of increased dimensionality. Whereas, it is clear from the results of Fig. \ref{fig:results_pasc}-\ref{fig:results_feret} that despite of having much lower dimensionality, the proposed LDRP descriptor is either outperforms LGHP or has the comparable performance against LGHP. The proposed LDRP descriptor outperforms the existing descriptors over six challenging face datasets having a varying number of samples and complexities, which shows the scalability of the proposed LDRP descriptor.

\begin{table}[!t]
\caption{The performance comparison of LDRP descriptors in terms of the ARP(\%) for $n=5$ number of retrieved images over the PubFig, PaSC, LFW, FERET, AT\&T, and ESSEX face databases by varying the values of $M_1$ and $M_2$ (i.e. multiscale parameters) for different radius and the number of local neighborhoods. The Chi-square distance is used. The highest ARP values are highlighted in bold for each database.}
\label{t1}
\begin{center}
\begin{tabular}{cccccccc}
\hline
\\[-0.65em] \multirow{2}{*}{$M_1$} & \multirow{2}{*}{$M_2$} & \multicolumn{6}{c}{Face Databases}\\ \\[-0.65em]
\cline{3-8} 
\\[-0.65em]  & & PubFig & PaSC & LFW & FERET & AT\&T & ESSEX \\ \\[-0.65em]
\hline
\\[-0.65em]
3 & 3 & 39.90 & 26.25 & 32.18 & 67.51 & 89.80 & 82.78\\ \\[-0.85em]
3 & 4 & 43.81 & 33.43 & 37.07 & 74.44 & 94.95 & 98.34\\ \\[-0.85em]
3 & 5 & 45.83 & 36.92 & 39.36 & 75.62 & 95.35 & 98.88\\ \\[-0.85em]
3 & 6 & 47.76 & 39.12 & 41.04 & \textbf{75.91} & 96.10 & 99.05\\ \\[-0.85em]
3 & 7 & 48.61 & 40.96 & 42.15 & 75.90 & \textbf{96.30} & 99.11\\ \\[-0.85em]
4 & 4 & 43.24 & 38.28 & 38.42 & 70.61 & 94.50 & 98.44\\ \\[-0.85em]
4 & 5 & 46.53 & 42.09 & 41.00 & 73.47 & 94.45 & 99.01\\ \\[-0.85em]
4 & 6 & 48.08 & 44.20 & 42.75 & 74.07 & 95.65 & 99.07\\ \\[-0.85em]
4 & 7 & \textbf{49.22} & 45.74 & \textbf{44.05} & 74.07 & 96.05 & \textbf{99.14}\\ \\[-0.85em]
5 & 5 & 44.17 & 42.99 & 37.26 & 69.12 & 92.65 & 98.91\\ \\[-0.85em]
5 & 6 & 46.50 & 45.42 & 40.47 & 71.21 & 94.35 & 99.05\\ \\[-0.85em]
5 & 7 & 48.01 & \textbf{46.86} & 42.51 & 71.87 & 95.30 & 99.10\\ \\[-0.85em]
6 & 6 & 44.20 & 43.86 & 37.55 & 67.86 & 93.55 & 98.87\\ \\[-0.85em]
6 & 7 & 46.32 & 45.97 & 40.47 & 69.86 & 95.10 & 99.03\\ \\[-0.85em]
7 & 7 & 43.66 & 44.08 & 37.58 & 66.71 & 93.70 & 98.86\\ \\[-0.65em]
\hline
\end{tabular}
\end{center}
\end{table}

\begin{table}[!t]
\caption{The ARP(\%) using proposed LDRP descriptor with Euclidean, Cosine, L1, D1, and Chi-square distance measures over the PubFig, PaSC, LFW, FERET, AT\&T, and ESSEX face databases. The number of retrieved images ($n$) is 5. The highest ARP values are highlighted in bold for each database.}
\label{t2}
\begin{center}
\begin{tabular}{lcccccc}
\hline
\\[-0.65em]\multirow{2}{*}{Distance} & \multicolumn{6}{c}{Face Databases}\\ \\[-0.65em]
\cline{2-7} 
\\[-0.65em] & PubFig & PaSC & LFW & FERET & AT\&T & ESSEX \\ \\[-0.65em]
\hline
\\[-0.65em]
Euclidean & 35.49 & 27.53 & 30.17 & 65.09 & 88.45 & 95.83\\ \\[-0.85em]
Cosine & 37.24 & 30.93 & 32.37 & 67.23 & 93.40 & 97.46\\ \\[-0.85em]
L1 & 44.91 & 36.96 & 38.57 & 75.54 & 95.50 & 98.80\\ \\[-0.85em]
D1 & 45.25 & 37.62 & 39.14 & 75.53 & 95.65 & 98.86\\ \\[-0.85em]
Chi-square & \textbf{47.76} & \textbf{39.12} & \textbf{41.04} & \textbf{75.91} & \textbf{96.10} & \textbf{99.05}\\ \\[-0.65em]
\hline
\end{tabular}
\end{center}
\end{table}

\begin{table}[!t]
\caption{The ARP(\%) using LGHP descriptor \cite{lghp} with Euclidean, Cosine, L1, D1, and Chi-square distance measures over the PubFig, PaSC, LFW, FERET, AT\&T, and ESSEX face databases. The number of retrieved images ($n$) is 5. The highest ARP values are highlighted in bold for each database.}
\label{t3}
\begin{center}
\begin{tabular}{lcccccc}
\hline
\\[-0.65em]\multirow{2}{*}{Distance} & \multicolumn{6}{c}{Face Databases}\\ \\[-0.65em]
\cline{2-7} 
\\[-0.65em] & PubFig & PaSC & LFW & FERET & AT\&T & ESSEX \\ \\[-0.65em]
\hline
\\[-0.65em]
Euclidean & 34.28 & 24.21 & 27.82 & 49.11 & 81.90 & 90.57\\ \\[-0.85em]
Cosine & 36.10 & 25.82 & 29.51 & 54.60 & 83.10 & 91.75\\ \\[-0.85em]
L1 & 43.89 & 30.25 & 36.30 & 69.60 & 91.65 & 95.09\\ \\[-0.85em]
D1 & 43.96 & 30.29 & 36.33 & 69.71 & 91.75 & 95.11\\ \\[-0.85em]
Chi-square & \textbf{47.05} & \textbf{32.93} & \textbf{39.53} & \textbf{74.88} & \textbf{94.00} & \textbf{95.73}\\ \\[-0.65em]
\hline
\end{tabular}
\end{center}
\end{table}

\begin{figure*}[!t]
  \begin{subfigure}{.5\textwidth}
    \centering
    \includegraphics[clip=true, trim = 0 0 0.8cm 15cm, width=.98\linewidth]{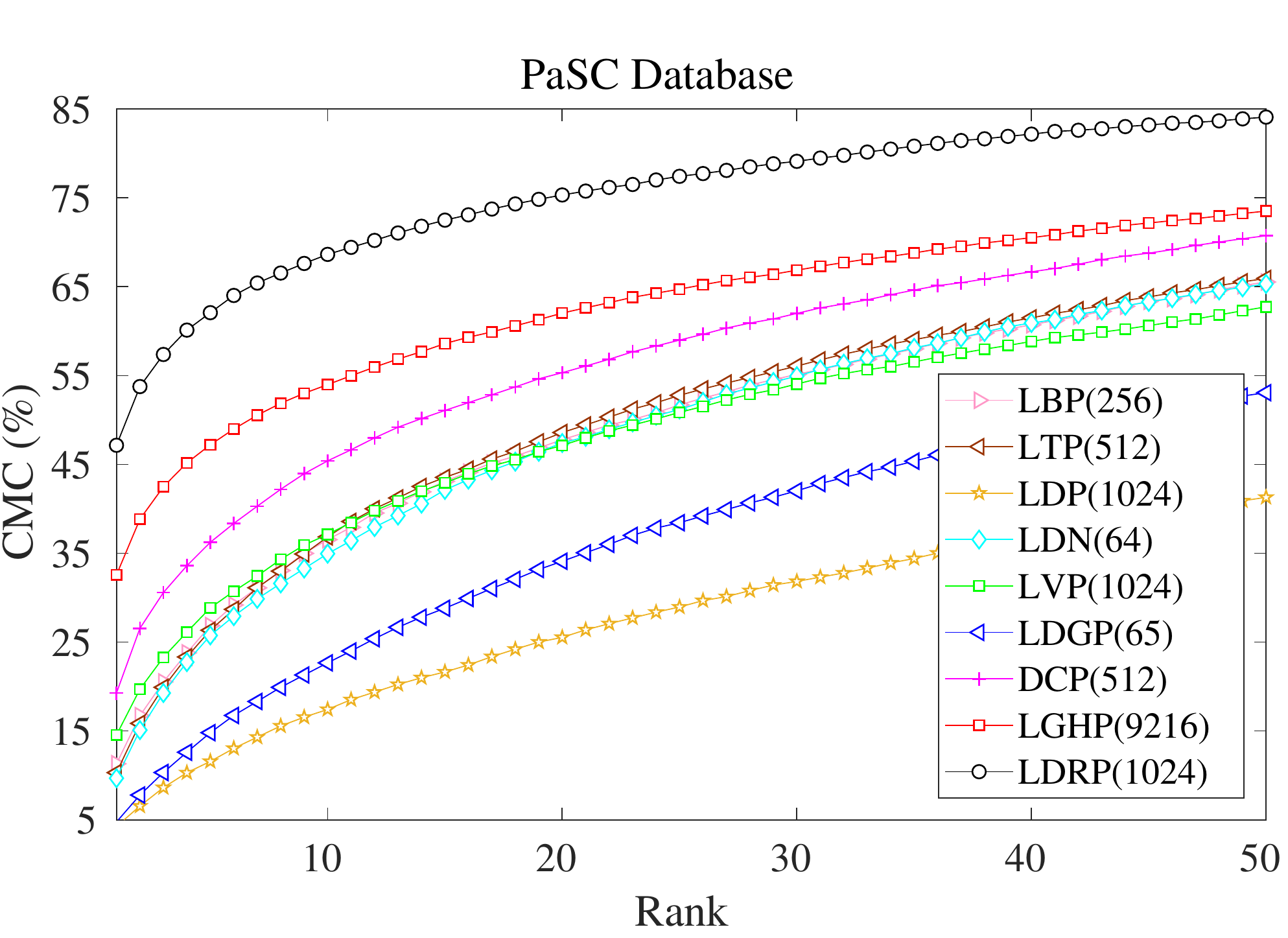}
    \caption{CMC over PaSC}
    \label{fig:pasc-cmc}
  \end{subfigure}%
  \begin{subfigure}{.5\textwidth}
    \centering
    \includegraphics[clip=true, trim = 0 0 0.8cm 15cm, width=.98\linewidth]{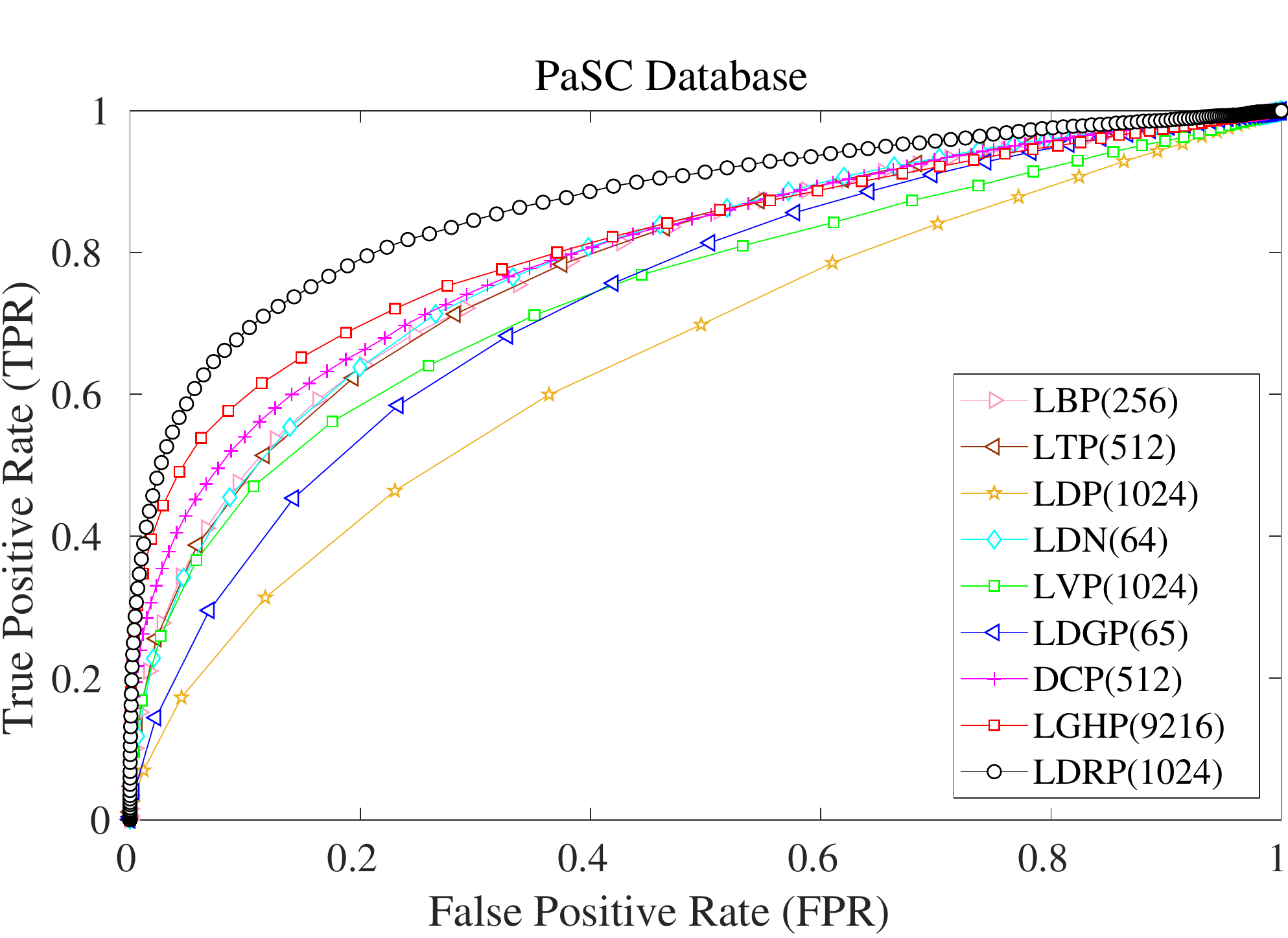}
    \caption{ROC over PaSC}
    \label{fig:pasc-roc}
  \end{subfigure}

  \begin{subfigure}{.5\textwidth}
    \centering
    \includegraphics[clip=true, trim = 0 0 0.8cm 15cm, width=.98\linewidth]{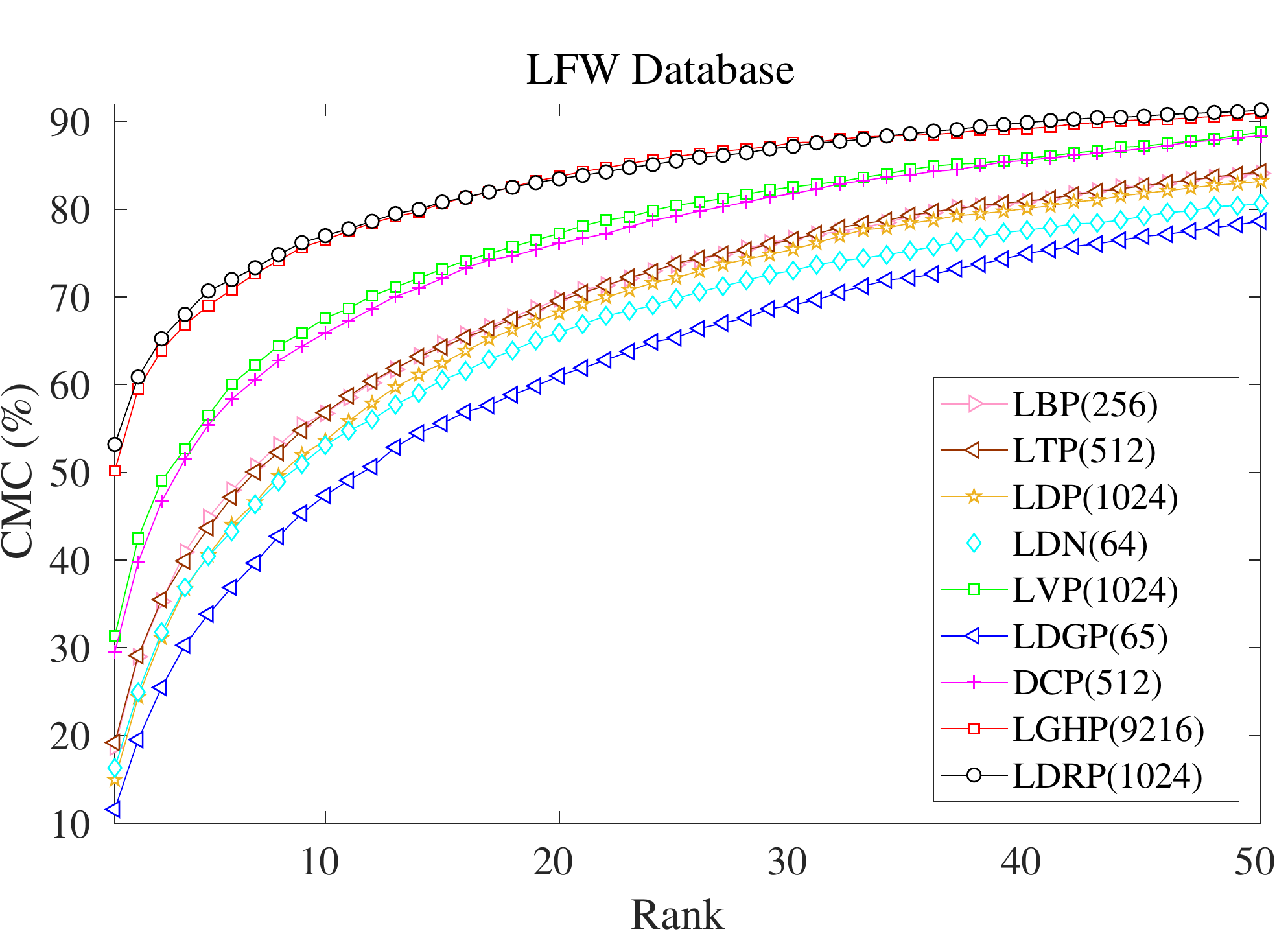}
    \caption{CMC over LFW}
    \label{fig:lfw-cmc}
  \end{subfigure}%
  \begin{subfigure}{.5\textwidth}
    \centering
    \includegraphics[clip=true, trim = 0 0 0.8cm 15cm, width=.98\linewidth]{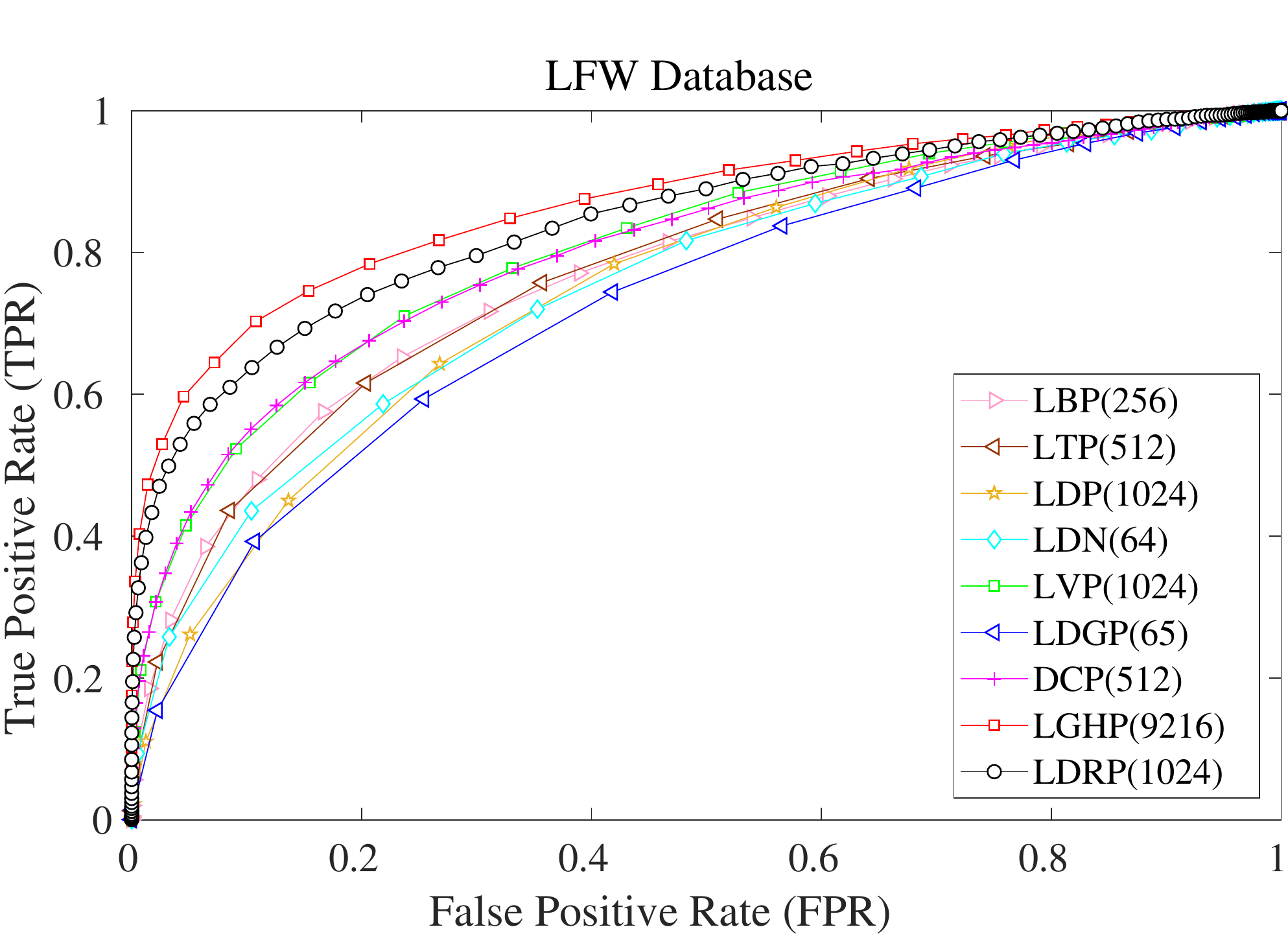}
    \caption{ROC over LFW}
    \label{fig:lfw-roc}
  \end{subfigure}

  \begin{subfigure}{.5\textwidth}
    \centering
    \includegraphics[clip=true, trim = 0 0 0.8cm 15cm, width=.98\linewidth]{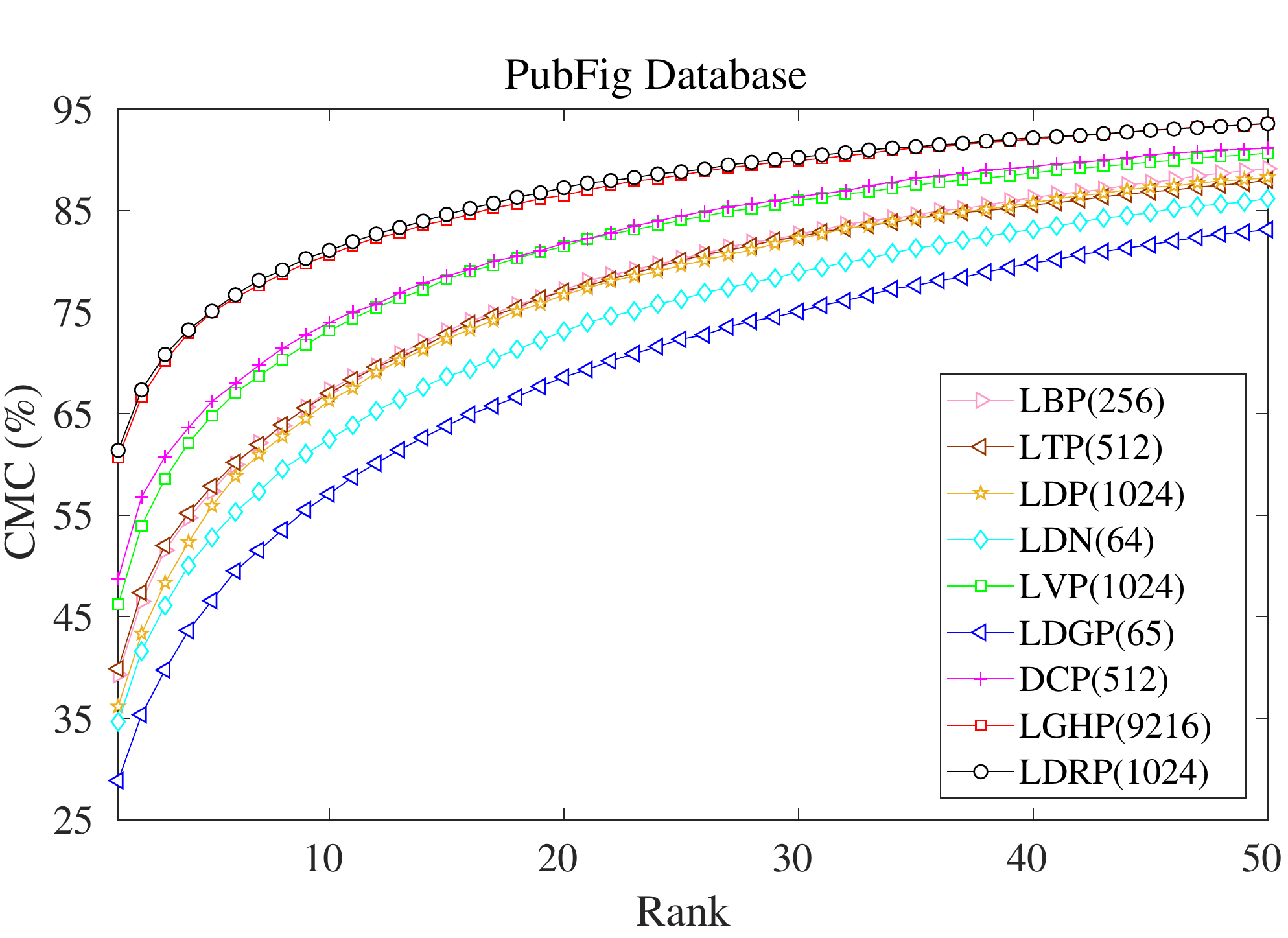}
    \caption{CMC over PubFig}
    \label{fig:pubfig-cmc}
  \end{subfigure}%
  \begin{subfigure}{.5\textwidth}
    \centering
    \includegraphics[clip=true, trim = 0 0 0.8cm 15cm, width=.98\linewidth]{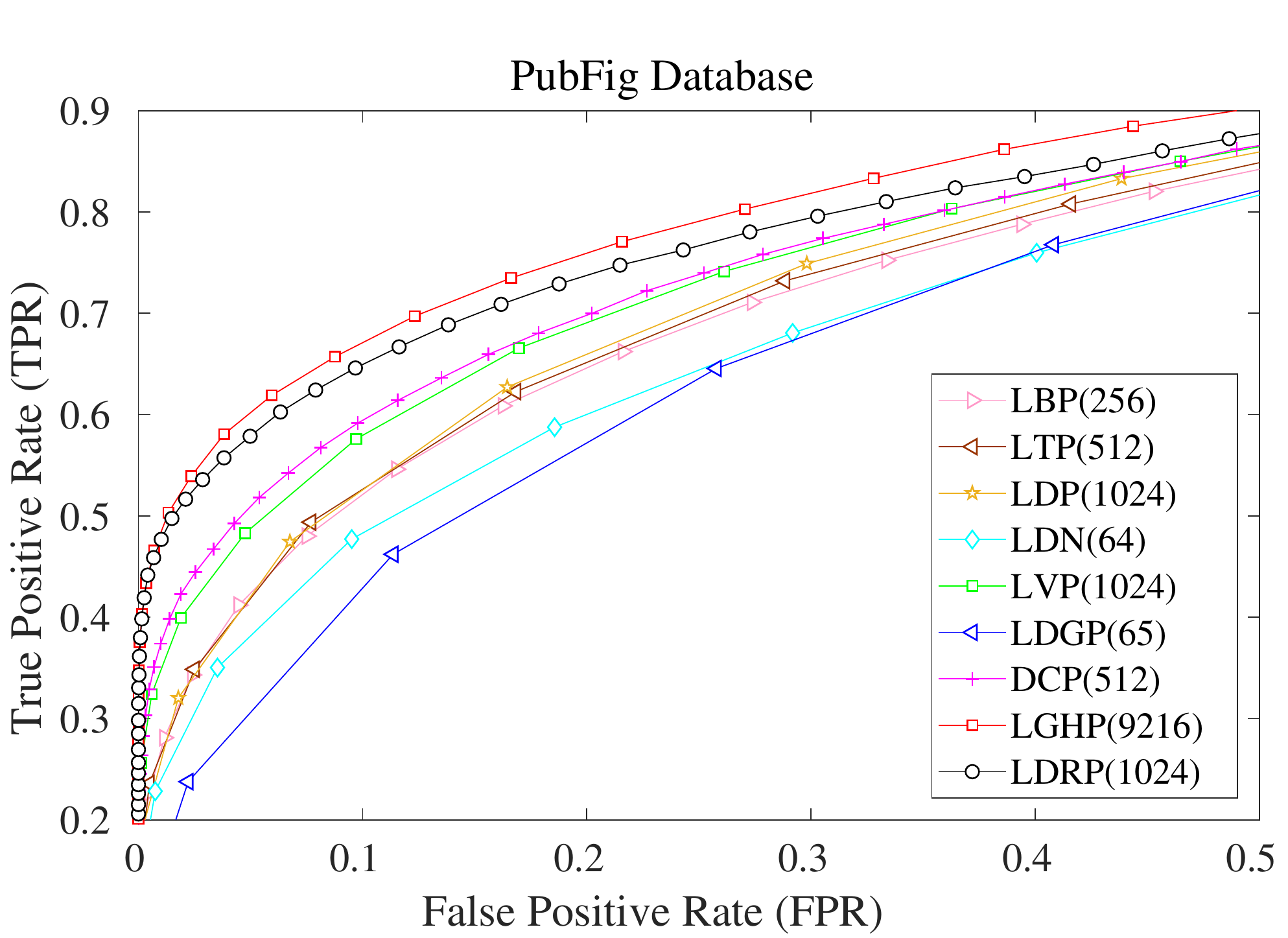}
    \caption{ROC over PubFig}
    \label{fig:pubfig-roc}
  \end{subfigure}
  
  \caption{The results over PaSC, LFW, and PubFig databases in terms of the CMC and ROC for face recognition.}
  \label{fig:results-recognition1}
\end{figure*}

\begin{figure*}[!t]
  \begin{subfigure}{.5\textwidth}
    \centering
    \includegraphics[clip=true, trim = 0 0 0.8cm 15cm, width=.98\linewidth]{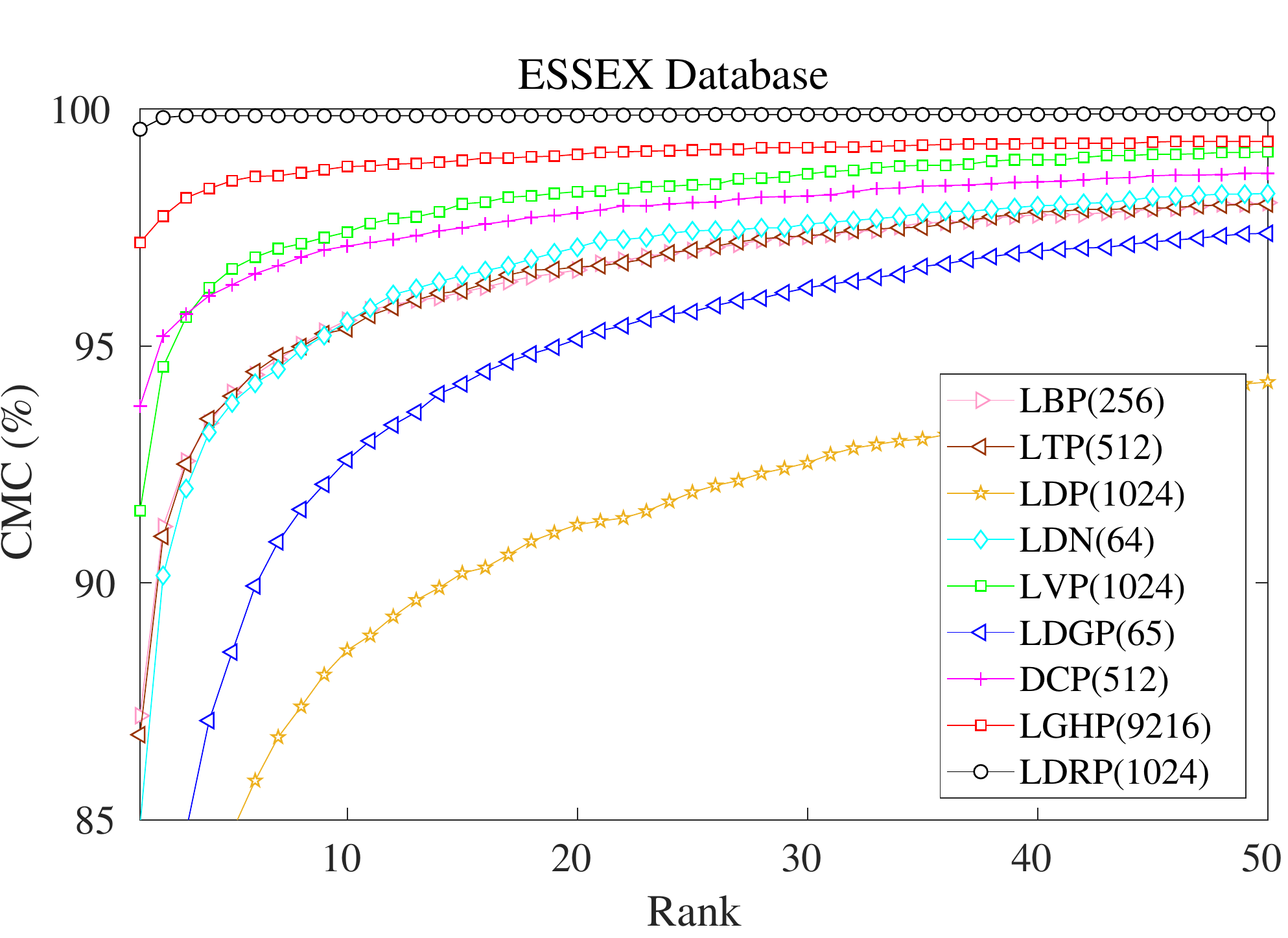}
    \caption{CMC over ESSEX}
    \label{fig:essex-cmc}
  \end{subfigure}%
  \begin{subfigure}{.5\textwidth}
    \centering
    \includegraphics[clip=true, trim = 0 0 0.8cm 15cm, width=.98\linewidth]{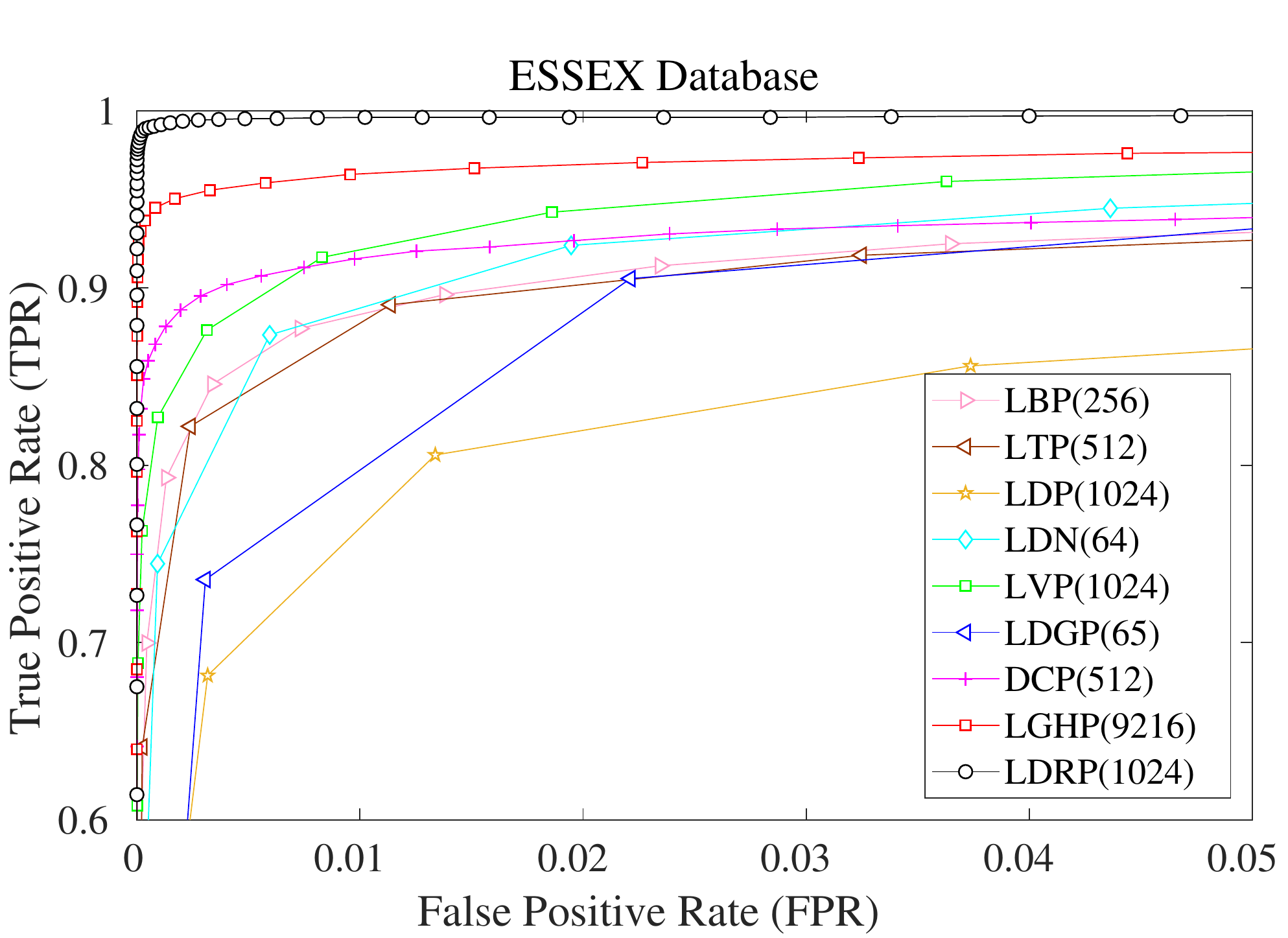}
    \caption{ROC over ESSEX}
    \label{fig:essex-roc}
  \end{subfigure}
  
  \begin{subfigure}{.5\textwidth}
    \centering
    \includegraphics[clip=true, trim = 0 0 0.8cm 15cm, width=.98\linewidth]{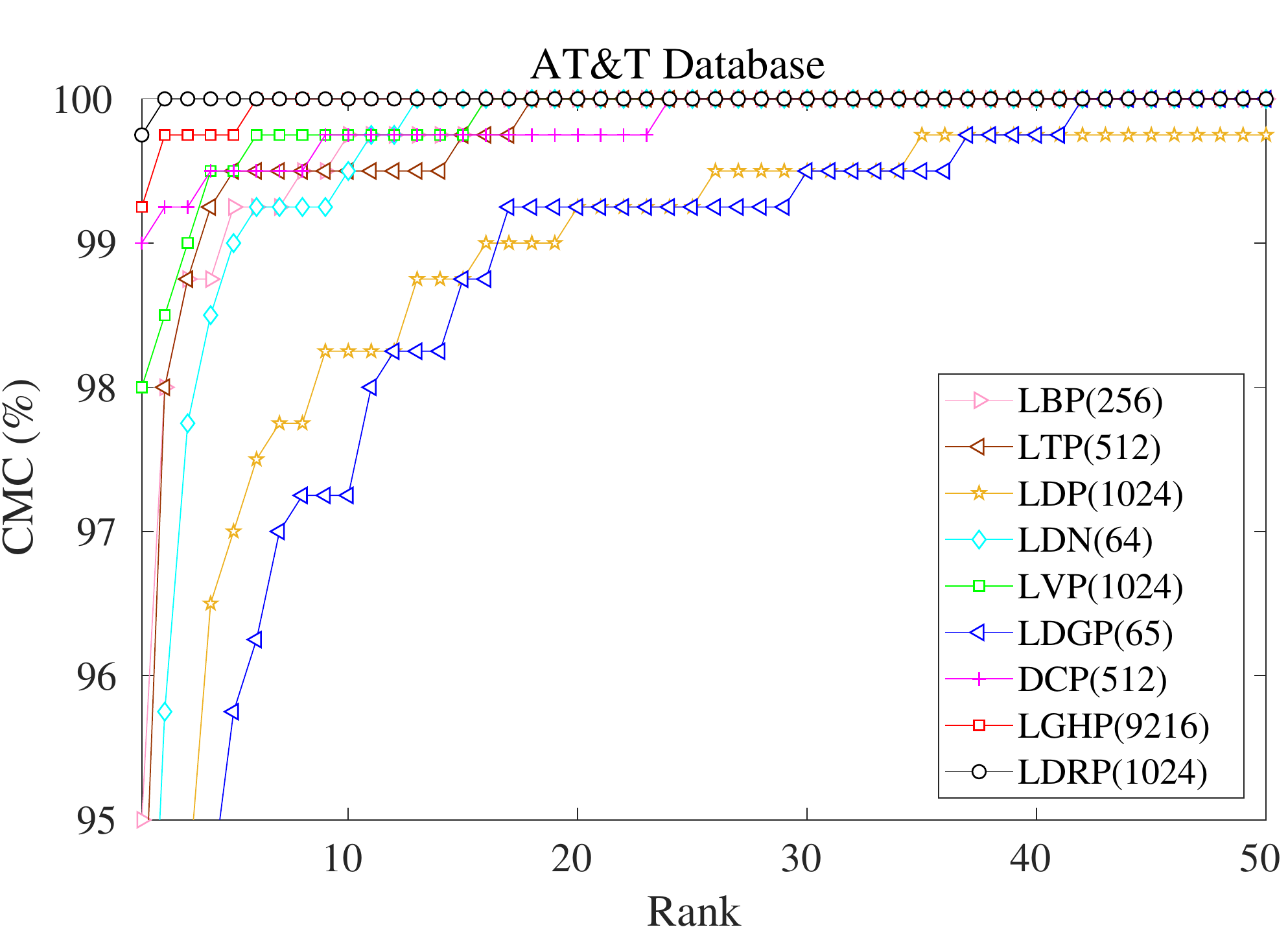}
    \caption{CMC over AT\&T}
    \label{fig:att-cmc}
  \end{subfigure}%
  \begin{subfigure}{.5\textwidth}
    \centering
    \includegraphics[clip=true, trim = 0 0 0.8cm 15cm, width=.98\linewidth]{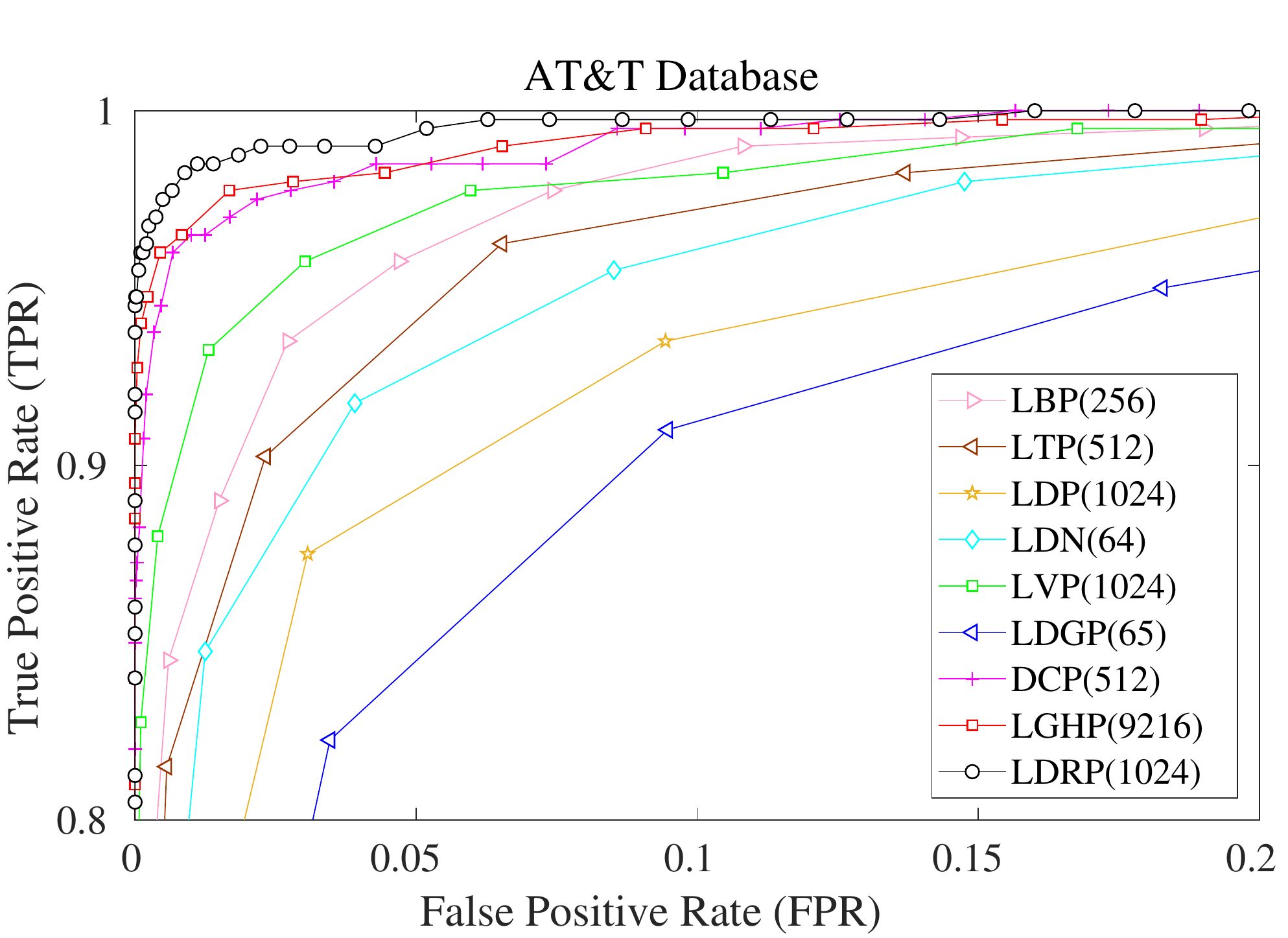}
    \caption{ROC over AT\&T}
    \label{fig:att-roc}
  \end{subfigure}
  
  \begin{subfigure}{.5\textwidth}
    \centering
    \includegraphics[clip=true, trim = 0 0 0.8cm 15cm, width=.98\linewidth]{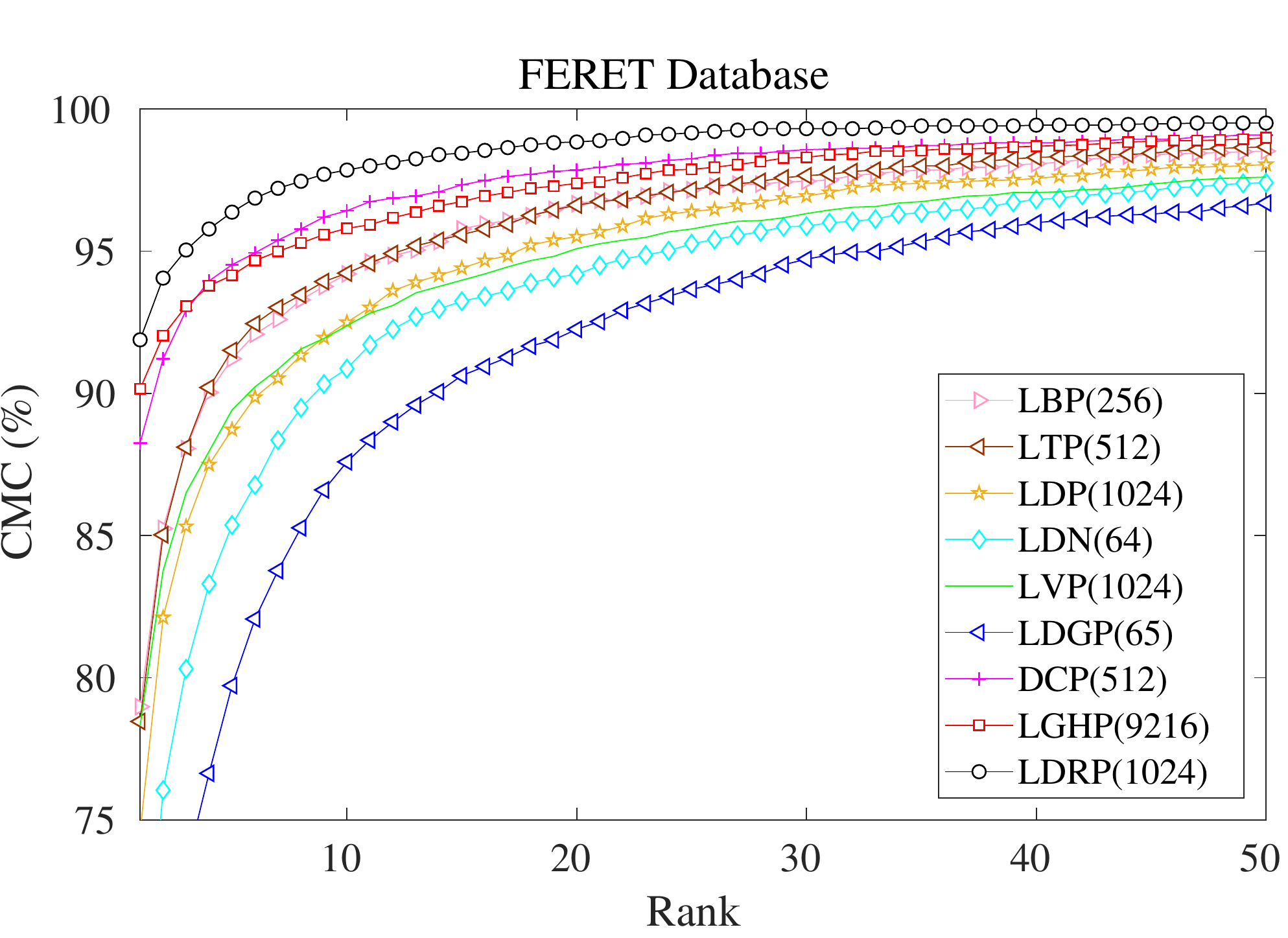}
    \caption{CMC over FERET}
    \label{fig:feret-cmc}
  \end{subfigure}%
  \begin{subfigure}{.5\textwidth}
    \centering
    \includegraphics[clip=true, trim = 0 0 0.8cm 15cm, width=.98\linewidth]{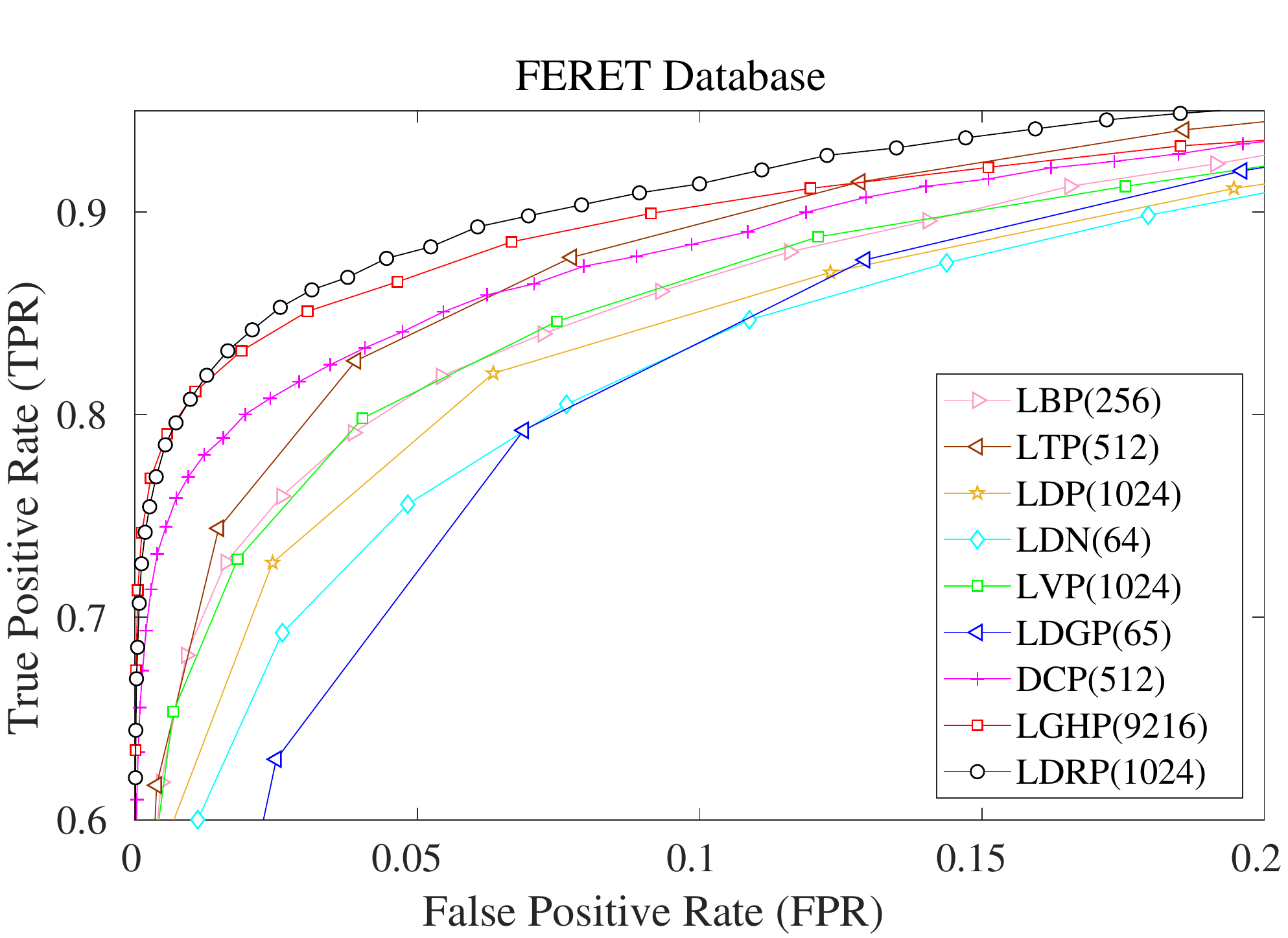}
    \caption{ROC over FERET}
    \label{fig:feret-roc}
  \end{subfigure}
  
  \caption{The results over ESSEX, AT\&T, and FERET databases in terms of the CMC and ROC for face recognition.}
  \label{fig:results-recognition2}
\end{figure*}

\section{Performance Analysis}
This section is devoted to the performance analysis of the proposed descriptor. First, the effect of size of local neighborhood and multi-scale is analyzed, then the effect of distance measure is tested over each database, after that the performance comparison is done for face recognition, next the robustness of proposed descriptor is analyzed against pose and expression, next the performance comparison of proposed descriptor is performed with pre-trained CNN features, next the experiment is conducted over large-scale dataset to show the scalability of the proposed descriptor, and finally the performance comparison is made with deep learning based DLib face descriptor. 

\subsection{Effect of Local Neighborhood}
In the previous experiment, the LDRP parameters are as follows: $N=8$, $M_1=3$, and $M_2=6$. In this subsection, the performance of LDRP is tested by varying the values of $M_1$ and $M_2$ from 3 to 7. The ARP(\%) values for $n=5$ number of retrieved images over the PubFig, PaSC, LFW, FERET, AT\&T, and ESSEX face databases are summarized in Table \ref{t1}. The highest ARP for a particular database is highlighted in bold. It is observed that LDRP with $M_1=3$ and $M_2=6$ is having highest precision only over the FERET database. It is due to the huge pose variations present in the FERET database. The performance of LDRP is improved over each database for upper limit as a maximum (i.e. $M_2=7$). From this experiment, it is also clear that $M_1=4$ and $M_2=7$ are better suited for unconstrained scenario. Though, $M_1=3$ and $M_2=6$ are used in the previous results, the performance of LDRP can be further improved by considering $M_1=4$ and $M_2=7$.

\subsection{Effect of Distance Measures}
In order to find out the suitable distance measure for the proposed descriptor, this experiment is conducted by using the different distance measures. The Euclidean, Cosine, L1, D1, and Chi-square distances are used in this experiment \cite{ltrp}, \cite{mdlbp}. The ARP in percentage over the PubFig, PaSC, LFW, FERET, AT\&T, and ESSEX databases for $n=5$ number of top matches are displayed in Table \ref{t2} using the proposed LDRP descriptor. In this experiment, the default parameter values are used for LDRP descriptor (i.e., $N=8$, $M_1=3$, and $M_2=6$). The best result over a database is highlighted in bold. It is observed from the results that the Chi-square distance measure is better suited with the proposed LDRP descriptor for a face retrieval task. The effect of distances is also tested with LGHP descriptor \cite{lghp} in Table \ref{t3} and interestingly, the Chi-square distance is also better suited for LGHP. The Chi-square distance is performing better because the descriptors are in the form of histograms and representing the occurrences of patterns in some form.

\subsection{Face Recognition Performance Comparison}
Most of the local descriptors are proposed for the face recognition problem. Thus, a face recognition experiment is also conducted in this paper to understand the behavior of proposed LDRP descriptor in recognition framework. The face recognition performance is compared in terms of the Cumulative Match Characteristic (CMC) and Receiver Operating Characteristic (ROC) metrics. The CMC curve is computed by finding the cumulative recognition rate against the rank. The ROC curve is generated from the True Positive Rate (TPR) and False Positive Rate (FPR). The TPR and FPR are computed by varying a threshold over intra-class (i.e. genuine) and inter-class (i.e. impostor) scores. The scores are computed using the Chi-square distance measure. The multiple thresholds are chosen between the lowest and highest scores with a step value of $0.001$. If the lowest and highest scores are $s_l$ and $s_h$, respectively, then the thresholds are $s_l$, $s_l+0.001$, $s_l+2\times0.001$, $\dots$, $s_h$. Thus, the number of thresholds is $\frac{s_h-s_l}{0.001}+1$. The CMC and ROC curves over each face database are plotted in Fig. \ref{fig:results-recognition1} and Fig. \ref{fig:results-recognition2}. The dimension of the descriptors is also mentioned along with the descriptor name. LDRP outperforms all the descriptors over PaSC database in recognition framework also as depicted in Fig. \ref{fig:pasc-cmc}-\ref{fig:pasc-roc}. It is also pointed that the performance of LDRP is also pretty good over unconstrained databases such as LFW and PubFig. It can be seen in Fig. \ref{fig:lfw-cmc}-\ref{fig:pubfig-roc} that the LDRP(1024) is comparable to the LGHP(9216) in-spite of having much lower dimensional feature vector. The LDRP is better in CMC as compared to ROC over LFW and PubFig unconstrained databases, possibly due to the random directional edges which lead to the less inter-class score variability. This problem can be minimized by applying some alignment technique before applying the LDRP descriptor. The recognition result of LDRP is quite impressive over ESSEX and AT\&T face databases as well (see Fig. \ref{fig:essex-cmc}-\ref{fig:att-roc}). The LDRP outperforms other descriptors over FERET database also in recognition framework as shown in Fig. \ref{fig:feret-cmc}-\ref{fig:feret-roc} which justifies the robustness against pose and expression.

\subsection{Pose and Expression Robustness Analysis}

\begin{table}[!t]
\caption{The performance comparison of LDRP descriptor with other descriptors in terms of the accuracy under different poses and expression of FERET database. The poses such as `\textit{quarter} (\textit{q})', `\textit{half} (\textit{h})', `\textit{orthogonal} (\textit{o})', and `\textit{random} (\textit{r})' are used as the Probe and the frontal poses `\textit{frontal} (\textit{f})' are taken as the Gallery. For expression, the random expression `\textit{fb}' is used as Probe and the neutral expression `\textit{fa}' is taken as the Gallery. The Chi-square distance is used as the similarity measure. The highest accuracy values are highlighted in bold in each column.}
\label{t4}
\begin{center}
\begin{tabular}{cccccc}
\hline \\[-0.65em] 
\multirow{3}{*}{Descriptors} & \multicolumn{5}{c}{Robustness Type}\\ \\[-0.65em]
\cline{2-6} \\[-0.65em]
 & \multicolumn{4}{c}{Pose}	& \multirow{2}{*}{Expression} \\ \\[-0.65em]
\cline{2-5} \\[-0.65em]
 & Quarter & Half & Orthogonal & Random & \\ \\[-0.65em]
\cline{1-6}
\\[-0.65em]
LBP(256) & 66.62 & 37.29 & 23.17 & 56.00 & 85.14\\ \\[-0.85em]
LTP(512) & 67.05 & 35.02 & 20.91 & 55.49 & 83.38\\ \\[-0.85em]
LDP(1024) & 61.93 & 28.70 & 16.50 & 55.38 & 85.64\\ \\[-0.85em]
LDN(64) & 54.12 & 26.04 & 16.37 & 46.58 & 74.56\\ \\[-0.85em]
LVP(1024) & 59.52 & 24.78 & 13.10 & 51.76 & 90.18\\ \\[-0.85em]
LDGP(65) & 43.04 & 20.86 & 11.46 & 37.47 & 63.22\\ \\[-0.85em]
DCP(512) & 77.70 & 39.57 & 22.42 & 65.01 & 93.45\\ \\[-0.85em]
LGHP(9216) & 76.70 & 35.02 & 19.77 & 62.53 & 94.96\\ \\[-0.85em]
LDRP(1024) & \textbf{80.68} & \textbf{41.85} & \textbf{25.32} & \textbf{70.19} & \textbf{95.21}\\
\hline
\end{tabular}
\end{center}
\end{table}

In order to reveal the improved robustness of LDRP descriptor against pose and expression, this experiment is conducted over different poses and expression of FERET database. The images of FERET database have different poses such as `Frontal Pose (f)', `Quarter Pose (q)', `Half Pose (h)', `Orthogonal Pose (o)', and `Random Pose (r)'. Note that the orthogonal category `o' is having faces nearly at $90^o$ as compared to the frontal category `f' either in the left or in the right direction. Similarly the half category `h' and quarter category `q' are having faces nearly at $45^o$ and $25^o$ respectively as compared to the frontal category `f' either in the left or in the right direction. Among frontal faces `f', the category `fa' has the neutral expression faces, whereas the category `fb' has the random expression faces. In this experiment, the performance is judged on the basis of the recognition accuracy (i.e. similar to CMC at rank 1). Gallery and probe sets are considered to find the recognition accuracy over probe set. In order to test the pose robustness, the frontal face category `f'=`fa'+`fb' is taken as a combined gallery set and rest of varying sets are considered as the probe sets individually. While in case of expression experiment, the neutral frontal face (`fa') is taken as the gallery set and random expression frontal face (`fb') is considered as the probe set. The recognition accuracy using different descriptors is summarized in Table \ref{t4} for both pose and expression experiments. It is clear from Table \ref{t4} that the LDRP descriptor beats the other descriptors in all poses. It can be also seen in Table \ref{t4} that LDRP surpasses in case of expression experiment as well with 95.21\% of recognition accuracy.

\begin{figure*}[!t]
  \begin{subfigure}{.5\textwidth}
    \centering
    \includegraphics[width=.98\linewidth]{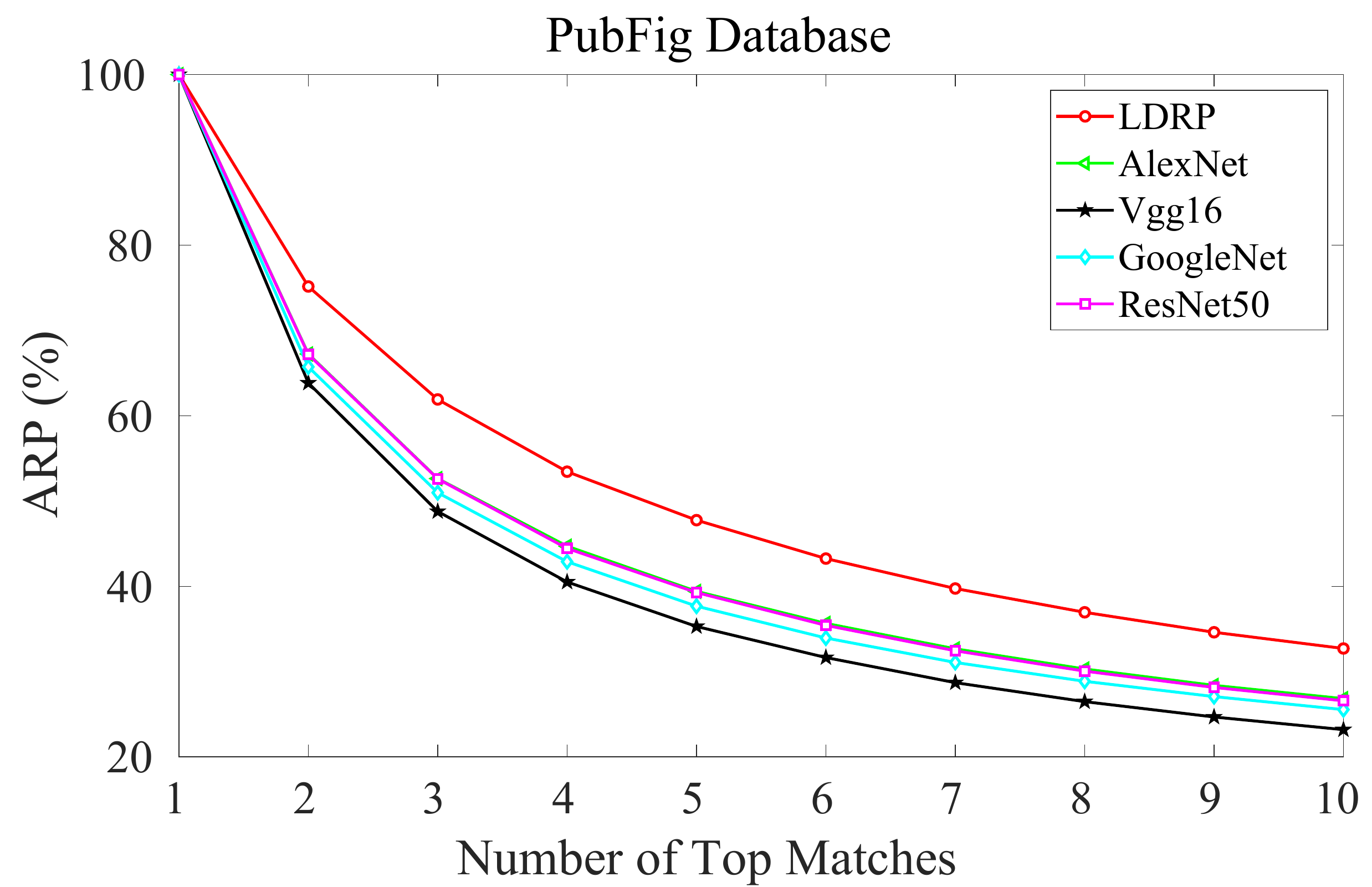}
    \caption{Over PubFig Database}
    \label{fig:pubfig-cnn}
  \end{subfigure}%
  \begin{subfigure}{.5\textwidth}
    \centering
    \includegraphics[width=.98\linewidth]{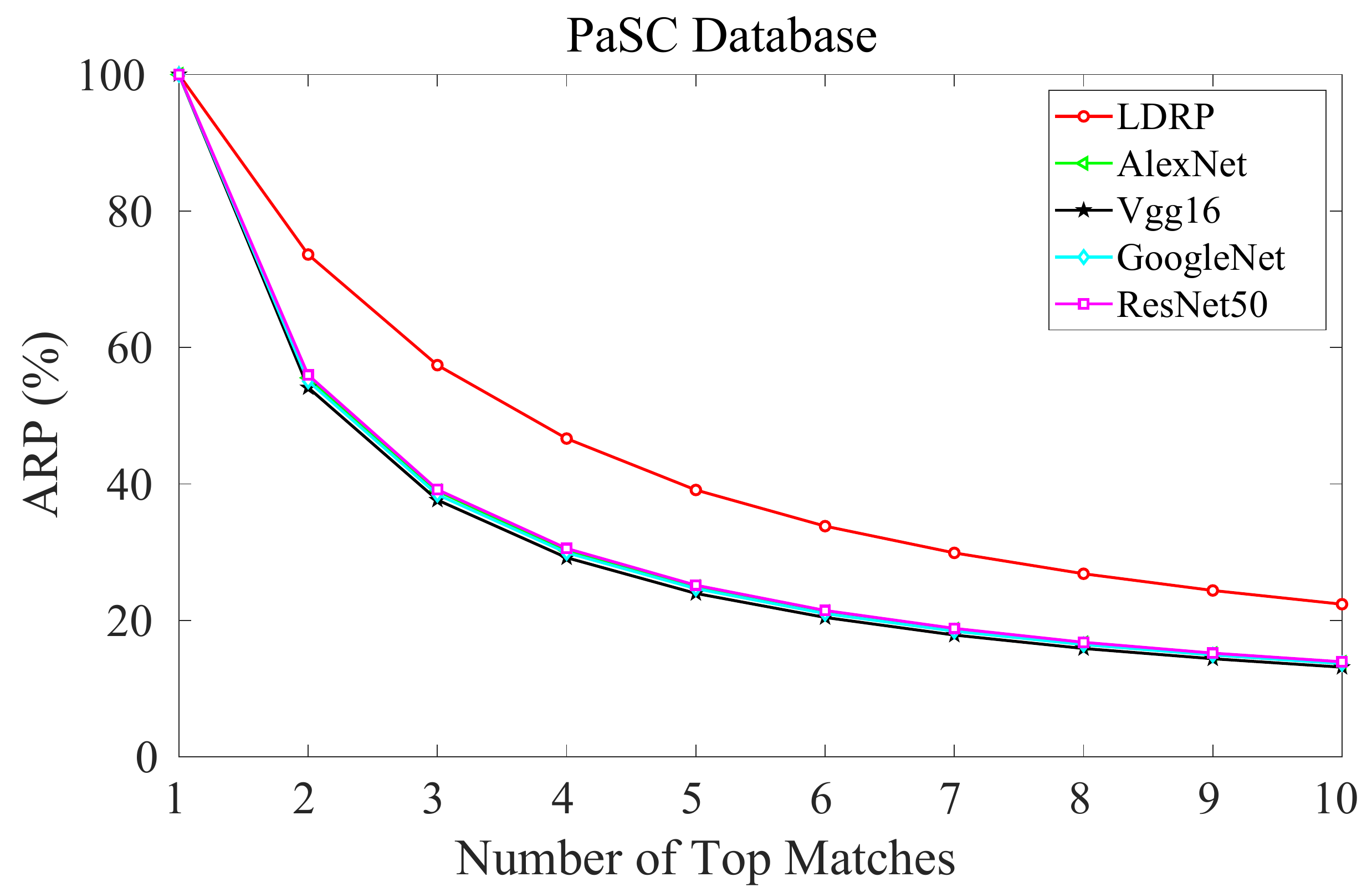}
    \caption{Over PaSC Database}
    \label{fig:pasc-cnn}
  \end{subfigure}
  
  \begin{subfigure}{.5\textwidth}
    \centering
    \includegraphics[width=.98\linewidth]{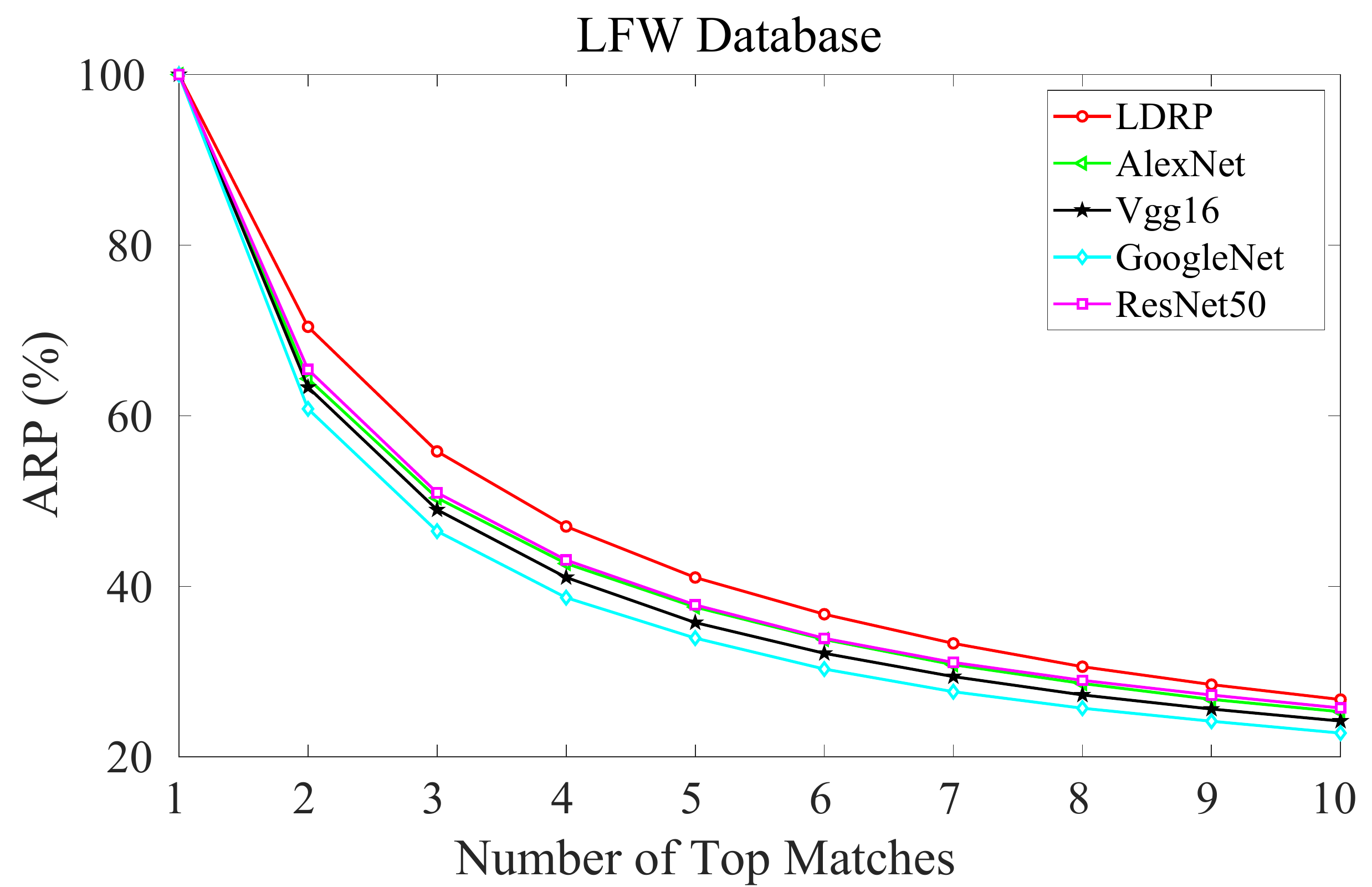}
    \caption{Over LFW Database}
    \label{fig:lfw-cnn}
  \end{subfigure}%
  \begin{subfigure}{.5\textwidth}
    \centering
    \includegraphics[width=.98\linewidth]{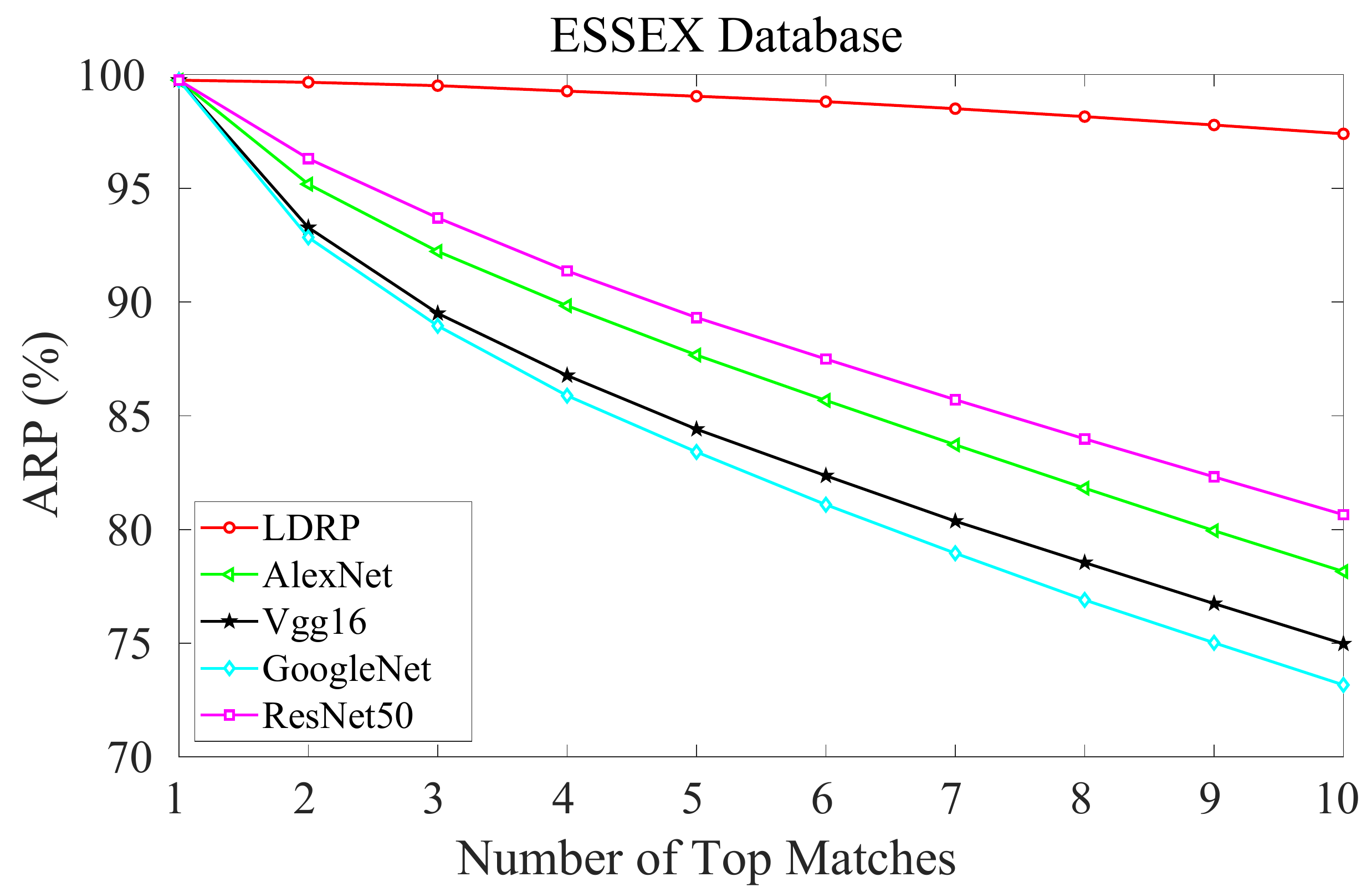}
    \caption{Over ESSEX Database}
    \label{fig:essex-cnn}
  \end{subfigure}
  
  \begin{subfigure}{.5\textwidth}
    \centering
    \includegraphics[width=.98\linewidth]{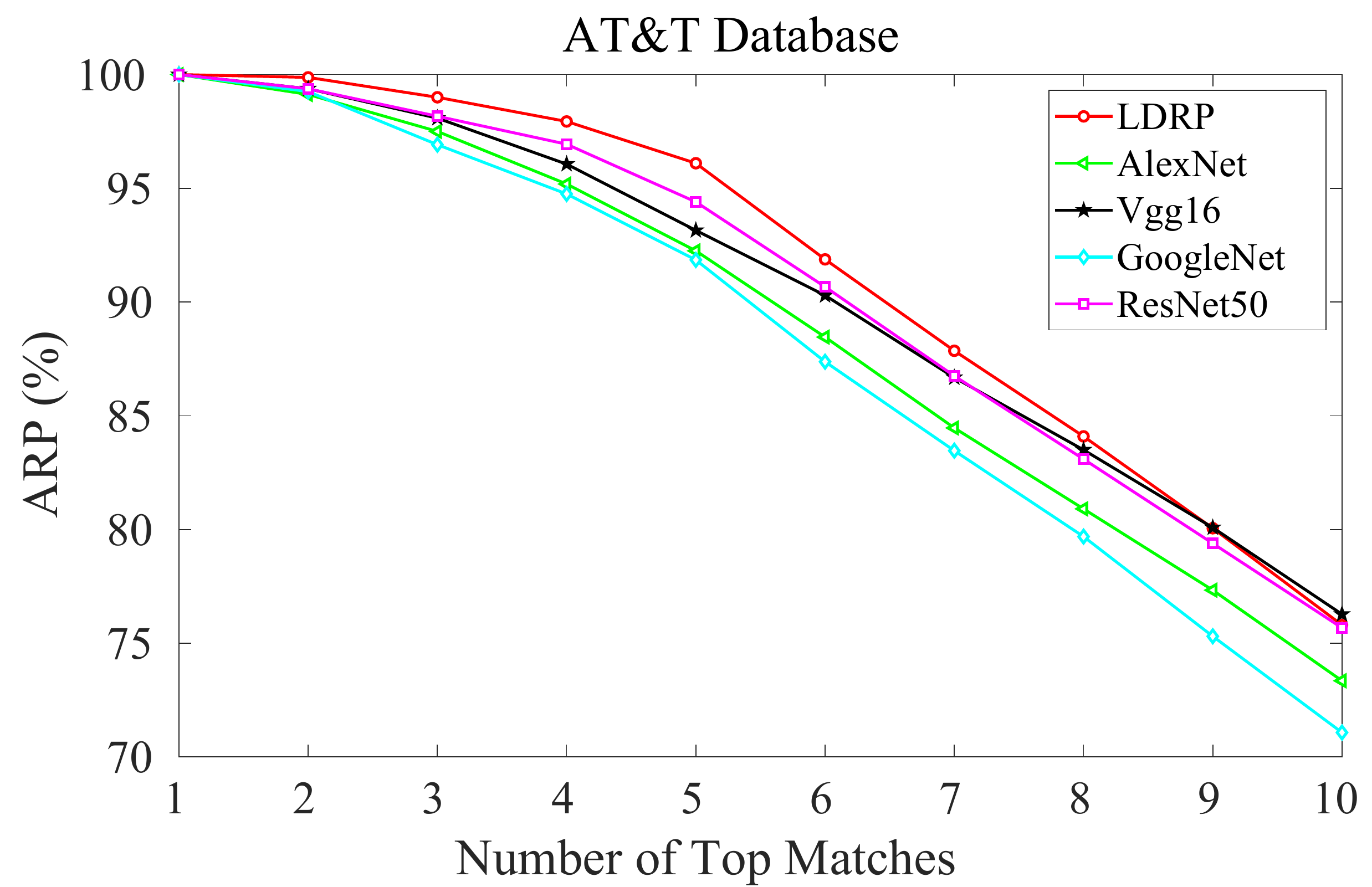}
    \caption{Over AT\&T Database}
    \label{fig:att-cnn}
  \end{subfigure}%
  \begin{subfigure}{.5\textwidth}
    \centering
    \includegraphics[width=.98\linewidth]{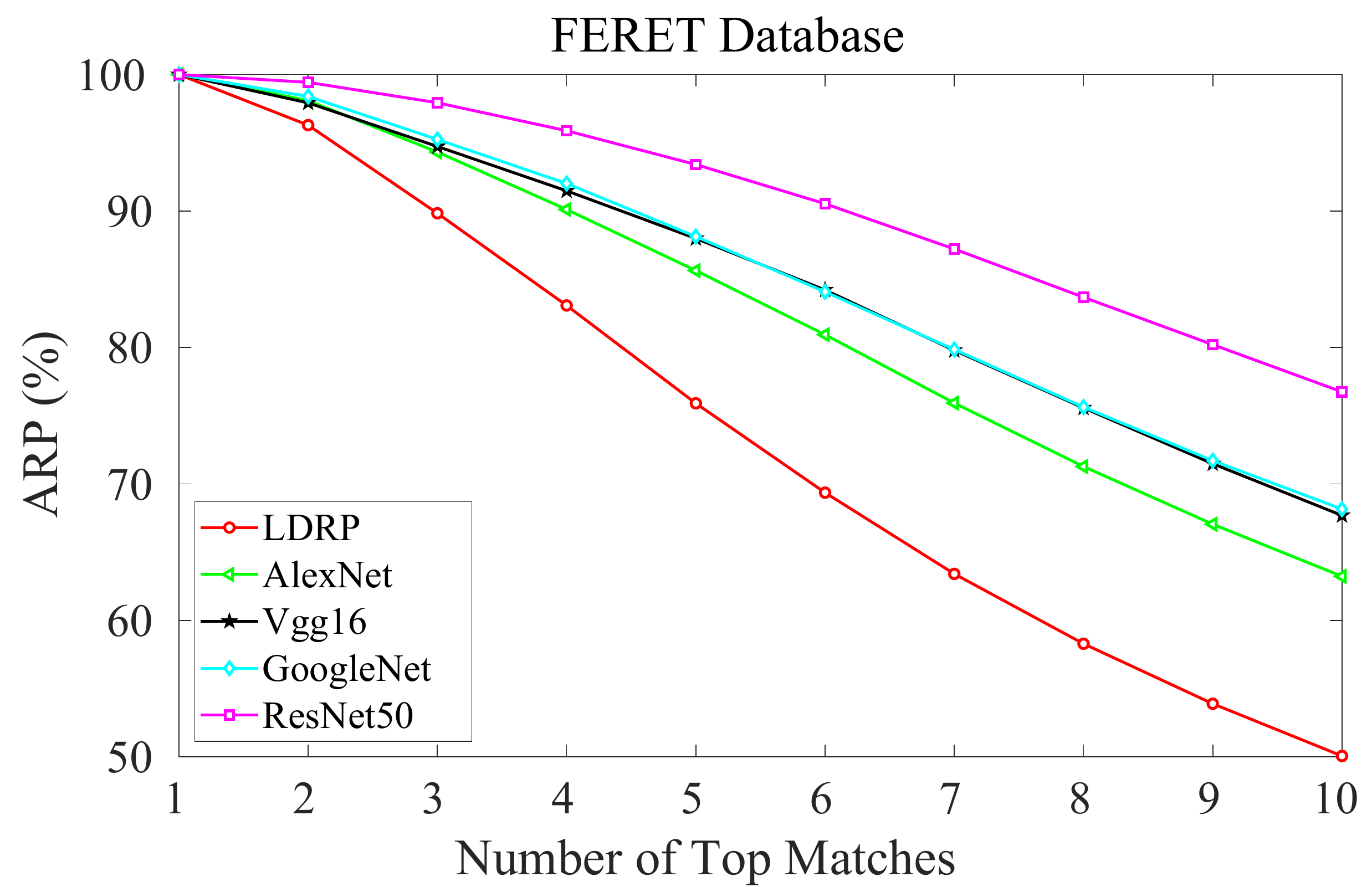}
    \caption{Over FERET Database}
    \label{fig:feret-cnn}
  \end{subfigure}
  
  \caption{The results comparison with ImageNet \cite{imagenet} pre-trained CNN models including AlexNet \cite{alex}, Vgg16 \cite{vgg}, GoogleNet \cite{googlenet}, and ResNet50 \cite{resnet}.}
  \label{fig:results-cnn}
\end{figure*}

\begin{figure}[!t]
  \centering
  \includegraphics[width=0.7\columnwidth]{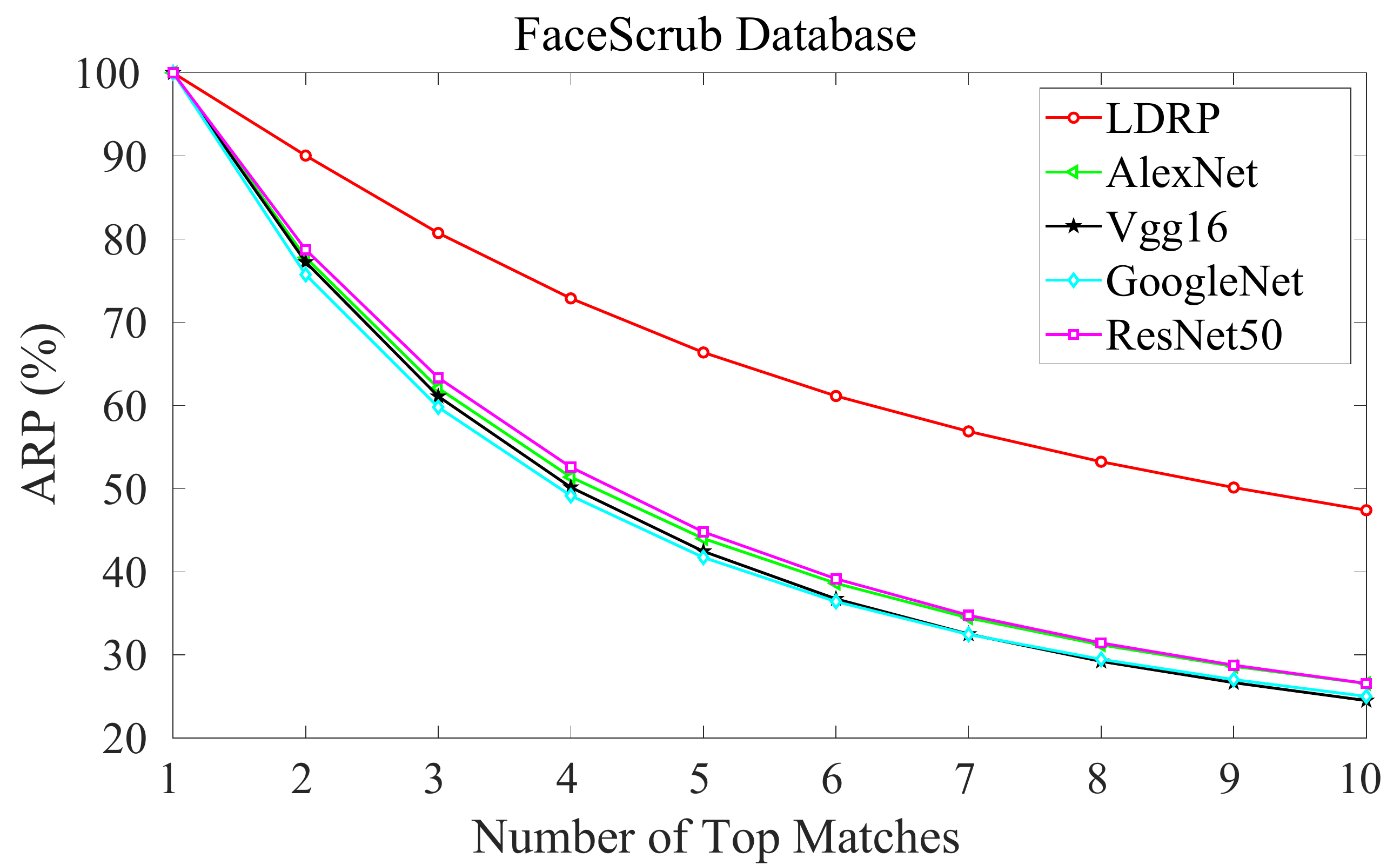}
  \caption{The results comparison over large-scale FaceScrub face dataset with ImageNet \cite{imagenet} pre-trained CNN models including AlexNet \cite{alex}, Vgg16 \cite{vgg}, GoogleNet \cite{googlenet}, and ResNet50 \cite{resnet}.}
  \label{fig:results-facescrub}
\end{figure}

\subsection{Comparison with ImageNet Pre-trained CNN Models}
In this section, face retrieval results of the proposed LDRP method are compared with the pre-trained CNN models. Several pre-trained CNN models are available in MATLAB. These models are trained over large-scale ImageNet database \cite{imagenet}. The AlexNet \cite{alex}, Vgg16 \cite{vgg}, GoogleNet \cite{googlenet}, and ResNet50 \cite{resnet} models are used for the comparison. These pre-trained models are used as a feature extractor and the absolute features of final dense layer of dimension 1000 is used as the descriptor. The comparison results over each dataset are plotted in Fig. \ref{fig:results-cnn}. It is clear from these results that the proposed LDRP feature outperforms the ImageNet based CNN features for most of the databases. Moreover, the CNN features are very sensitive to the blurred images as pointed out in Fig. \ref{fig:essex-cnn}. It shows the weakness of deep learning based features in  terms of the training data dependency. Thus, the proposed method is more generalized compared to CNN methods.

\subsection{Experiments over Large-Scale Face Dataset}
In order to show the scalability of the proposed method, this experiment is performed over large-scale FaceScrub dataset \cite{ng2014data} available from the challenge 1 of MegaFace challenge\footnote{http://megaface.cs.washington.edu/participate/challenge.html} \cite{kemelmacher2016megaface}. This dataset contains $91,712$ faces from $526$ subjects. The cropped version of FaceScrub dataset is used in this experiment. The first $10$ faces of each subject are used as the probe while the complete dataset is used as the gallery. The ARP over large-scale FaceSrub face dataset is presented in Fig. \ref{fig:results-facescrub}. The result is compared with the ImageNet \cite{imagenet} pre-trained CNN models, namely AlexNet \cite{alex}, Vgg16 \cite{vgg}, GoogleNet \cite{googlenet}, and ResNet50 \cite{resnet}. The proposed LDRP descriptor outperforms the pre-trained ImageNet CNN models over large-scale FaceScrub face dataset. This observation supports the scalability of the proposed LDRP face descriptor.

\begin{figure*}[!t]
  \begin{subfigure}{.5\textwidth}
    \centering
    \includegraphics[clip=true, trim = 0 0 0.8cm 16cm, width=.98\linewidth]{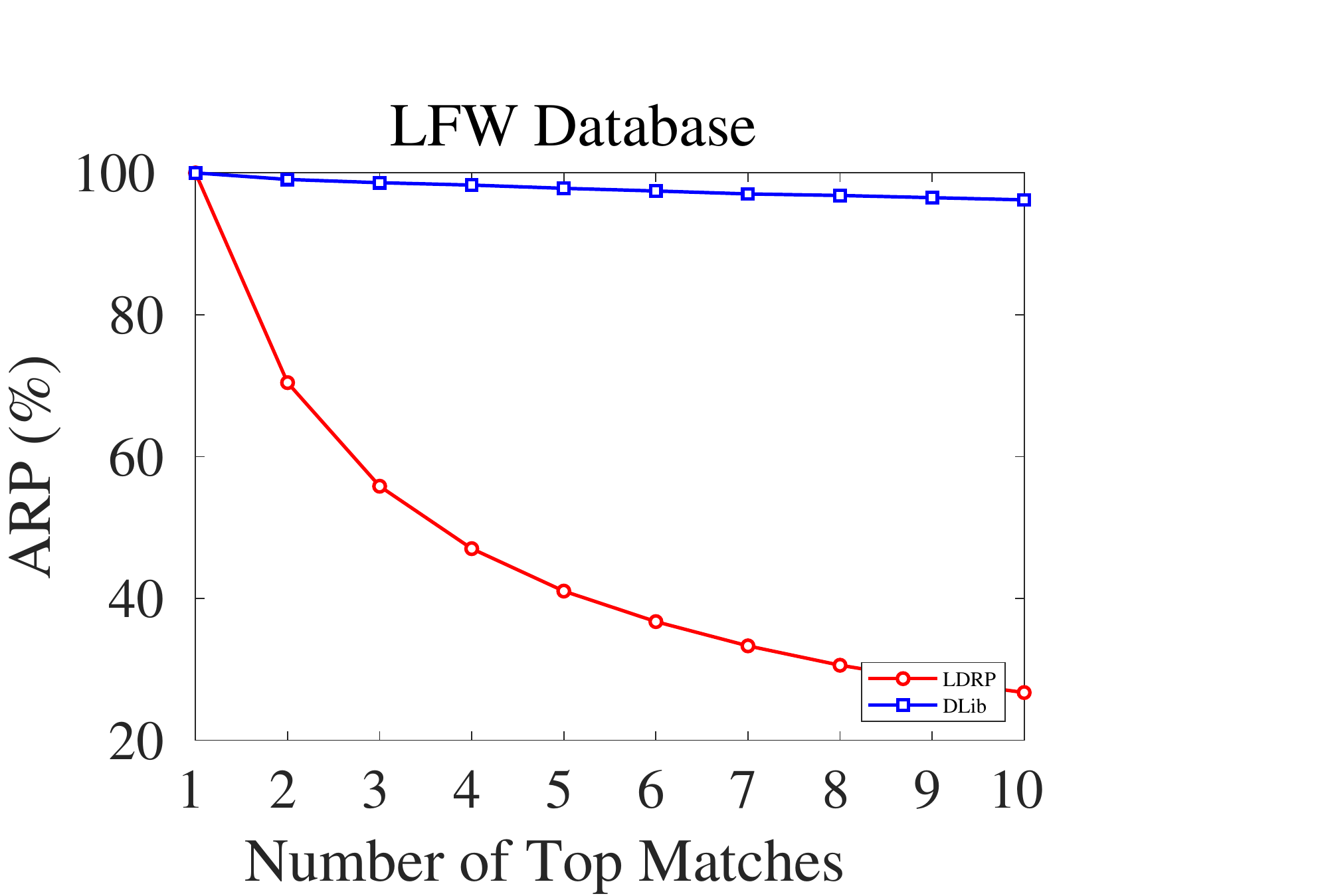}
    \caption{Over LFW Database}
    \label{fig:lfw-dlib}
  \end{subfigure}%
  \begin{subfigure}{.5\textwidth}
    \centering
    \includegraphics[clip=true, trim = 0 0 0.8cm 16cm, width=.98\linewidth]{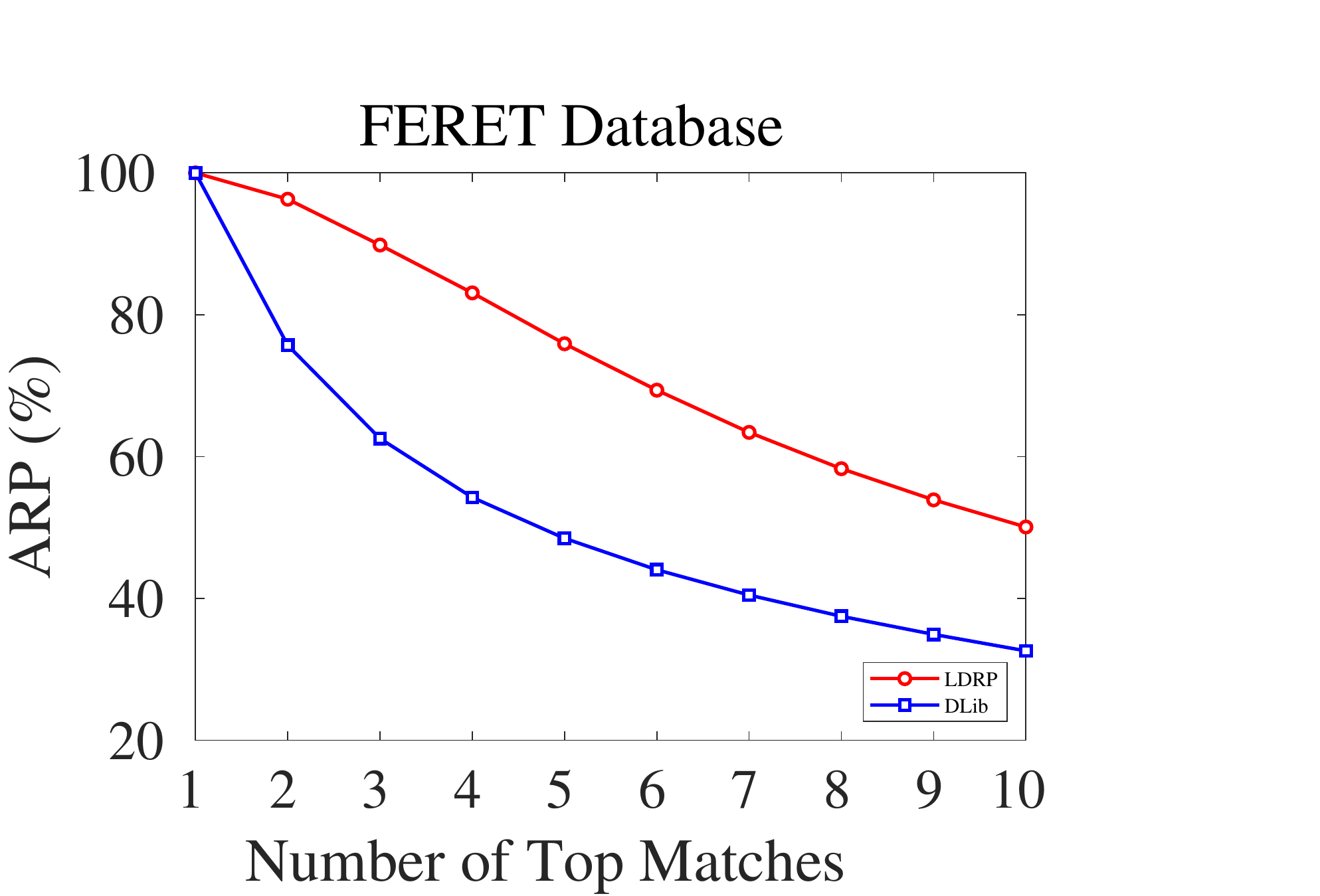}
    \caption{Over FERET Database}
    \label{fig:feret-dlib}
  \end{subfigure}
  
  \begin{subfigure}{.5\textwidth}
    \centering
    \includegraphics[clip=true, trim = 0 0 0.8cm 16cm, width=.98\linewidth]{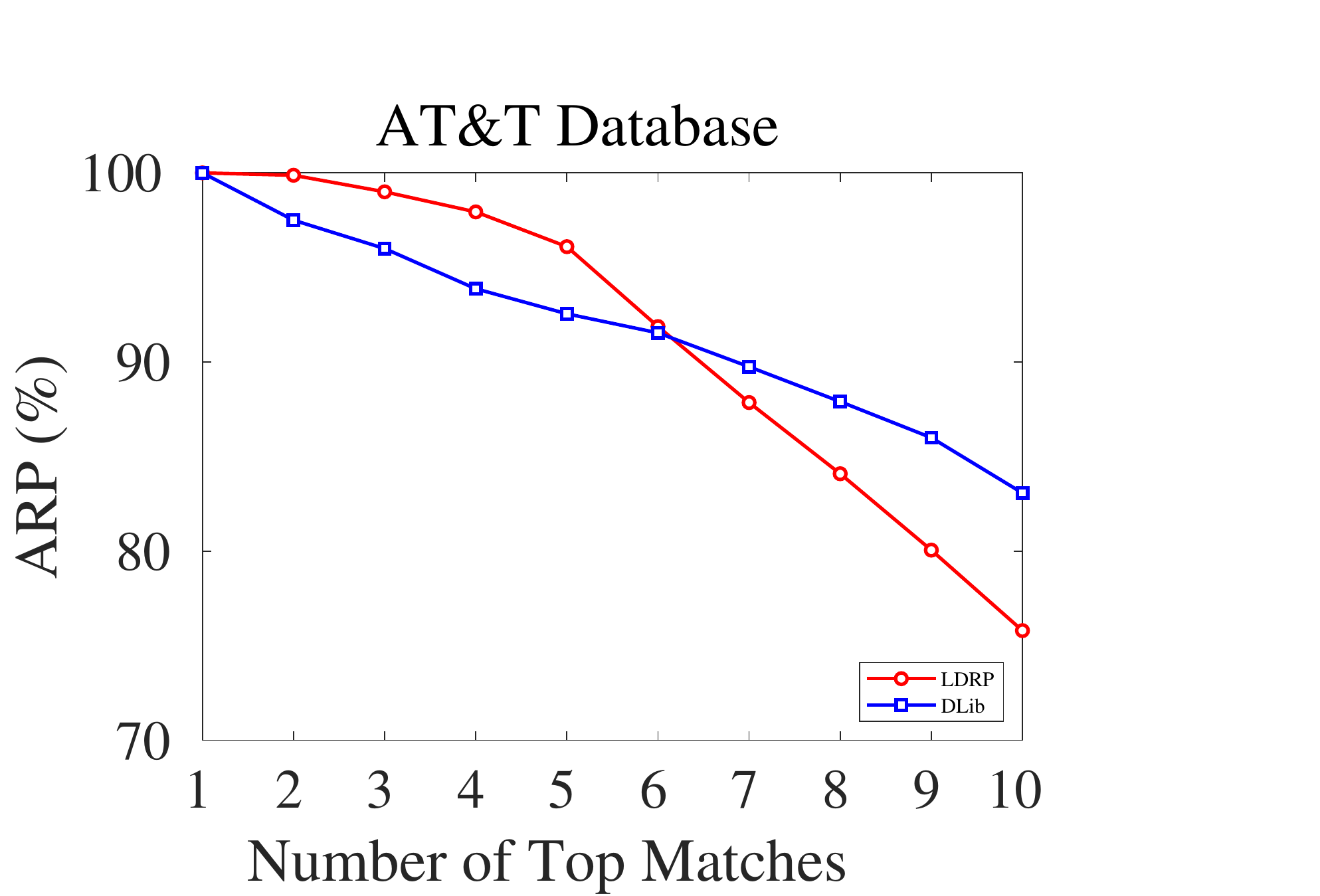}
    \caption{Over AT\&T Database}
    \label{fig:att-dlib}
  \end{subfigure}%
  \begin{subfigure}{.5\textwidth}
    \centering
    \includegraphics[clip=true, trim = 0 0 0.8cm 16cm, width=.98\linewidth]{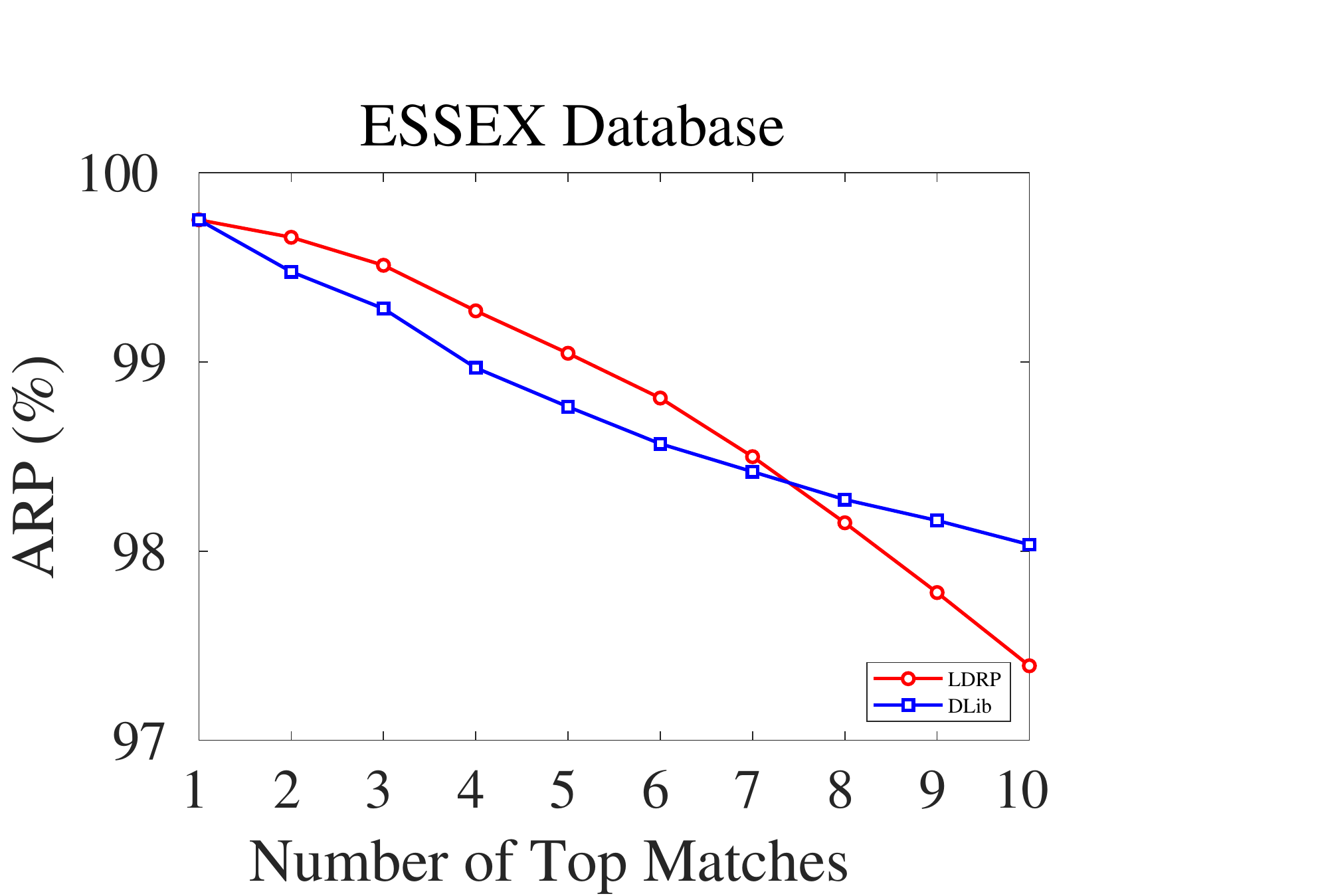}
    \caption{Over ESSEX Database}
    \label{fig:essex-dlib}
  \end{subfigure}
  
  \caption{The results comparison with deep learning based DLib face descriptor \cite{dlib} over LFW, FERET, AT\&T, and ESSEX datasets.}
  \label{fig:results-dlib}
\end{figure*}

\subsection{Comparison with Deep Learning based DLib Face Descriptor}
In order to show the importance of proposed LDRP method in terms of the less data dependency, its performance is compared with the deep learning based DLib face descriptor \cite{dlib}. The DLib descriptor is trained over face images and has shown very promising performance over LFW databases. The results comparison between hand-crafted LDRP and deep learning based DLib descriptors are illustrated in Fig. \ref{fig:results-dlib} over LFW, FERET, AT\&T, and ESSEX datasets. It can be seen that DLib completely outperforms LDRP over LFW dataset, because it is trained on such images. Whereas, over new face images in challenging scenariossuch as severe pose variations, background difference, blurred images, etc., the proposed LDRP descriptor outperforms the DLib descriptor. It can be visualized in the results over FERET, AT\&T, and ESSEX databases in Fig. \ref{fig:results-dlib}. It points out that the proposed LDRP descriptor is more generic and less data dependent than the DLib descriptor.

\section{Conclusion}
In this paper, a local directional relation pattern (LDRP) is proposed that utilizes the wider neighborhood information to increase the discriminative ability and local relations to increase the robustness. The LDRP first converts the wider local neighborhood into local directional codes in order to decrease the dimension of the descriptor by exploiting the relation among the directional neighbors at multiple radius, then it transforms the center pixel into the range of local directional relation codes, and finally the descriptor is computed by utilizing the relation of transformed center pixel with directional relation codes. The proposed LDRP descriptor is tested in an image retrieval framework over six very challenging face databases such as PaSC, LFW, FERET, etc. Some databases are totally unconstrained while some are having very severe variations in pose, expressions, etc. The retrieval results of LDRP are compared with the state-of-the-art face descriptors like LBP, LDN, DCP, LGHP, etc. The results are also compared in the recognition framework. The experimental results confirm the superiority of the LDRP descriptor as compared to the existing face descriptors. It is noticed that the performance of LDRP can be further boosted by considering more wider neighborhoods. The LDRP descriptor also outperforms the ImageNet pre-trained CNN models over each face dataset. Moreover, the proposed descriptor is scalable over large-scale dataset such as FaceScrub face dataset. The Chi-square distance measure is found to be best suited with LDRP for face retrieval. The image alignment can be used with LDRP to boost the discriminative power for unconstrained face recognition. The potential applications of the proposed descriptor are image matching, image retrieval, image stitching, texture classification, etc.

\section*{Acknowledgment}
This research is funded by IIIT Sri City, India through the Faculty Seed Research Grant.


\bibliographystyle{spmpsci}

\bibliography{reference}

\begin{thebibliography}{10}
\providecommand{\url}[1]{{#1}}
\providecommand{\urlprefix}{URL }
\expandafter\ifx\csname urlstyle\endcsname\relax
  \providecommand{\doi}[1]{DOI~\discretionary{}{}{}#1}\else
  \providecommand{\doi}{DOI~\discretionary{}{}{}\begingroup
  \urlstyle{rm}\Url}\fi

\bibitem{dlib}
Dlib face descriptor.
\newblock \url{https://github.com/ageitgey/face\_recognition}.
\newblock Accessed: 2019-05-01

\bibitem{lbp}
Ahonen, T., Hadid, A., Pietikainen, M.: Face description with local binary
  patterns: Application to face recognition.
\newblock IEEE transactions on pattern analysis and machine intelligence
  \textbf{28}(12), 2037--2041 (2006)

\bibitem{ahonen2008}
Ahonen, T., Rahtu, E., Ojansivu, V., Heikkila, J.: Recognition of blurred faces
  using local phase quantization.
\newblock In: Pattern Recognition, 2008. ICPR 2008. 19th International
  Conference on, pp. 1--4. IEEE (2008)

\bibitem{gemf}
Arandjelovic, O.: Gradient edge map features for frontal face recognition under
  extreme illumination changes.
\newblock In: BMVC 2012: Proceedings of the British machine vision association
  conference, pp. 1--11. BMVA Press (2012)

\bibitem{pasc}
Beveridge, J.R., Phillips, P.J., Bolme, D.S., Draper, B.A., Givens, G.H., Lui,
  Y.M., Teli, M.N., Zhang, H., Scruggs, W.T., Bowyer, K.W., et~al.: The
  challenge of face recognition from digital point-and-shoot cameras.
\newblock In: Biometrics: Theory, Applications and Systems (BTAS), 2013 IEEE
  Sixth International Conference on, pp. 1--8. IEEE (2013)

\bibitem{cao2010}
Cao, Z., Yin, Q., Tang, X., Sun, J.: Face recognition with learning-based
  descriptor.
\newblock In: Computer Vision and Pattern Recognition (CVPR), 2010 IEEE
  Conference on, pp. 2707--2714. IEEE (2010)

\bibitem{lghp}
Chakraborty, S., Singh, S., Chakraborty, P.: Local gradient hexa pattern: A
  descriptor for face recognition and retrieval.
\newblock IEEE Transactions on Circuits and Systems for Video Technology
  (2016)

\bibitem{ldgp}
Chakraborty, S., Singh, S.K., Chakraborty, P.: Local directional gradient
  pattern: a local descriptor for face recognition.
\newblock Multimedia Tools and Applications \textbf{76}(1), 1201--1216 (2017)

\bibitem{chan2013}
Chan, C.H., Tahir, M.A., Kittler, J., Pietik{\"a}inen, M.: Multiscale local
  phase quantization for robust component-based face recognition using kernel
  fusion of multiple descriptors.
\newblock IEEE Transactions on Pattern Analysis and Machine Intelligence
  \textbf{35}(5), 1164--1177 (2013)

\bibitem{wld}
Chen, J., Shan, S., He, C., Zhao, G., Pietikainen, M., Chen, X., Gao, W.: Wld:
  A robust local image descriptor.
\newblock IEEE transactions on pattern analysis and machine intelligence
  \textbf{32}(9), 1705--1720 (2010)

\bibitem{imagenet}
Deng, J., Dong, W., Socher, R., Li, L.J., Li, K., Fei-Fei, L.: Imagenet: A
  large-scale hierarchical image database.
\newblock In: Computer Vision and Pattern Recognition, 2009. CVPR 2009. IEEE
  Conference on, pp. 248--255. Ieee (2009)

\bibitem{dcp}
Ding, C., Choi, J., Tao, D., Davis, L.S.: Multi-directional multi-level
  dual-cross patterns for robust face recognition.
\newblock IEEE transactions on pattern analysis and machine intelligence
  \textbf{38}(3), 518--531 (2016)

\bibitem{ding2016comprehensive}
Ding, C., Tao, D.: A comprehensive survey on pose-invariant face recognition.
\newblock ACM Transactions on intelligent systems and technology (TIST)
  \textbf{7}(3), 37 (2016)

\bibitem{ding2015multi}
Ding, C., Xu, C., Tao, D.: Multi-task pose-invariant face recognition.
\newblock IEEE Transactions on Image Processing \textbf{24}(3), 980--993 (2015)

\bibitem{ding2012}
Ding, L., Ding, X., Fang, C.: Continuous pose normalization for pose-robust
  face recognition.
\newblock IEEE Signal Processing Letters \textbf{19}(11), 721--724 (2012)

\bibitem{iold}
Dubey, S.R., Singh, S.K., Singh, R.K.: Rotation and illumination invariant
  interleaved intensity order-based local descriptor.
\newblock IEEE Transactions on Image Processing \textbf{23}(12), 5323--5333
  (2014)

\bibitem{ldep}
Dubey, S.R., Singh, S.K., Singh, R.K.: Local diagonal extrema pattern: a new
  and efficient feature descriptor for ct image retrieval.
\newblock IEEE Signal Processing Letters \textbf{22}(9), 1215--1219 (2015)

\bibitem{lwp}
Dubey, S.R., Singh, S.K., Singh, R.K.: Local wavelet pattern: A new feature
  descriptor for image retrieval in medical ct databases.
\newblock IEEE Transactions on Image Processing \textbf{24}(12), 5892--5903
  (2015)

\bibitem{lbdp}
Dubey, S.R., Singh, S.K., Singh, R.K.: Local bit-plane decoded pattern: A novel
  feature descriptor for biomedical image retrieval.
\newblock IEEE Journal of Biomedical and Health Informatics \textbf{20}(4),
  1139--1147 (2016)

\bibitem{mdlbp}
Dubey, S.R., Singh, S.K., Singh, R.K.: Multichannel decoded local binary
  patterns for content-based image retrieval.
\newblock IEEE Transactions on Image Processing \textbf{25}(9), 4018--4032
  (2016)

\bibitem{elaiwat2014}
Elaiwat, S., Bennamoun, M., Boussaid, F., El-Sallam, A.: 3-d face recognition
  using curvelet local features.
\newblock IEEE Signal Processing Letters \textbf{21}(2), 172--175 (2014)

\bibitem{lvp}
Fan, K.C., Hung, T.Y.: A novel local pattern descriptor—local vector pattern
  in high-order derivative space for face recognition.
\newblock IEEE transactions on image processing \textbf{23}(7), 2877--2891
  (2014)

\bibitem{resnet}
He, K., Zhang, X., Ren, S., Sun, J.: Deep residual learning for image
  recognition.
\newblock In: Proceedings of the IEEE conference on computer vision and pattern
  recognition, pp. 770--778 (2016)

\bibitem{hong2015novel}
Hong, D., Liu, W., Su, J., Pan, Z., Wang, G.: A novel hierarchical approach for
  multispectral palmprint recognition.
\newblock Neurocomputing \textbf{151}, 511--521 (2015)

\bibitem{huang2011local}
Huang, D., Shan, C., Ardabilian, M., Wang, Y., Chen, L.: Local binary patterns
  and its application to facial image analysis: a survey.
\newblock IEEE Transactions on Systems, Man, and Cybernetics, Part C
  (Applications and Reviews) \textbf{41}(6), 765--781 (2011)

\bibitem{lfw}
Huang, G.B., Ramesh, M., Berg, T., Learned-Miller, E.: Labeled faces in the
  wild: A database for studying face recognition in unconstrained environments.
\newblock Tech. rep., Technical Report 07-49, University of Massachusetts,
  Amherst (2007)

\bibitem{lqp}
Hussain, S.U., Napol{\'e}on, T., Jurie, F.: Face recognition using local
  quantized patterns.
\newblock In: British machive vision conference, pp. 11--pages (2012)

\bibitem{jabid2010facial}
Jabid, T., Kabir, M.H., Chae, O.: Facial expression recognition using local
  directional pattern (ldp).
\newblock In: Image Processing (ICIP), 2010 17th IEEE International Conference
  on, pp. 1605--1608. IEEE (2010)

\bibitem{kan2016multi}
Kan, M., Shan, S., Zhang, H., Lao, S., Chen, X.: Multi-view discriminant
  analysis.
\newblock IEEE transactions on pattern analysis and machine intelligence
  \textbf{38}(1), 188--194 (2016)

\bibitem{kemelmacher2016megaface}
Kemelmacher-Shlizerman, I., Seitz, S.M., Miller, D., Brossard, E.: The megaface
  benchmark: 1 million faces for recognition at scale.
\newblock In: Proceedings of the IEEE Conference on Computer Vision and Pattern
  Recognition, pp. 4873--4882 (2016)

\bibitem{alex}
Krizhevsky, A., Sutskever, I., Hinton, G.E.: Imagenet classification with deep
  convolutional neural networks.
\newblock In: Advances in neural information processing systems, pp. 1097--1105
  (2012)

\bibitem{pubfig}
Kumar, N., Berg, A.C., Belhumeur, P.N., Nayar, S.K.: Attribute and simile
  classifiers for face verification.
\newblock In: Computer Vision, 2009 IEEE 12th International Conference on, pp.
  365--372. IEEE (2009)

\bibitem{lei2014}
Lei, Z., Pietik{\"a}inen, M., Li, S.Z.: Learning discriminant face descriptor.
\newblock IEEE Transactions on Pattern Analysis and Machine Intelligence
  \textbf{36}(2), 289--302 (2014)

\bibitem{dlbp}
Liao, S., Law, M.W., Chung, A.C.: Dominant local binary patterns for texture
  classification.
\newblock IEEE transactions on image processing \textbf{18}(5), 1107--1118
  (2009)

\bibitem{mblbp}
Liao, S., Zhu, X., Lei, Z., Zhang, L., Li, S.Z.: Learning multi-scale block
  local binary patterns for face recognition.
\newblock In: International Conference on Biometrics, pp. 828--837. Springer
  (2007)

\bibitem{liu2016recognizing}
Liu, L., Cheng, L., Liu, Y., Jia, Y., Rosenblum, D.S.: Recognizing complex
  activities by a probabilistic interval-based model.
\newblock In: AAAI, vol.~30, pp. 1266--1272 (2016)

\bibitem{brint}
Liu, L., Long, Y., Fieguth, P.W., Lao, S., Zhao, G.: Brint: binary rotation
  invariant and noise tolerant texture classification.
\newblock IEEE Transactions on Image Processing \textbf{23}(7), 3071--3084
  (2014)

\bibitem{liu2015action2activity}
Liu, Y., Nie, L., Han, L., Zhang, L., Rosenblum, D.S.: Action2activity:
  Recognizing complex activities from sensor data.
\newblock In: IJCAI, vol. 2015, pp. 1617--1623 (2015)

\bibitem{liu2016action}
Liu, Y., Nie, L., Liu, L., Rosenblum, D.S.: From action to activity:
  sensor-based activity recognition.
\newblock Neurocomputing \textbf{181}, 108--115 (2016)

\bibitem{lu2015simultaneous}
Lu, J., Erin~Liong, V., Zhou, J.: Simultaneous local binary feature learning
  and encoding for face recognition.
\newblock In: Proceedings of the IEEE International Conference on Computer
  Vision, pp. 3721--3729 (2015)

\bibitem{lu2015learning}
Lu, J., Liong, V.E., Zhou, X., Zhou, J.: Learning compact binary face
  descriptor for face recognition.
\newblock IEEE transactions on pattern analysis and machine intelligence
  \textbf{37}(10), 2041--2056 (2015)

\bibitem{anmrr}
Lu, K., He, N., Xue, J., Dong, J., Shao, L.: Learning view-model joint
  relevance for 3d object retrieval.
\newblock IEEE Transactions on Image Processing \textbf{24}(5), 1449--1459
  (2015)

\bibitem{lumini2017}
Lumini, A., Nanni, L., Brahnam, S.: Ensemble of texture descriptors and
  classifiers for face recognition.
\newblock Applied Computing and Informatics \textbf{13}(1), 79--91 (2017)

\bibitem{ltrp}
Murala, S., Maheshwari, R., Balasubramanian, R.: Local tetra patterns: a new
  feature descriptor for content-based image retrieval.
\newblock IEEE Transactions on Image Processing \textbf{21}(5), 2874--2886
  (2012)

\bibitem{ng2014data}
Ng, H.W., Winkler, S.: A data-driven approach to cleaning large face datasets.
\newblock In: 2014 IEEE International Conference on Image Processing (ICIP),
  pp. 343--347. IEEE (2014)

\bibitem{lbptexture}
Ojala, T., Pietikainen, M., Maenpaa, T.: Multiresolution gray-scale and
  rotation invariant texture classification with local binary patterns.
\newblock IEEE Transactions on pattern analysis and machine intelligence
  \textbf{24}(7), 971--987 (2002)

\bibitem{feret1}
Phillips, P.J., Moon, H., Rizvi, S.A., Rauss, P.J.: The feret evaluation
  methodology for face-recognition algorithms.
\newblock IEEE Transactions on pattern analysis and machine intelligence
  \textbf{22}(10), 1090--1104 (2000)

\bibitem{feret}
Phillips, P.J., Wechsler, H., Huang, J., Rauss, P.J.: The feret database and
  evaluation procedure for face-recognition algorithms.
\newblock Image and vision computing \textbf{16}(5), 295--306 (1998)

\bibitem{lbpbook}
Pietik{\"a}inen, M., Hadid, A., Zhao, G., Ahonen, T.: Local binary patterns for
  still images.
\newblock In: Computer vision using local binary patterns, pp. 13--47. Springer
  (2011)

\bibitem{abhi2015}
Punnappurath, A., Rajagopalan, A.N., Taheri, S., Chellappa, R., Seetharaman,
  G.: Face recognition across non-uniform motion blur, illumination, and pose.
\newblock IEEE Transactions on Image Processing \textbf{24}(7), 2067--2082
  (2015)

\bibitem{drldp}
PVSSR, C.M., et~al.: Dimensionality reduced local directional pattern (dr-ldp)
  for face recognition.
\newblock Expert Systems with Applications \textbf{63}, 66--73 (2016)

\bibitem{priclbp}
Qi, X., Xiao, R., Li, C.G., Qiao, Y., Guo, J., Tang, X.: Pairwise rotation
  invariant co-occurrence local binary pattern.
\newblock IEEE Transactions on Pattern Analysis and Machine Intelligence
  \textbf{36}(11), 2199--2213 (2014)

\bibitem{lgop}
Ren, C.X., Lei, Z., Dai, D.Q., Li, S.Z.: Enhanced local gradient order features
  and discriminant analysis for face recognition.
\newblock IEEE transactions on cybernetics \textbf{46}(11), 2656--2669 (2016)

\bibitem{ldn}
Rivera, A.R., Castillo, J.R., Chae, O.O.: Local directional number pattern for
  face analysis: Face and expression recognition.
\newblock IEEE transactions on image processing \textbf{22}(5), 1740--1752
  (2013)

\bibitem{ldtp}
Ryu, B., Rivera, A.R., Kim, J., Chae, O.: Local directional ternary pattern for
  facial expression recognition.
\newblock IEEE Transactions on Image Processing  (2017)

\bibitem{att}
Samaria, F.S., Harter, A.C.: Parameterisation of a stochastic model for human
  face identification.
\newblock In: Applications of Computer Vision, 1994., Proceedings of the Second
  IEEE Workshop on, pp. 138--142. IEEE (1994)

\bibitem{lfw1}
Sanderson, C., Lovell, B.C.: Multi-region probabilistic histograms for robust
  and scalable identity inference.
\newblock In: International Conference on Biometrics, pp. 199--208. Springer
  (2009)

\bibitem{facenet}
Schroff, F., Kalenichenko, D., Philbin, J.: Facenet: A unified embedding for
  face recognition and clustering.
\newblock In: Proceedings of the IEEE Conference on Computer Vision and Pattern
  Recognition, pp. 815--823 (2015)

\bibitem{vgg}
Simonyan, K., Zisserman, A.: Very deep convolutional networks for large-scale
  image recognition.
\newblock arXiv preprint arXiv:1409.1556  (2014)

\bibitem{essex}
Spacek, L.: University of essex face database (2002).
\newblock \urlprefix\url{http://cswww.essex.ac.uk/mv/allfaces/}

\bibitem{googlenet}
Szegedy, C., Liu, W., Jia, Y., Sermanet, P., Reed, S., Anguelov, D., Erhan, D.,
  Vanhoucke, V., Rabinovich, A.: Going deeper with convolutions.
\newblock In: Proceedings of the IEEE conference on computer vision and pattern
  recognition, pp. 1--9 (2015)

\bibitem{deepface}
Taigman, Y., Yang, M., Ranzato, M., Wolf, L.: Deepface: Closing the gap to
  human-level performance in face verification.
\newblock In: Proceedings of the IEEE conference on computer vision and pattern
  recognition, pp. 1701--1708 (2014)

\bibitem{ltp}
Tan, X., Triggs, B.: Enhanced local texture feature sets for face recognition
  under difficult lighting conditions.
\newblock IEEE transactions on image processing \textbf{19}(6), 1635--1650
  (2010)

\bibitem{tang20133d}
Tang, H., Yin, B., Sun, Y., Hu, Y.: 3d face recognition using local binary
  patterns.
\newblock Signal Processing \textbf{93}(8), 2190--2198 (2013)

\bibitem{viola}
Viola, P., Jones, M.: Rapid object detection using a boosted cascade of simple
  features.
\newblock In: Computer Vision and Pattern Recognition, 2001. CVPR 2001.
  Proceedings of the 2001 IEEE Computer Society Conference on, vol.~1, pp.
  I--I. IEEE (2001)

\bibitem{vu2013}
Vu, N.S.: Exploring patterns of gradient orientations and magnitudes for face
  recognition.
\newblock IEEE Transactions on Information Forensics and Security
  \textbf{8}(2), 295--304 (2013)

\bibitem{wolf2011}
Wolf, L., Hassner, T., Taigman, Y.: Effective unconstrained face recognition by
  combining multiple descriptors and learned background statistics.
\newblock IEEE transactions on pattern analysis and machine intelligence
  \textbf{33}(10), 1978--1990 (2011)

\bibitem{wright2009}
Wright, J., Yang, A.Y., Ganesh, A., Sastry, S.S., Ma, Y.: Robust face
  recognition via sparse representation.
\newblock IEEE transactions on pattern analysis and machine intelligence
  \textbf{31}(2), 210--227 (2009)

\bibitem{xie2010fusing}
Xie, S., Shan, S., Chen, X., Chen, J.: Fusing local patterns of gabor magnitude
  and phase for face recognition.
\newblock IEEE transactions on image processing \textbf{19}(5), 1349--1361
  (2010)

\bibitem{yang2013comparative}
Yang, B., Chen, S.: A comparative study on local binary pattern (lbp) based
  face recognition: Lbp histogram versus lbp image.
\newblock Neurocomputing \textbf{120}, 365--379 (2013)

\bibitem{ldp}
Zhang, B., Gao, Y., Zhao, S., Liu, J.: Local derivative pattern versus local
  binary pattern: face recognition with high-order local pattern descriptor.
\newblock IEEE transactions on image processing \textbf{19}(2), 533--544 (2010)

\bibitem{hgpp}
Zhang, B., Shan, S., Chen, X., Gao, W.: Histogram of gabor phase patterns
  (hgpp): A novel object representation approach for face recognition.
\newblock IEEE Transactions on Image Processing \textbf{16}(1), 57--68 (2007)

\bibitem{lgbphs}
Zhang, W., Shan, S., Gao, W., Chen, X., Zhang, H.: Local gabor binary pattern
  histogram sequence (lgbphs): A novel non-statistical model for face
  representation and recognition.
\newblock In: Computer Vision, 2005. ICCV 2005. Tenth IEEE International
  Conference on, vol.~1, pp. 786--791. IEEE (2005)

\bibitem{zhang2009face}
Zhang, X., Gao, Y.: Face recognition across pose: A review.
\newblock Pattern Recognition \textbf{42}(11), 2876--2896 (2009)

\bibitem{zhao2003face}
Zhao, W., Chellappa, R., Phillips, P.J., Rosenfeld, A.: Face recognition: A
  literature survey.
\newblock ACM computing surveys (CSUR) \textbf{35}(4), 399--458 (2003)

\end{thebibliography}

\end{document}